\documentclass[twocolumn]{article}

\usepackage[sc]{mathpazo} 
\usepackage[T1]{fontenc} 
\usepackage{microtype} 

\usepackage[left=.7in, right=.7in, top=32mm,columnsep=15pt]{geometry} 
\usepackage{multicol} 
\usepackage[hang, small,labelfont=bf,up,textfont=it,up]{caption} 
\usepackage{booktabs} 
\usepackage{float} 
\usepackage{hyperref} 

\usepackage{paralist} 

\usepackage{abstract} 

\usepackage{titlesec} 
\usepackage[pdftex]{graphicx}
\usepackage{paralist,overpic}
\usepackage{fancyhdr,array}
\usepackage{footmisc,xcolor}
\usepackage{listings}
\usepackage{fancyvrb}
\usepackage{amsmath,epsfig,amssymb,bbm,stmaryrd,mathrsfs}
\usepackage{pstricks}
\usepackage{pstricks}
\usepackage{subfigure}
\usepackage{wrapfig}

\DeclareMathAlphabet{\mathpzc}{OT1}{pzc}{m}{it}

\newcommand{\argmin}{\operatornamewithlimits{arg\;\!min\;}}

\newcommand{\x}{\boldsymbol{x}}

\newcommand{\argmax}{\operatornamewithlimits{arg\;\!max\;}}

\RequirePackage{algorithm,overpic}
\RequirePackage[noend]{algpseudocode}

\renewcommand\thesection{\Roman{section}} 
\renewcommand\thesubsection{\Roman{subsection}} 
\titleformat{\section}[block]{\large\scshape\centering}{\thesection.}{1em}{} 
\titleformat{\subsection}[block]{\large}{\thesubsection.}{1em}{} 

\usepackage{fancyhdr} 
\title{\vspace{-15mm}\fontsize{18pt}{10pt}\selectfont\textbf{Sweep Distortion Removal from THz Images via Blind Demodulation}} 

\author{
\large
\textsc{Alireza Aghasi${}^*$, Barmak Heshmat${}^*$,}\\ \textsc{Albert Redo-Sanchez${}^*$, Justin Romberg${}^\dagger$}
\\ \textsc{and Ramesh Raskar${}^*$}\\[2mm] 
\normalsize ${}^*$ Massachusetts Institute of Technology\\${}^\dagger$ \normalsize Georgia Institute of Technology \\ 
\normalsize \href{mailto:aghasi@mit.edu}{aghasi@mit.edu} 
\vspace{-5mm}
}
\date{}

\begin{document}

\maketitle 

\thispagestyle{fancy} 


\begin{abstract}
Heavy sweep distortion induced by alignments and inter-reflections of layers of a sample is a major burden in recovering 2D and 3D information in time resolved spectral imaging. This problem cannot be addressed by conventional denoising and signal processing techniques as it heavily depends on the physics of the acquisition. Here we propose and implement an algorithmic framework based on low-rank matrix recovery and alternating minimization that exploits the forward model for THz acquisition. The method allows recovering the original signal in spite of the presence of temporal-spatial distortions. We address a blind-demodulation problem, where based on several observations of the sample texture modulated by an undesired sweep pattern, the two classes of signals are separated. The performance of the method is examined in both synthetic and experimental data, and the successful reconstructions are demonstrated. The proposed general scheme can be implemented to advance inspection and imaging applications in THz and other time-resolved sensing modalities.
\end{abstract}



\section{Introduction}
Due to fine time resolution and broad spectral coverage, terahertz time domain spectroscopy (THz-TDS) has become a leading method in THz spectroscopy \cite{shen2011terahertz,tonouchi2007cutting}, imaging \cite{seco2013goya} and nondestructive testing \cite{redo2013review} of dielectric structures. There is an extensive literature on improving the THz imaging capability. In general, a large body of literature is focused on improving the hardware as THz-TDS power levels are usually at sub microwatt levels \cite{heshmat2012nanoplasmonic, heshmat2012carbon, heshmat2013enhanced}. This perspective can be extended to improving THz acquisition methodologies such as compressive \cite{chan2008single} and wide field acquisitions \cite{li2013nipkow, shen2012spinning, lee2005real}. On the signal processing side the mainstream research has been focused on improvement of signal to noise ratio (SNR) \cite{chen2010frequency, galvao2003optimal}. Many of these SNR improvement methods are developed for transmission mode spectroscopy and have appreciable alignment with infrared spectroscopy \cite{deng2010terahertz, burnett2009broadband, yin2012terahertz}. However, when it comes to imaging, inspection, and content extraction of surfaces and layered structures, THz-TDS can suffer significantly from phase distortions along the sample surface. These sweeping distortions in THz time domain imaging appear as dominant challenge for in depth imaging and content extraction in densely layered structures, irregular surfaces or any 2D geometry. Unfortunately, these heavy distortions are directly induced by the nature of the sample and THz-TDS measurement scheme, and cannot be addressed by minor hardware improvement, conventional denoising or SNR enhancement techniques.

In this paper we propose and demonstrate a mathematical framework that enables demodulation of the recorded signal from sweep distoritions induced by slight depth variations or the layered structure inter-reflections. We view the problem as a demodulation of the distortion profiles from the sample texture, based on the reflected wave measured at different instances of time. The problem of interest is modeled as a bilinear inverse problem (BIP), which can be addressed using a low-rank matrix recovery framework. We propose an alternating minimization scheme, where a prior structure is considered for each factor in the BIP. More specifically, we assume the distortion profiles belong to a low dimensional subspace (which can be extracted from the raw data) and the layer structure is of binary nature. The model considered for the layer texture is in fact a shape-based approximation (see \cite{aghasi2013geometric, aghasi2011parametric, miller2012environmental} for examples), a phase corresponding to the main texture and another phase representing the anomalies and inclusions. Using the given priors for each factor in the BIP, we alternatively solve the demodulation problem to exclude the undesired sweep distortions from the THz images.

Our mathematical presentation mainly relies on multidimensional calculus. We use bold characters to denote vectors and matrices. Considering a matrix $\boldsymbol{A}$ and the index sets $\Gamma_1$, and $\Gamma_2$, we use $\boldsymbol{A}_{\Gamma_1,:}$ to denote the matrix obtained by restricting the rows of $\boldsymbol{A}$ to $\Gamma_1$. Similarly, $\boldsymbol{A}_{:,\Gamma_2}$ denotes the restriction of $\boldsymbol{A}$ to the columns specified by $\Gamma_2$, and $\boldsymbol{A}_{\Gamma_1,\Gamma_2}$ is the submatrix with the rows and columns restricted to $\Gamma_1$ and $\Gamma_2$, respectively.

The remainder of this paper is organized as follows. In Section \ref{sec:phys} we overview the physics of the problem and discuss the main demodulation problem to be addressed. In Section \ref{sec:meth} we present the main algorithmic framework to address the demodulation problem using an alternating minimization scheme. Finally, in section \ref{sec:exp} the  sensitivity and performance of the algorithm are assessed with synthetic data. We further experimentally demonstrate the blind demodulation of the data recorded from both single and multilayered structures and report the algorithm outcomes for experimental data.  We conclude this section with some remarks and future directions of research.

\section{Physics of the Problem}\label{sec:phys}

\begin{figure}[t]
\begin{tabular}{cccc}
\hspace{-1.4cm}\begin{overpic}[trim={7.5cm 7.4cm  6.3cm 8cm},clip, width=1.2in]{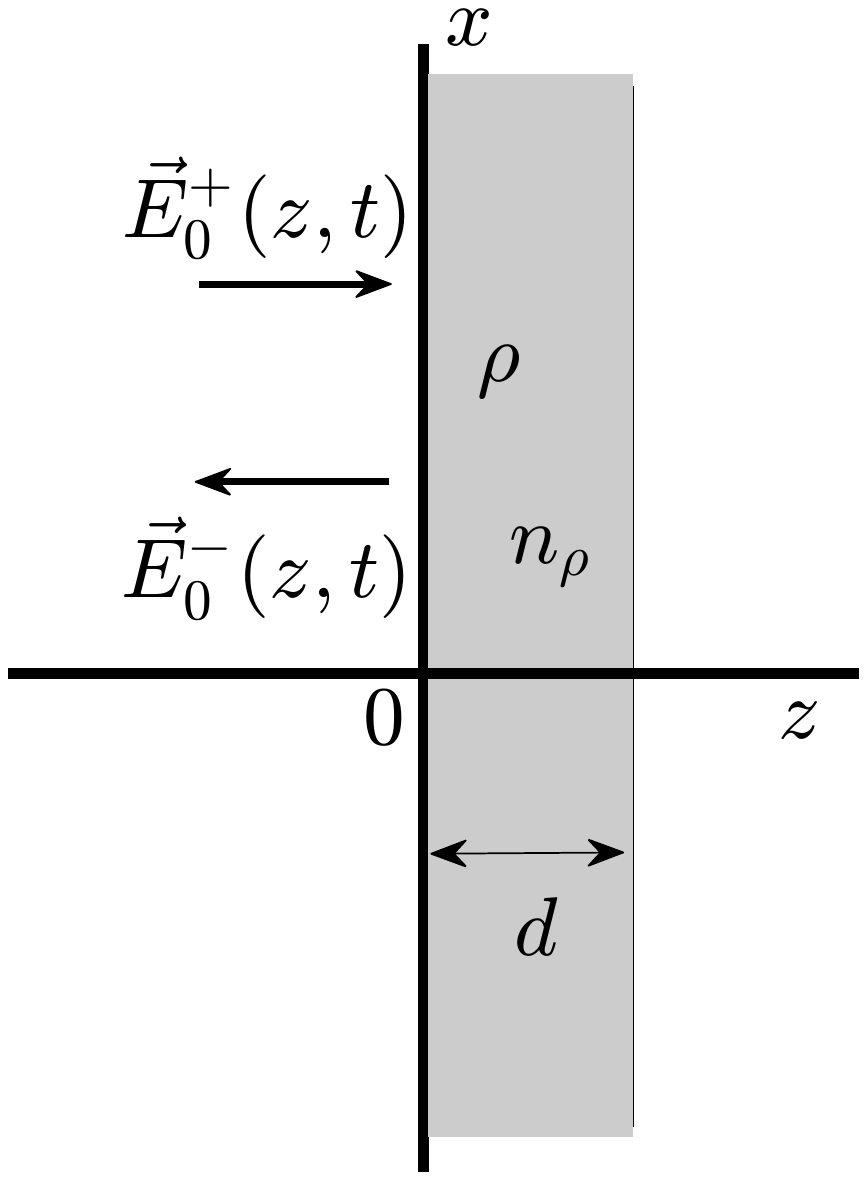}
 \put (26,-3) {(a)} \end{overpic} &\hspace{-1.6cm} \begin{overpic}[trim={0 -1cm  0cm 0},clip, width=2.22in, height=1.8in]{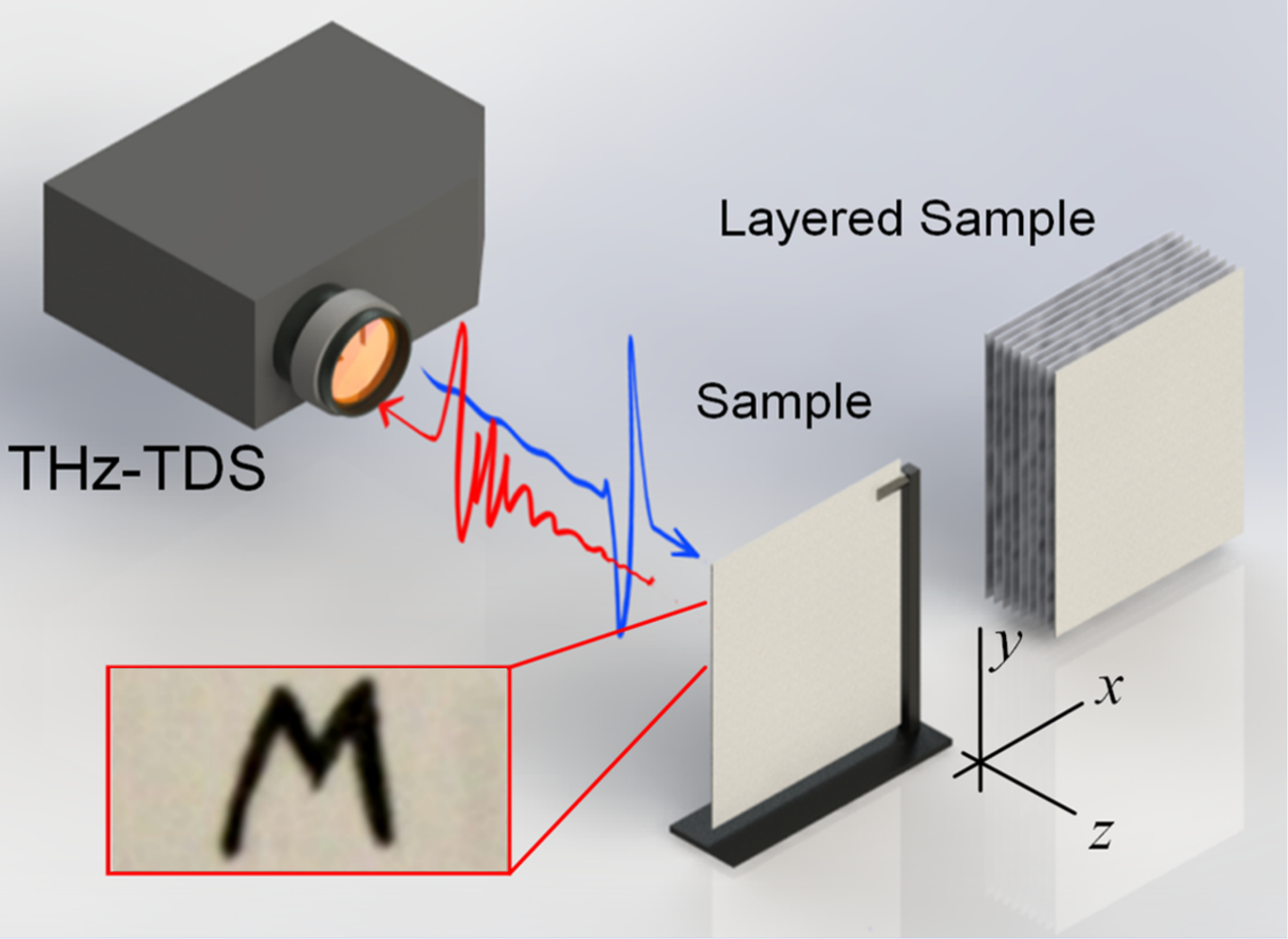}
 \put (47,-3) {(b)} \end{overpic}
\\[.2cm]
\hspace{-.2cm}\begin{overpic}[trim={4.5cm 7cm 4.5cm 7.5cm}, clip,width=1.67in,height=1.1in]{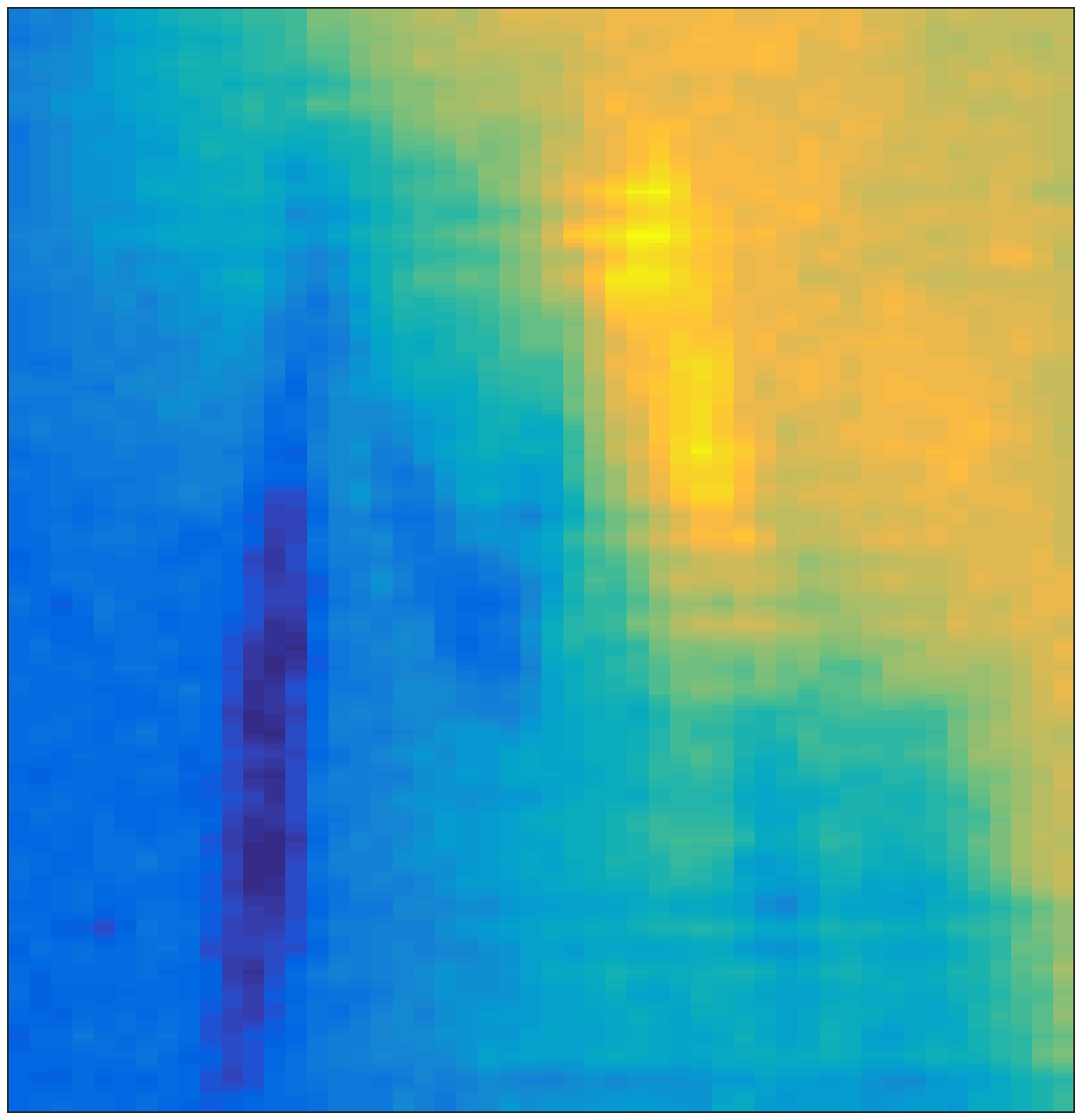}\put (45,-6) {(c)} \end{overpic}
&
\hspace{-.1cm}\begin{overpic}[trim={4.5cm 7cm 4.5cm 7.5cm},clip,width=1.67in,height=1.1in]{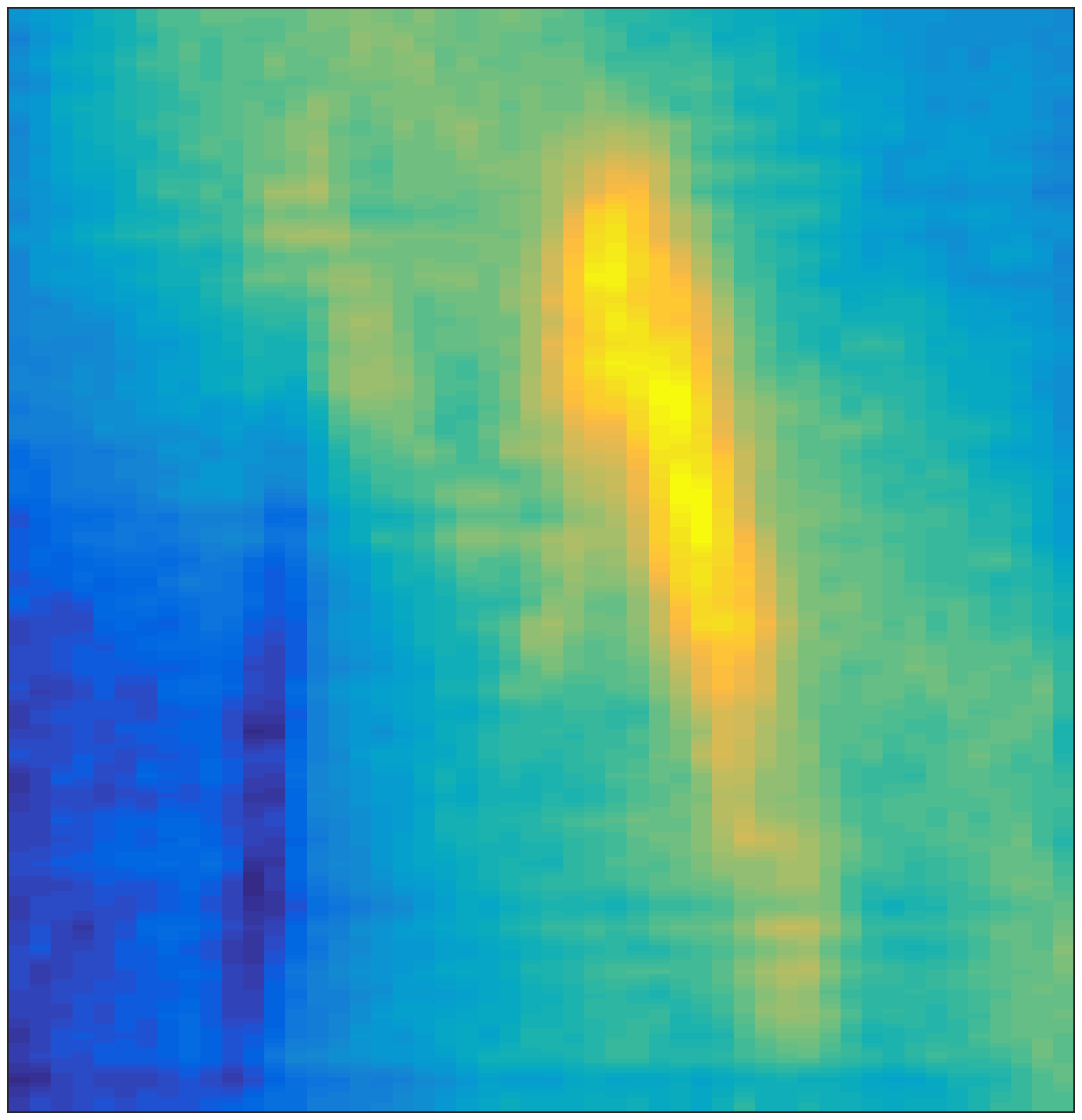}\put (45,-6) {(d)} \end{overpic}
\end{tabular}
\caption{ (a) A dielectric slab, emitted and reflected electrical fields; (b) setup schematics, blue is the emitted THz field and the red waveform is indicative of a typical reflected signal in one pixel; (c,d) induction of sweep distortion: observation of the returned field at two different time instances}\label{fig1}
\end{figure}

Consider an electric field that is linearly polarized in the $x$-direction, propagating along the $z$-direction in the free space. The waveform is considered to be a finite duration THz pulse $\chi(t)$, for which the traveling field along every point $(x,y)$ is
\begin{equation*}
\vec{E}_0^+(z,t) = \chi(t-\frac{z}{c})\vec{a}_x,
\end{equation*}
and $c$ is the wave speed. Due to confocal nature of the measurement at each $(x,y)$ position, the problem of interest can be considered as extracting the contents of a dielectric slab placed perpendicular to the wave propagation axis, based on analyzing the reflected field $\vec{E}_0^-(z,t)$ (Figure \ref{fig1}(a), (b)).

For a homogeneous layer of width $d$, with reflection coefficient $\rho$ and refraction index $n_\rho$, the returned signal can be analytically expressed by convolving the pulse $\chi(.)$ with a train of impulse functions with decaying coefficients. More specifically \cite{scharstein1992transient},
\begin{equation}\label{eqp2}
\vec{E}_0^-(z,t) = \rho u\left(t+\frac{z}{c}\right)\vec{a}_x,
\end{equation}
where 
\begin{equation}\label{eqp3}
u(\tau) = \chi(\tau)\ast \left( \delta(\frac{\tau}{\tau_\rho}) - \frac{1-\rho^2}{\rho^2} \sum_{m=1}^\infty\rho^{2m}\delta\left(\frac{\tau}{\tau_\rho} - 2m\right) \right),
\end{equation}
and $\tau_\rho = n_\rho d/c$. 
Each impulse term in (\ref{eqp3}) corresponds to a reflection. Particularly, the first term corresponds to the reflection from the front surface of the slab, the second term ($m=1$) corresponds to the back surface reflection and the remaining terms correspond to the returned waves after a number of inter-reflections within the slab.

The response in (\ref{eqp2}) can still be reasonably accurate when the dielectric slab is homogeneous along the $z$-axis, i.e., $\rho(x,y,z) = \rho_0(x,y)$ for $0\leq z\leq d$. It only requires plugging the point-wise values of $\rho$ and $\tau_\rho$ in the formulation.

In the case of perfect or close to perfect reflection, where $\rho\approx 1$, the first impulse term in (\ref{eqp2}) dominates all the other terms and the returned wave is a simple modulation of $\rho$ with the pulse waveform. When $0<\rho\ll 1$, the most dominant terms in $u(\tau)$ correspond to the first and second impulses, and the remaining terms decay exponentially fast. Specifically, in many practical scenarios where the pulse width is sufficiently small and the slab width $d$ is large, the first reflected waveforms does not overlap with the subsequent reflections. In this case, still a modulation of $\rho$ with the pulse waveform is observed in different time intervals.

For multilayer dielectric slabs, the impulse response can still be cast as an impulse train, however deriving closed-form expressions for the coefficients is a sophisticated task and beyond the scope of this paper. When the reflection coefficients of the layers are small (hence transmission is the leading phenomenon), the dominant terms associated with different layers remain reasonably distinct from one another. If the pulse width is sufficiently small, for each layer the corresponding dominant reflection is still proportional to the reflection coefficient of that layer. 

In all the situations stated above, for a measurement location $(x,y,z_0)$ with a fixed $z$ component, and sampling time $t=t_0$, the reflected signal is $\rho(x,y)u(t_0+z_0/c)$, which is a constant multiple of $\rho(x,y)$. In other words, the reflected images should contain the same pattern as $\rho$. However, due to non-ideal configurations such as a tilted or uneven sample, the effective measurement points are $(x,y,z_0 + \varepsilon(x,y))$, where $\varepsilon(x,y)$ is a function of small magnitude. For instance, in the case of a tilted sample $\varepsilon(x,y) = \alpha_1 x+\alpha_2 y$, where $\alpha_1$ and $\alpha_2$ are small constants.

Technically, the demodulation problem of interest in this paper corresponds to extracting the sample structure $\rho(x,y)$, based on observing the reflected signal $\vec{E}_0^-(x,y,z,t)$ at a fixed $z$ coordinate and time samples $t=t_1,\cdots,t_M$. Due to the non-ideal configurations stated above, the returned signal, sampled at a time $t_j$, is in the form of $\rho(x,y)u_j(x,y)+n_j(x,y)$, where $u_j(x,y)$ depends on the pulse waveform, sampling time and $\varepsilon(x,y)$. The additive term $n_j(x,y)$ models the noise uncertainty and undesired electromagnetic interactions. We will refer to the $u_j(x,y)$ factor as a sweep distortion profile, which is often a slowly varying function when the misconfiguration is small.

%

Specifically, inspired by the applications presented in this paper, we are interested in single or multilayer composite slabs of binary nature, where for each layer
\begin{equation}\label{eqcat}
\rho(x,y,z) = \left\{\begin{array}{lc} \rho^0 & (x,y)\in D_0\\ \rho^1 & (x,y)\in D_1\end{array}\right .,
\end{equation}
and $\rho^0$ and $\rho^1$ are constant. An example of a composite slab which roughly follows such model is a card marked with deep ink (see example in Figure \ref{fig1}(b)).

Figures \ref{fig1}(c), (d) show examples of the observed reflection from a binary slab (letter ``M'' printed on a card) at two different time instances. We are willing to make a binary characterization of the sample content by demodulating and excluding the multiplicative sweep distortion profiles affecting the desired image. In the sequel we will present an inversion scheme to characterize the slab contents based on such multiplicative observations.

%

\section{Methods}\label{sec:meth}
\subsection{General Setup}\label{sec:meth:gen}
For a more concise formulation we use the more compact spatial notation $\x=(x,y)$. Based on the discussions in the previous section, our observation is in the form of 
\begin{equation}\label{eq1}
y_j(\x) = \rho(\x)u_j(\x) + n_j(\x),\quad j=1,\cdots,M,\;\;\;\x\in D,
\end{equation}
where $y_j(\x)$ represents the $j$-th observed reflected image dependent on the planar coordinate $\x$, $\rho(\x)$ is the true dielectric texture profile common across all the layers, $u_j(\x)$ is the sweep distortion profile corrupting layer $j$, $n_j(\x)$ models the noise and uncertainty and $D$ is the domain of imaging for $\x$. As a reasonable model, we consider a normal distribution for the noise, that is, $n_j(\x)\sim\mathcal{N}(0,\sigma^2)$ are independent and identically distributed (i.i.d) random variables for $j=1,\cdots,M$ and $\x\in D$.

The ultimate goal is recovering the image $\rho(\x)$ and the distortion profile $u_j(\x)$ based on the observation of $y_j(\x)$. Our strategy to address the problem is considering known subspace models for the signals $u_j(\x)$. More specifically, $u_j(\x)\in\mathcal{S}^j$, where $\mathcal{S}^j = \mbox{span}\big(s^j_1(\x),\cdots,s^j_{N_j}(\x)\big)$ and $N_j$ is the $j$-th subspace dimension. In Section \ref{sec:subspace} we introduce an algorithm to construct a set of low dimensional subspaces $\mathcal{S}^1,\cdots, \mathcal{S}^M$ from the set of observations $y_1(\x), \cdots y_M(\x)$. Also, based on the structure of the problem, we may benefit from prior assumptions about $\rho(\x)$ as will be detailed in the sequel.

\subsubsection{Problem Reformulation as a Low-Rank Recovery Scheme}\label{lowrank}
Consider discretizing the domain $D$ into a collection of pixels $\{\x_i\}_{i=1}^P$. Based on the model in (\ref{eq1}), a natural way of inverting the data for the image and sweep distortion profiles is through the following minimization:
\begin{equation}\label{eq2}
\min_{\boldsymbol{\rho},\boldsymbol{u}_1,\cdots,\boldsymbol{u}_M}\sum_{j=1}^M\Big\|\boldsymbol{y}_j - \boldsymbol{\rho}\odot \boldsymbol{u}_j\Big\|_2^2\!\quad s.t. \quad \! \boldsymbol{u}_j \in \mbox{span}\left(\boldsymbol{s}^j_1,\cdots,\boldsymbol{s}^j_{N_j}\right).
\end{equation}
Here $\odot$ denotes the Hadamard (pointwise) product and the bold scripts are vectors of length $P$, corresponding to the variables in (\ref{eq1}). Consider $\boldsymbol{S}^j$ to be a matrix representation of the subspace $\mathcal{S}^j$ and $\boldsymbol{Q}$ is a subspace in which the signal $\boldsymbol{\rho}$ lives (in the case of no such assumption, $\boldsymbol{Q}$ can be simply the identity matrix spanning the canonical basis). Under this assumption, the minimization in (\ref{eq2}) can be cast as
\begin{equation}\label{eq3}
\min_{\boldsymbol{\beta},\boldsymbol{\alpha}_1,\cdots,\boldsymbol{\alpha}_M}\sum_{j=1}^M \Big \|\boldsymbol{y}_j - (\boldsymbol{Q}\boldsymbol{\beta})\odot (\boldsymbol{S}^j\boldsymbol{\alpha}_j)\Big\|_2^2,
\end{equation}
where the search for the unknown vectors is performed in the corresponding subspaces. Suppose $\boldsymbol{\alpha} = [{\boldsymbol{\alpha}_1}^T, \cdots \boldsymbol{\alpha}_M^T]^T$ and $\boldsymbol{P}_j$ to be a suitable selection matrix (subset of the rows of the identity matrix) such that $\boldsymbol{\alpha}_j = \boldsymbol{P}_j\boldsymbol{\alpha}$. For the point-wise product of the image and the sweep distortion profiles we have
\begin{align}\nonumber
\big(\boldsymbol{\rho}\odot \boldsymbol{u}_j\big)_i &=\;\;\! (\boldsymbol{Q}\boldsymbol{\beta})_i\odot \big(\boldsymbol{S}^{j}\boldsymbol{\alpha_j}\big)_i\;\;=\big(\boldsymbol{Q}_{i,:}\boldsymbol{\beta}\big) \boldsymbol{S}^{j}_{i,:}\boldsymbol{P}_j\boldsymbol{\alpha}\\\nonumber &= \big(\boldsymbol{Q}_{i,:}\boldsymbol{\beta}\big)^T \boldsymbol{S}^{j}_{i,:}\boldsymbol{P}_j\boldsymbol{\alpha} = \mbox{tr}\Big(\boldsymbol{\alpha}\boldsymbol{\beta}^T \boldsymbol{Q}_{i,:}^T \boldsymbol{S}^{j}_{i,:}\boldsymbol{P}_j\Big)\\ &= \Big \langle \boldsymbol{\beta}\boldsymbol{\alpha}^T, \boldsymbol{Q}_{i,:}^T \boldsymbol{S}^{j}_{i,:}\boldsymbol{P}_j \Big\rangle.\label{eq4}
\end{align}
In other words considering $\boldsymbol{X} = \boldsymbol{\beta}\boldsymbol{\alpha}^T$ to be the outer-product of the unknown vectors $\boldsymbol{\beta}$ and $\boldsymbol{\alpha}$, the noise-free observations correspond to the inner-products of $\boldsymbol{X}$ with known matrices $\boldsymbol{Q}_{i,:}^T \boldsymbol{S}^{j}_{i,:}\boldsymbol{P}_j$. Basically, if we define a linear operator
\begin{equation}
\Big(\mathcal{A}(\boldsymbol{X})\Big)_{i,j} = \Big \langle \boldsymbol{X}, \boldsymbol{Q}_{i,:}^T \boldsymbol{S}^{j}_{i,:}\boldsymbol{P}_j \Big\rangle, \label{eq5}
\end{equation}
the noise-free observations are linearly dependent on $\boldsymbol{X}$, while the original version of the problem in (\ref{eq3}) is nonlinear in terms of $\boldsymbol{\beta}$ and $\boldsymbol{\alpha}$. However, since the outer-product of two nonzero vectors is a rank-one matrix, an equivalent reformulation of (\ref{eq3}) is the minimization
\begin{equation}
\min_{\boldsymbol{X}}\;\;\mbox{rank}(\boldsymbol{X})\quad s.t. \quad \big\| \boldsymbol{Y} - \mathcal{A}(\boldsymbol{X})\big\|_F \leq \epsilon,  \label{eq5.5}
\end{equation}
where $\boldsymbol{Y} = [\boldsymbol{y}_1,\cdots,\boldsymbol{y}_M]$ and $\|.\|_F$ denotes the Frobenius norm. Despite the linear relationship between the unknown variable and the noise-free observations, the rank constraint in (\ref{eq5.5}) makes it a combinatorial problem. As a remedy, a well-known approximation strategy is to use the nuclear norm of $\boldsymbol{X}$, denoted by $\|\boldsymbol{X}\|_*$, as a convex proxy to $\mbox{rank}(\boldsymbol{X})$:
\begin{equation}
\min_{\boldsymbol{X}}\;\;\|\boldsymbol{X}\|_*\quad s.t. \quad \big\| \boldsymbol{Y} - \mathcal{A}(\boldsymbol{X})\big\|_F \leq \epsilon.  \label{eq5.6}
\end{equation}
It is an interesting fact that under certain conditions on $\mathcal{A}$ (e.g., see \cite{recht2010guaranteed}), the approximate solution of (\ref{eq5.6}) can coincide with the solution of (\ref{eq5.5}). Once a rank-one solution $\boldsymbol{X}^*$ is available, the factors $\boldsymbol{\alpha}$ and $\boldsymbol{\beta}$ can be determined up to a constant multiple ($\boldsymbol{\beta}=\boldsymbol{X}_{:,1}^*$ and $\boldsymbol{\alpha}^T = \boldsymbol{X}_{1,:}^*/\boldsymbol{X}_{1,1}^*$).

Other than the relaxation technique, there are other methods reported in the literature to address instances of (\ref{eq5.5}). We are specifically interested in alternating minimization approach \cite{jain2013low}, since it allows incorporating prior information into the reconstruction of each bilinear factor.

\subsubsection{Alternating Minimization: A Maximum a Posteriori Framework}

In minimizing the non-convex program (\ref{eq2}) a reliable technique would be to setup an alternating minimization scheme. Basically, by fixing any of the operands in the bilinear term in (\ref{eq2}) (either $\rho$ or $u_j$) the problem turns into a standard least squares problem. The process would be to initialize one of the operands in the bilinear model, say $\rho$; with this quantity fixed, the resulting least squares in terms of $u_j$ is solved. By plugging in the acquired solutions for $u_j$'s, we can solve another least squares problem in terms of $\rho$, and proceed with such alternation until convergence.

This alternation approach is capable of approximating the solution to (\ref{eq5.5}). Again under certain incoherence conditions on $\mathcal{A}$ (slightly stricter than the ones stated for nuclear norm minimization \cite{jain2013low, recht2010guaranteed}), the approximate solution coincides with the solution of (\ref{eq5.5}). The major advantage of using an alternation scheme over the nuclear norm surrogate is the possibility of incorporating prior information beyond the subspace constraint on each factors of the problem in (\ref{eq1}) (e.g., see \cite{lee2013near}).

Basically, the proposed scheme corresponds to an alternation between the maximum likelihood (ML) estimates of $\rho$ and $u_j$.  When some level of prior knowledge about the $\rho$ and/or $u_j$ exists, the framework could be generalized to an alternation between the maximum a posteriori (MAP) estimates of $\rho$ and $u_j$, as will be detailed in the sequel.

Consider $f(\rho,\{u_j\}_{j=1}^M\big|\{y_j\}_{j=1}^M)$ to be the joint probability density function (pdf) of $\rho$ and $u_j$'s given the measurements $y_j$. The MAP estimates of $\rho$ and $u_j$ correspond to the maximizing parameters of the posterior distribution:
\begin{equation}\label{eq6}
\Big\{\rho^{MAP}, \{u^{MAP}_{j}\}_{j=1}^M\Big\} = \argmax_{\boldsymbol{\rho},\boldsymbol{u}_1,\cdots,\boldsymbol{u}_M} f(\rho,\{u_j\}_{j=1}^M\big|\{y_j\}_{j=1}^M).
\end{equation}
As a natural generalization, the alternating MAP estimation corresponds to the process sketched in Algorithm \ref{alg:1}.
\begin{algorithm}[htb!]
\caption{MAP Alternation Scheme}\label{alg:1}
\begin{algorithmic}[1]
\State $k\gets 0$, initialize $\rho^{(0)}$
\Repeat
\State $\{u^{(k)}_{j}\}_{j=1}^M\! =\! \argmax_{u_1,\cdots,u_M} f(\{u_j\}_{j=1}^M \big|\rho^{(k)}, \{y_j\}_{j=1}^M)$
\State $\rho^{(k+1)} = \argmax_{\boldsymbol{\rho}} f(\rho\big|\{u^{(k)}_{j}\}_{j=1}^M, \{y_j\}_{j=1}^M)$
\State $k\gets k+1$
\Until{convergence}
\end{algorithmic}
\end{algorithm}

In the sequel we elaborate on how to implement the proposed scheme using certain facts about the problem. We would like to note that we make some reasonable assumptions and provide some slight modifications in implementing the steps outlined in Algorithm \ref{alg:1}.

\subsubsection{Iterative Reconstruction of the Sweep Distortion Profiles}\label{sec:subspace}
In this section we mainly focus on addressing the first stage of Algorithm \ref{alg:1} (line 3), which is determining the distortion profiles $u_j$ based on the observations $\{y_j\}_{i=1}^M$ and the knowledge of $\rho$. 

To address this problem we only assume that based on the physics of the problem, the signals $u_j$ reside in a low dimensional subspace. In other words, $\boldsymbol{u}_j = \boldsymbol{S}^{j}\boldsymbol{\alpha}_j$ for the discrete representation. We however make no prior assumptions about the coefficient vector $\boldsymbol{\alpha}_j$. Since $\boldsymbol{\rho}\odot\boldsymbol{u}_j = \mbox{diag}(\boldsymbol{\rho})\boldsymbol{u}_j$, we simply consider solving the least squares problems of the form 
\begin{equation}\label{eq8}
\boldsymbol{\alpha}_j^{(k)} = \argmin_{\boldsymbol{\alpha}}\Big\|\boldsymbol{y}_j-\mbox{diag}(\boldsymbol{\rho}^{(k)})\boldsymbol{S}^{j}\boldsymbol{\alpha} \Big\|_2^2.
\end{equation}
The solution to (\ref{eq8}) is 
\begin{equation*}
\boldsymbol{\alpha}_j^{(k)} =\Big(\mbox{diag}(\boldsymbol{\rho}^{(k)})\boldsymbol{S}^j \Big)^\dagger\boldsymbol{y}_j\;,
\end{equation*}
where $\boldsymbol{A}^\dagger$ denotes the pseudoinverse of $\boldsymbol{A}$. In a sense, by making no assumptions about $\boldsymbol{\alpha}_j$, instead of performing a MAP estimation (as outlined in Algorithm \ref{alg:1}), for a given $\rho^{(k)}$ we are performing an ML estimation of the sweep distortion profiles by solving the least squares problems (\ref{eq8}). 

%
%
%

The choice of the subspace plays a key role for this problem. In the remainder of this subsection we present a one-time process which generates embedding subspaces $\boldsymbol{S}^{j}$ based on the observations $\boldsymbol{y}_j$.

From a technical standpoint, $y_j(\x) \approx \rho(\x)u_j(\x)$, where as stated earlier, $u_j$ is a rather smooth function and for our application $\rho(\x)$ is a function of almost binary nature. Therefore, if we have a multi-scale representation of $y_j$, there should be a good overlap between $u_j$ and the course-level approximation of $y_j$. The high frequency components of $y_j$ should be mainly in hold of $\rho$ and the noise. 

Based on this argument, consider a wavelet representation of the signal $y_j$ as
\begin{equation*}
y_j(\x)=\sum_w c_{\ell_0,w}\phi_{\ell_0,w}(\x) + \sum_{\ell=\ell_0}^\infty \sum_w d_{\ell,w}\psi_{\ell,w}(\x),
\end{equation*}
where $\phi_{\ell_0,w}$ are the scaling functions, $\psi_{\ell,w}$ are the wavelet basis and $\ell_0$ is a fixed scaling level \cite{strang1996wavelets}. For a fixed subspace order $N_j$, our proposed subspace selection for $u_j$ corresponds to the top $N_j$ scaling/wavelet basis functions with the largest coefficients (in magnitude). In other words, we sort the wavelet coefficients in descending magnitude order and select the basis functions associated with the top $N_j$ coefficients. The selected basis (in discrete form) correspond to the columns of $\boldsymbol{S}^{j}$. This process is performed only once during the entire reconstruction and once the subspaces are determined, they remain fixed throughout the alternation scheme. 

\subsubsection{Iterative Reconstruction of the Image Profile}
 
Inspired by the second stage proposed in Algorithm \ref{alg:1}, the main objective in this section is finding the MAP estimate for $\rho(\x)$. In order to maintain a more compact formulation and closed-form expressions we make some independence assumptions about the pixel values as will be detailed in the sequel. 


Based on the binary model considered, we assume that each pixel in the image $\rho$ belongs to one of the two classes 0 or 1, and denote the pixel class by $\mathcal{C}(\x)$. Following the categorization in (\ref{eqcat}), the pixel values in class 0 concentrate around $\rho^0$ and the pixel values associated with class 1 concentrate around $\rho^1$. Consider $\boldsymbol{\rho}= [\rho_1,\cdots,\rho_P]^T$ and $\boldsymbol{\mathcal{C}}=[\mathcal{C}_1,\cdots,\mathcal{C}_P]^T$ to be vectors containing the pixel values and the class labels for the discrete image. 

For our estimation problem, we are willing to address the maximization problem
\begin{equation*}
\max_{\boldsymbol{\rho},\boldsymbol{\mathcal{C}}} G(\boldsymbol{\rho},\boldsymbol{\mathcal{C}}),
\end{equation*}
where
\begin{equation}\label{eq10}
G(\boldsymbol{\rho},\boldsymbol{\mathcal{C}})\triangleq f\big( \boldsymbol{\rho},\boldsymbol{\mathcal{C}}  \big| \big\{ \boldsymbol{u}_j^{(k)},\boldsymbol{y}_j\big\}_{j=1}^M  \big).
\end{equation}
In other words, given the observation $\boldsymbol{y}_j$ and the distortion profiles $\boldsymbol{u}_j^{(k)}$, we are willing to acquire MAP estimates of the image profile value at each pixel along with the binary class that pixel belongs to.

Generally speaking, for arbitrary random variables $\rho$, $y$, $u$ and $\mathcal{C}$ (discrete), using the Bayes' rule we have
\begin{align}\nonumber
f(\rho,\mathcal{C}\big| y,u) &= \frac{f(y\big| \rho,\mathcal{C},u)f(\rho,\mathcal{C},u)}{f(y|u)f(u)}\\&=\frac{f(y\big| \rho,\mathcal{C},u)f(\rho|\mathcal{C},u)}{f(y|u)}\mathbb{P}(\mathcal{C}|u), \label{eq11}
\end{align}
where we used $f(\mathcal{C},u) = \mathbb{P}(\mathcal{C}|u)f(u)$ to derive the second equality. Based on this result we have
\begin{align}\label{eqGMAP}
G(\boldsymbol{\rho},\boldsymbol{\mathcal{C}})\propto  f\Big(\{\boldsymbol{y}_j\}_{j=1}^M \Big|\boldsymbol{\rho},\boldsymbol{\mathcal{C}}\Big) f(\boldsymbol{\rho}|\boldsymbol{\mathcal{C}})\mathbb{P}(\boldsymbol{\mathcal{C}}),
\end{align}
where for the sake of convenience, in all the expressions we dropped the prior knowledge of $\{\boldsymbol{u}^{(k)}_{j}\}_{j=1}^M\vspace{-.12cm}$. Basically, in (\ref{eq10}), the vectors $\{\boldsymbol{u}^{(k)}_{j}\}_{j=1}^M$ are known and deterministic and our intention to derive (\ref{eq11}) is to make the MAP estimation in terms of probability density functions which all make such prior assumption. Also since the denominator term in (\ref{eq11}) is independent of $\rho$ and $\mathcal{C}$, in (\ref{eqGMAP}) we neglected the corresponding term as it will be a constant factor in the maximization.

To further proceed with simplifying the MAP objective, we assume that the pixel values in $\boldsymbol{\rho}$ are independent of each other. Based on this assumption and the i.i.d nature of the noise we can state that
\begin{equation}\label{eqpoint}
G(\boldsymbol{\rho},\boldsymbol{\mathcal{C}})\propto\prod_{i=1}^Pf(\rho_i|\mathcal{C}_i)\mathbb{P}(\mathcal{C}_i)\prod_{j=1}^Mf(y_{j,i}|\rho_i,\mathcal{C}_i),
\end{equation}
where $y_{j,i}$ is the $i$-th element of $\boldsymbol{y}_j$. The point-wise distributions appeared in (\ref{eqpoint}) can now be modeled based on the problem setup.  

Clearly, based on (\ref{eq1}), $f(y_{j,i}|\rho_i,\mathcal{C}_i)\sim \mathcal{N}(\rho_i u_{j,i}^{(k)},\sigma^2)$, which is an immediate result of the normal noise model. We also assume that 
\begin{equation*}
\mathbb{P}(\mathcal{C}_i)=p_{\mathcal{C}_i}=\left\{\begin{array}{lc}p_0& if\;\;\mathcal{C}_i=0\\ p_1 & if\;\;\mathcal{C}_i=1 \end{array}\right ., 
\end{equation*}
where $p_0$ and $p_1=1-p_0$ are somehow the prior estimates of the portion of pixels belonging to each class (e.g., knowing that roughly 20\% of the image pixels correspond to the anomaly). If no such information is available a priori, a reasonable assumption is $p_0=p_1=0.5$.

\begin{figure}[t]
\centering
\includegraphics[height=1.7in]{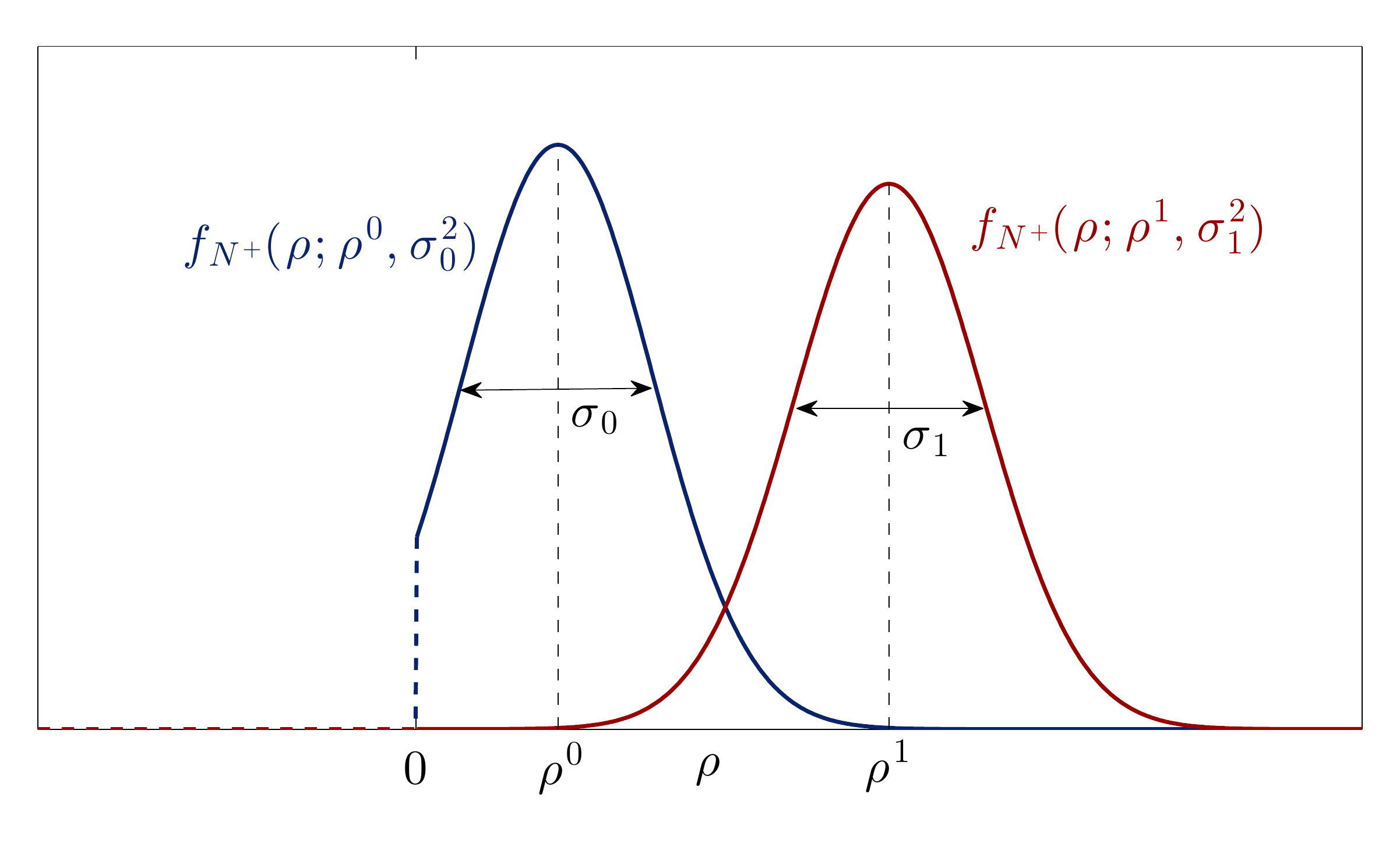}\vspace{-.3cm}
\caption{The distributions for $f(\rho|\mathcal{C}=0)$ and $f(\rho|\mathcal{C}=1)$ as truncated normal distributions}
\label{fig2}
\end{figure}

Finally, with reference to the term $f(\rho_i|\mathcal{C}_i)$, knowing that a pixel belongs to class 0 (or 1) we know that its value is likely to be close to $\rho^0$ (or $\rho^1$). In practice $f(\rho_i|\mathcal{C}_i)$ should model the uncertainty on how the values of each class concentrate around the mean class value. While we can use different distributions to model such concentration, for simplicity and in order to obtain closed form expressions, we assume that $f(\rho_i|\mathcal{C}_i=0)$ and $f(\rho_i|\mathcal{C}_i=1)$ are in the form of \emph{truncated normal distributions}, taking positive arguments and mainly concentrating around $\rho^0$ and $\rho^1$, respectively. By definition, a random variable $\rho$ truncated to values greater than zero takes the following pdf: 
\begin{equation}\label{tnd}
f_{N^+}(\rho;\tilde\mu,\tilde\sigma) = \left \{\begin{array}{lc} \frac{\gamma}{\tilde \sigma}\exp\big(-\frac{(\rho-\tilde\mu)^2}{2{\tilde\sigma}^2}\big)& \rho\geq 0\\ 0 & \rho< 0 \end{array}\right . .
\end{equation}
Here, $\tilde\mu$ and ${\tilde\sigma}^2$ are roughly the mean and variance and $\gamma$ is a normalizing factor to assure the pdf integrates to 1. To model an almost binary nature of the pixel values, we assume $f(\rho_i|\mathcal{C}_i=0) = f_{N^+}(\rho_i;\rho^0,\sigma_0^2)$ and $f(\rho_i|\mathcal{C}_i=1) = f_{N^+}(\rho_i;\rho^1,\sigma_1^2)$, or more concisely
\begin{equation*} 
f(\rho_i|\mathcal{C}_i) = f_{N^+}(\rho_i;\rho^{\mathcal{C}_i},\sigma_{{\mathcal{C}_i}}^2).
\end{equation*}
Figure \ref{fig2} shows the underlying truncated distributions. The positivity assumption in (\ref{tnd}) is imposed by the physics of the problem, where the reflection coefficient of the slab is always considered to be a positive quantity. 

Having the constituting terms modeled in (\ref{eqpoint}), we can proceed with the pixel value and class label MAP estimation. To maximize $G(\boldsymbol{\rho},\boldsymbol{\mathcal{C}})$ we may minimize the negative log function
\[g(\boldsymbol{\rho},\boldsymbol{\mathcal{C}}) = -\log G(\boldsymbol{\rho},\boldsymbol{\mathcal{C}}).
\]
Based on the proposed distribution models, one can easily verify that
\begin{align*}
g(\boldsymbol{\rho},\boldsymbol{\mathcal{C}})= \left\{\begin{array}{lc} K + \sum_{i=1}^P g(\rho_i,\mathcal{C}_i)& \rho_i\geq 0\\ +\infty& \rho_i< 0 \end{array}\right ., 
\end{align*}
where
\begin{align}\label{gci}
g(\rho_i,\mathcal{C}_i) \triangleq  \log\frac{\sigma_{{\mathcal{C}_i}}}{p_{\mathcal{C}_i}} +  \frac{(\rho_i-\rho^{\mathcal{C}_i})^2}{2\sigma_{{\mathcal{C}_i}}^2}+\sum_{j=1}^M\frac{(y_{j,i}-\rho_i u_{j,i}^{(k)})^2}{2\sigma^2},
\end{align}
and $K$ is a constant in terms of $\sigma$, $\gamma_i$ and the constant factor in (\ref{eqGMAP}). The minimization objective $g(\boldsymbol{\rho},\boldsymbol{\mathcal{C}})$ is separable in terms of the cost for each pixel and we can minimize each term individually. Also minimizing $g(\rho_i,\mathcal{C}_i)$ can be performed in a serial manner by first minimizing with respect to $\rho_i$ and then with respect to the class label $\mathcal{C}_i$. More concretely,
\begin{align}\nonumber
\min_{\boldsymbol{\mathcal{C}},\boldsymbol{\rho}\succeq \boldsymbol{0}} g(\boldsymbol{\rho},\boldsymbol{\mathcal{C}}) &= K + \sum_{i=1}^P \min_{\mathcal{C}_i,\rho_i\geq 0} g(\rho_i,\mathcal{C}_i)\\ & = K + \sum_{i=1}^P \min_{\mathcal{C}_i}\min_{\rho_i\geq 0} g(\rho_i,\mathcal{C}_i).
\end{align}
For a fixed given label $\mathcal{C}_i=\mathcal{C}$, we can minimize  $g(\rho_i,\mathcal{C})$ with respect to $\rho_i$ by setting $\partial g(\rho_i,\mathcal{C})/\partial \rho_i$ to zero. Based on the formulation in (\ref{gci}) a simple calculation yields 
\begin{align}\label{eqsol}
 \argmin_{\rho_i} g(\rho_i,\mathcal{C}) = \frac{\sigma^2\rho^{\mathcal{C}}+\sigma_{{\mathcal{C}}}^2\sum_{j=1}^M y_{j,i}u_{j,i}^{(k)}}{\sigma^2+\sigma_{{\mathcal{C}}}^2\sum_{j=1}^M{u_{j,i}^{(k)}}^2}.
\end{align}
The function $g(\rho_i,\mathcal{C})$ is convex in terms of $\rho_i$, and the solution to the constrained problem ($\rho_i\geq 0$) is the expression in (\ref{eqsol}) if feasible, or 0 otherwise. Basically,
\begin{align}\nonumber
\rho^*_i &\triangleq \argmin_{\rho_i\geq 0 } g(\rho_i,\mathcal{C})\\&= \max\left(\frac{\sigma^2\rho^{\mathcal{C}}+\sigma_{{\mathcal{C}}}^2\sum_{j=1}^M y_{j,i}u_{j,i}^{(k)}}{\sigma^2+\sigma_{{\mathcal{C}}}^2\sum_{j=1}^M{u_{j,i}^{(k)}}^2},0\right).
\end{align}
We are now only left with minimizing $g(\rho_i^*,\mathcal{C}_i)$ with respect to $\mathcal{C}_i$. Since the class labels $\mathcal{C}_i$ only takes binary values, we can conveniently find the corresponding minimizer through the following binary comparison:
\begin{align}\nonumber 
\mathcal{C}^*_i \triangleq \argmin_{\mathcal{C}_i\in\{0,1\}} g(\rho_i^*,\mathcal{C}_i)= \left\{\begin{array}{lc}1& if\;\;g(\rho^*_i,1)<g(\rho^*_i,0)\\ 0&if\;\;g(\rho^*_i,1)\geq g(\rho^*_i,0)  \end{array}\right ..
\end{align}
Equation (\ref{eqsol}) in some sense generates the optimal solution $\rho_i$, but leaves us with an ambiguity about the class label. The label can be obtained based on the above comparison, to find the ultimate MAP estimator for the image profile. Algorithm \ref{alg2} summarizes the steps sketched in this section, and presents the overall process to perform the demodulation task. 

\begin{algorithm}
\caption{Decoupling algorithm}\label{alg2}
\begin{algorithmic}[1]
\State $\mbox{\textbf{input}}$ $\{\boldsymbol{y}_j\}_{j=1}^M$, $\{N_j\}_{j=1}^M$, $\rho^0$, $\rho^1$, $\sigma_0^2$, $\sigma_1^2$, $\sigma^2$
\State $\mbox{\textbf{calculate}}$ $\{\boldsymbol{S}^j\}_{j=1}^M$ from $\{\boldsymbol{y}_j\}_{j=1}^M$ \Comment{\textcolor{gray}{As noted in Section \ref{sec:subspace}}}
\State $\boldsymbol{\rho}\gets \boldsymbol{1}$
\Repeat
\For{$j=1,\cdots,M$} 
\State $\boldsymbol{\alpha}_j \gets \big(\mbox{diag}(\boldsymbol{\rho})\boldsymbol{S^j} \big)^\dagger\boldsymbol{y}_j$
\State $\boldsymbol{u}_j \gets \boldsymbol{S}^j\boldsymbol{\alpha}_j$
\EndFor
\For{$i=1,\cdots,P$} 
\State $w_0 \gets \max\big(\frac{\sigma^2\rho^{0}+\sigma_{{0}}^2\sum_{j=1}^M y_{j,i}u_{j,i}}{\sigma^2+\sigma_{{0}}^2\sum_{j=1}^Mu_{j,i}^2},0\big)$ 
\State $w_1 \gets \max\big(\frac{\sigma^2\rho^{1}+\sigma_{{1}}^2\sum_{j=1}^M y_{j,i}u_{j,i}}{\sigma^2+\sigma_{{1}}^2\sum_{j=1}^Mu_{j,i}^2},0\big)$
\If {$g(w_0,0)\leq g(w_1,1)$}\Comment{\textcolor{gray}{$g(.,.)$ is defined in (\ref{gci})}}
    \State $\rho_i\gets w_0$
\Else
\State $\rho_i\gets w_1$
\EndIf 
\EndFor
\Until{convergence}
\State \textbf{return} $\boldsymbol{\rho}$, $\{\boldsymbol{u}_j\}_{j=1}^M$
\end{algorithmic}
\end{algorithm}

While the parameters $\sigma_0$ and $\sigma_1$ have statistical interpretations, from an optimization standpoint they can be considered as free parameters controlling the algorithm's performance. When $\sigma_0$ and $\sigma_1$ are set to be small quantities, the algorithm converges faster at the expense of more sensitivity to the initialization (possibility of recovering a local minimizer). Assigning relatively larger values to these quantities paves the path towards identifying the global minimizer in more number of iterations. This is also a more reliable parameter selection in the case of noisy observations.

\section{Simulation and Experiments}\label{sec:exp}
In this section we assess the performance of the proposed technique in demodulating simulated and real data. The simulation results mainly highlight the performance of the method for various number of frames and noise levels in the data. The second set of experiments demonstrate the method's success in removing multiplicative sweep distortion profiles from actual THz measurements. 

\subsection{Simulated Data}

\begin{figure}[!htb]
\begin{tabular}{ccc}
&
\hspace{-.43cm}\begin{overpic}[trim={4cm 7.5cm 4cm 7cm}, clip,width=1.05in,height=1in]{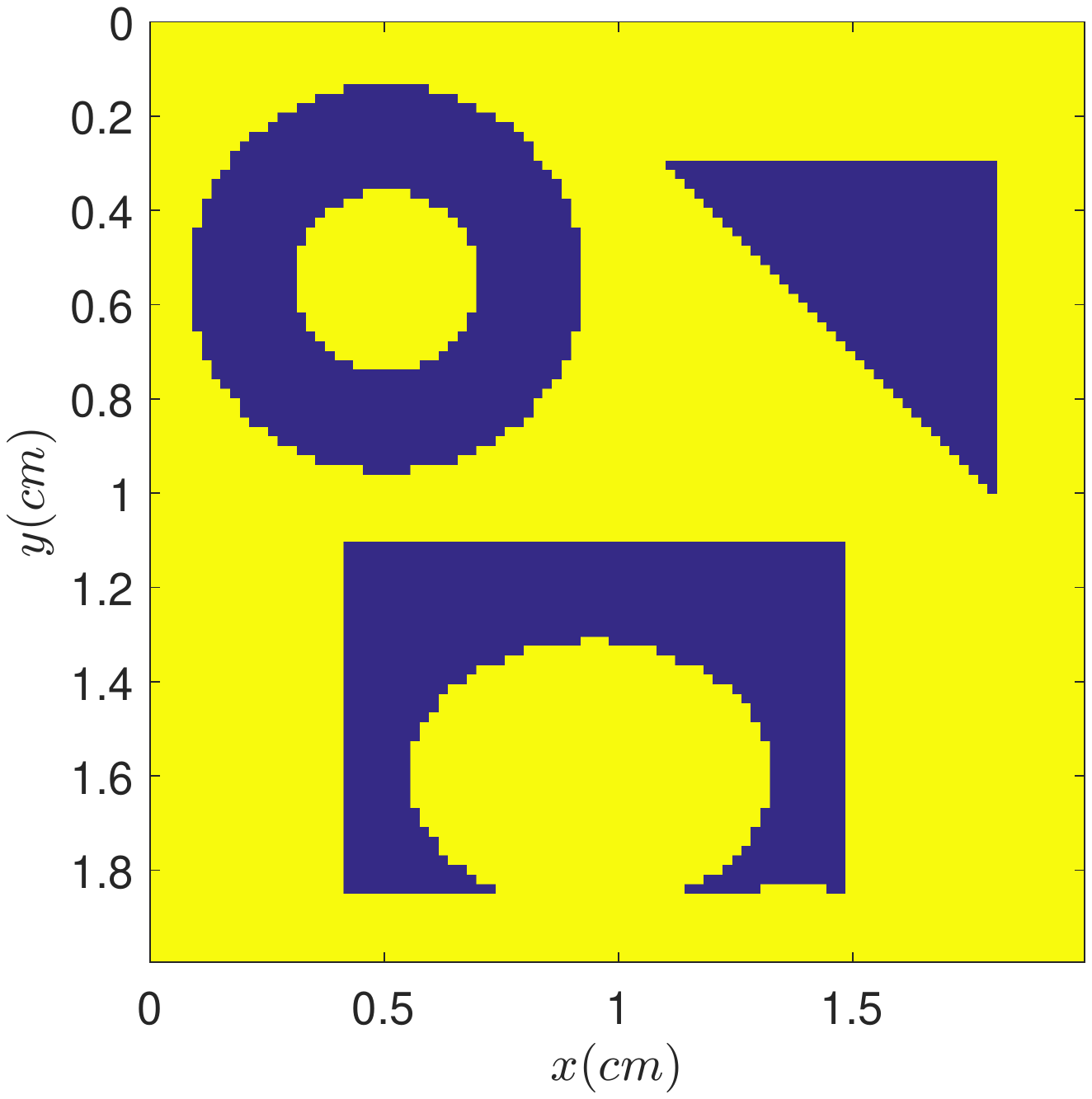}\put (48,-11) {(a)}\end{overpic}
&\\[.18cm]
\hspace{-.2cm}\begin{overpic}[trim={0.2cm 0.5cm 0.5cm 0.2cm}, clip,width=1.1in,height=1.05in]{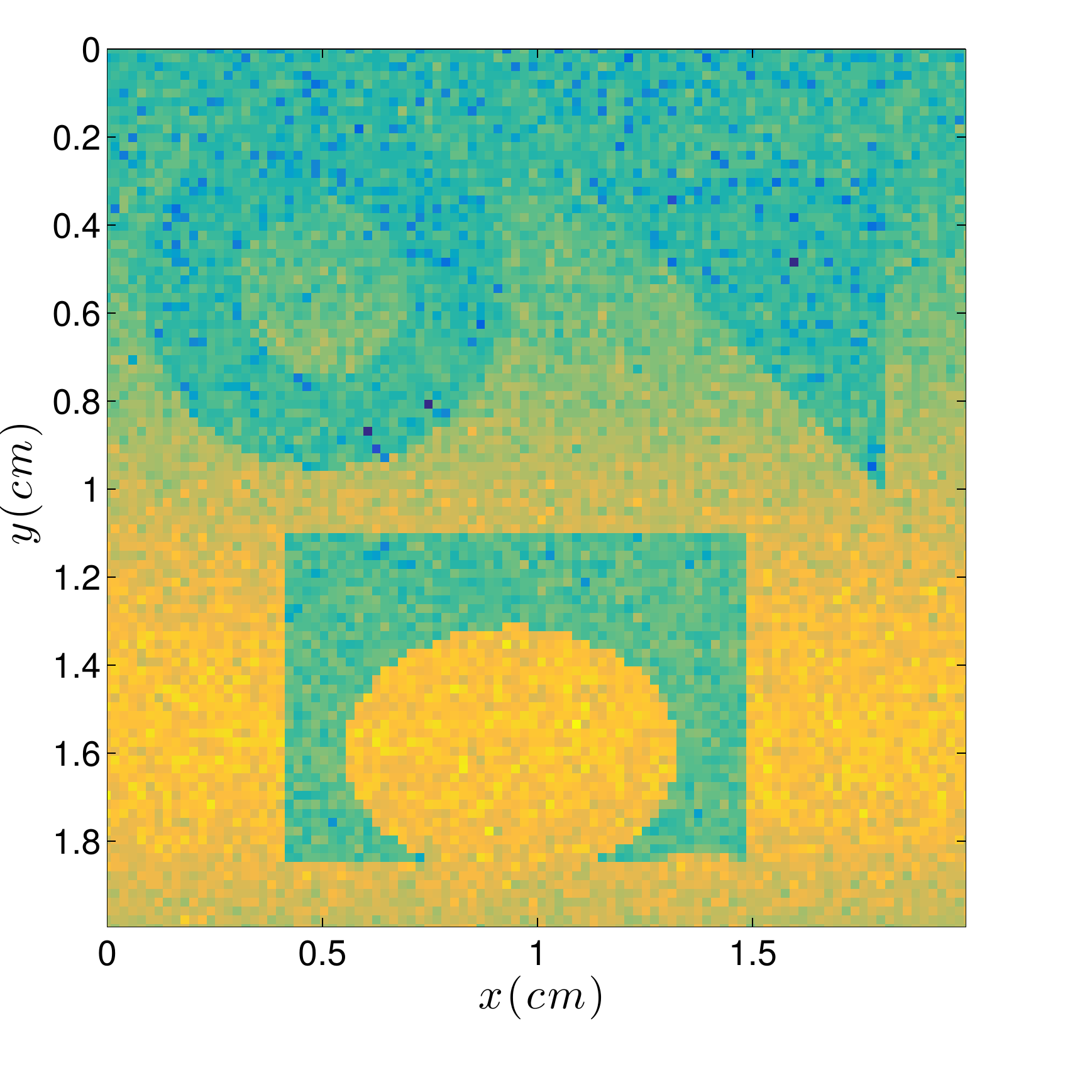}\put (45,-8) {(b.1)} \end{overpic}
&
\hspace{-.23cm}\begin{overpic}[trim={0.2cm 0.5cm 0.5cm 0.2cm}, clip,width=1.1in,height=1.05in]{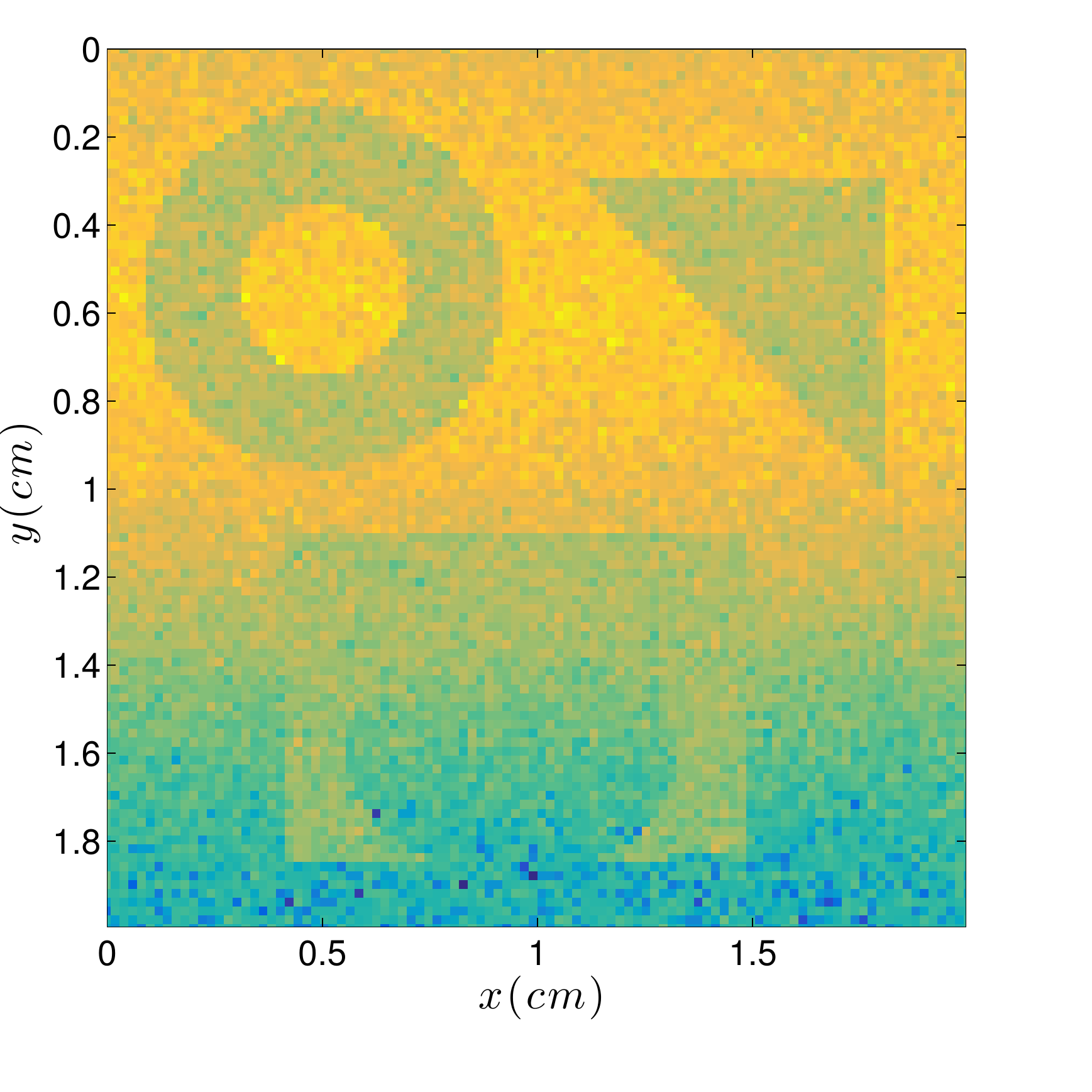}\put (45,-8) {(b.2)}\end{overpic}
&
\hspace{-.23cm}\begin{overpic}[trim={0.2cm 0.5cm 0.5cm 0.2cm}, clip,width=1.1in,height=1.05in]{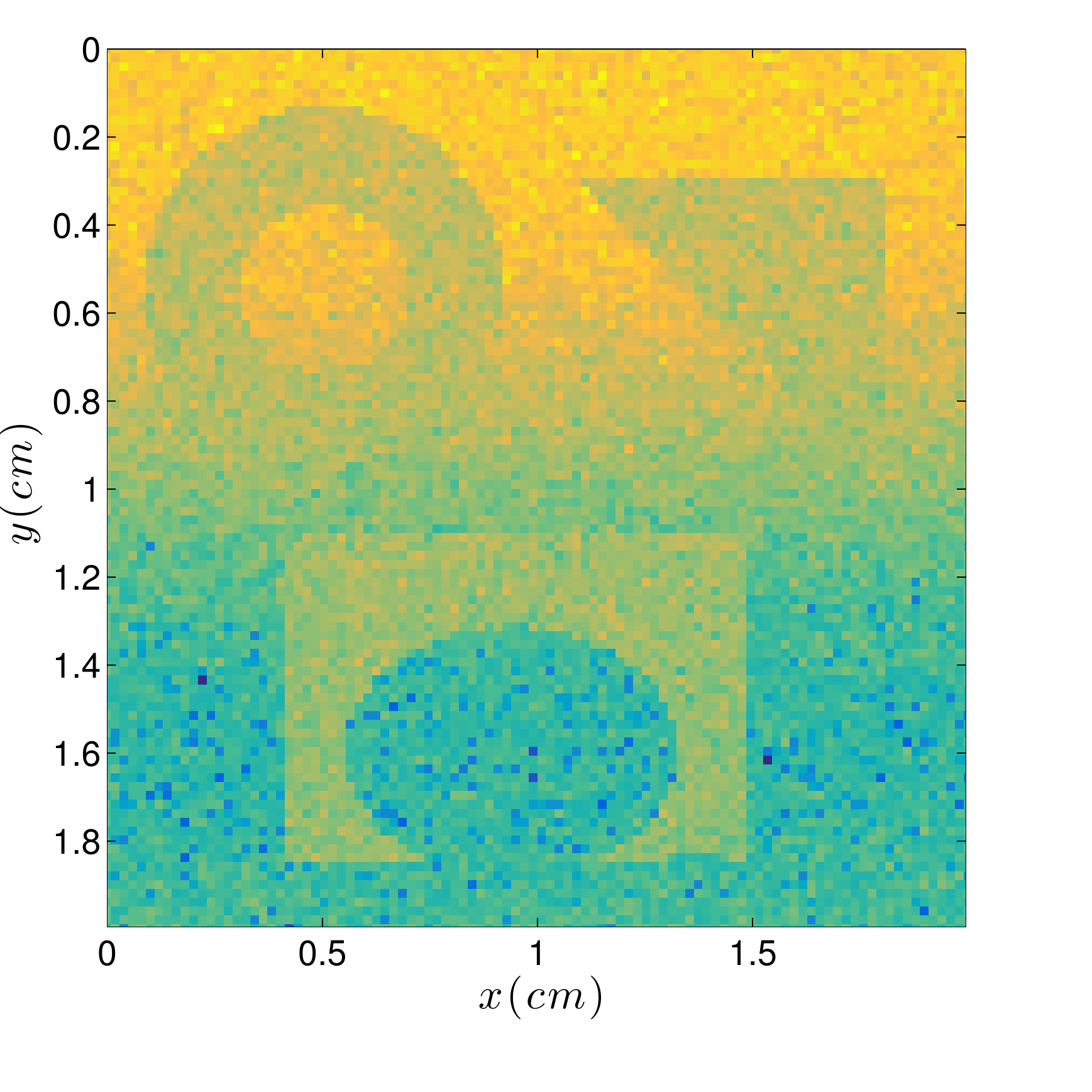}\put (45,-8) {(b.3)}
\end{overpic}
\end{tabular}
\\[.1cm]
\begin{tabular}{cc}
\hspace{-.3cm}\begin{overpic}[trim={0.2cm 0.5cm 0.5cm 0.2cm}, clip,width=1.75in,height=1.65in]{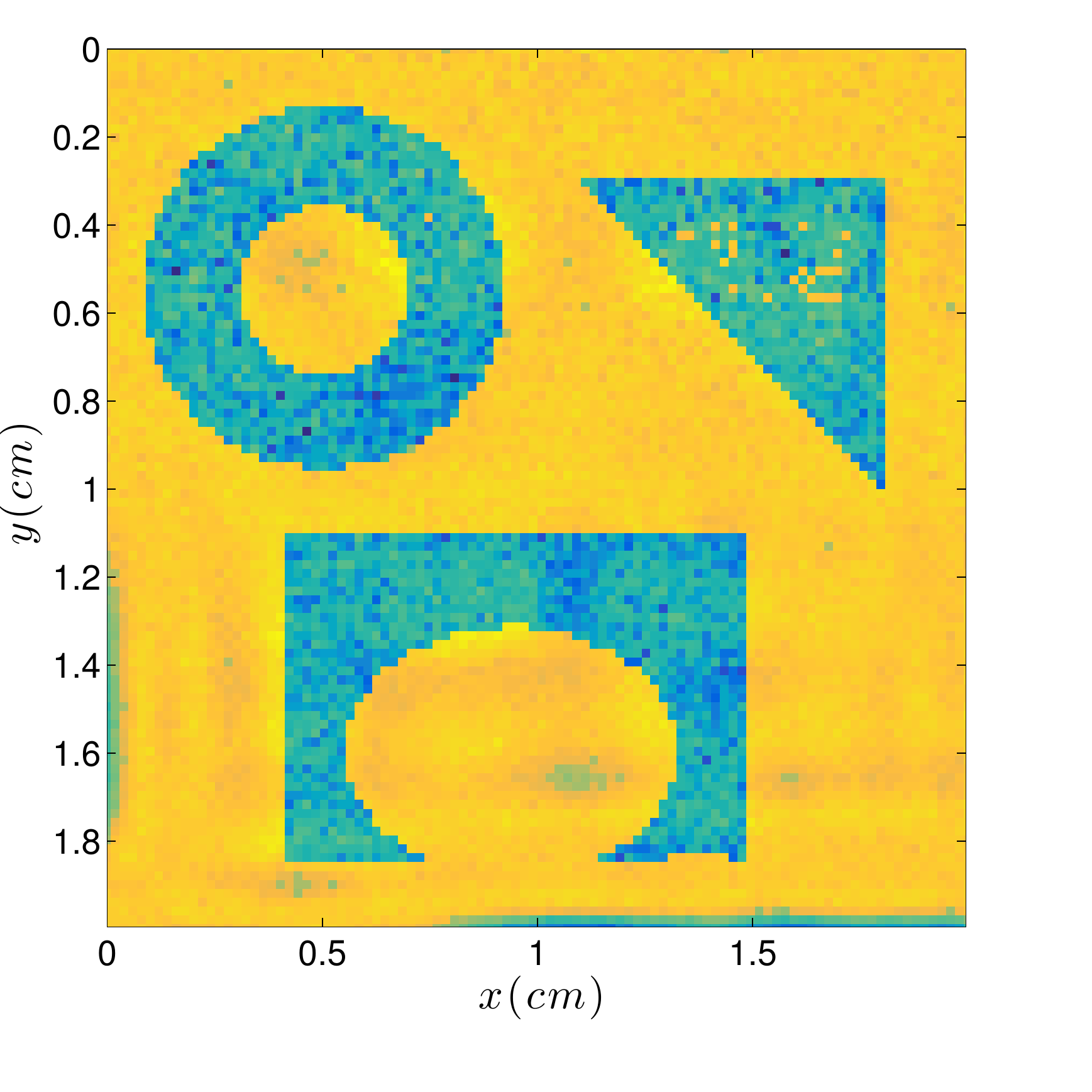}\put (45,-3) {(c.1)} \end{overpic}
&
\hspace{-.3cm}\begin{overpic}[trim={0.2cm 0.5cm 0.5cm 0.2cm}, clip,width=1.75in,height=1.65in]{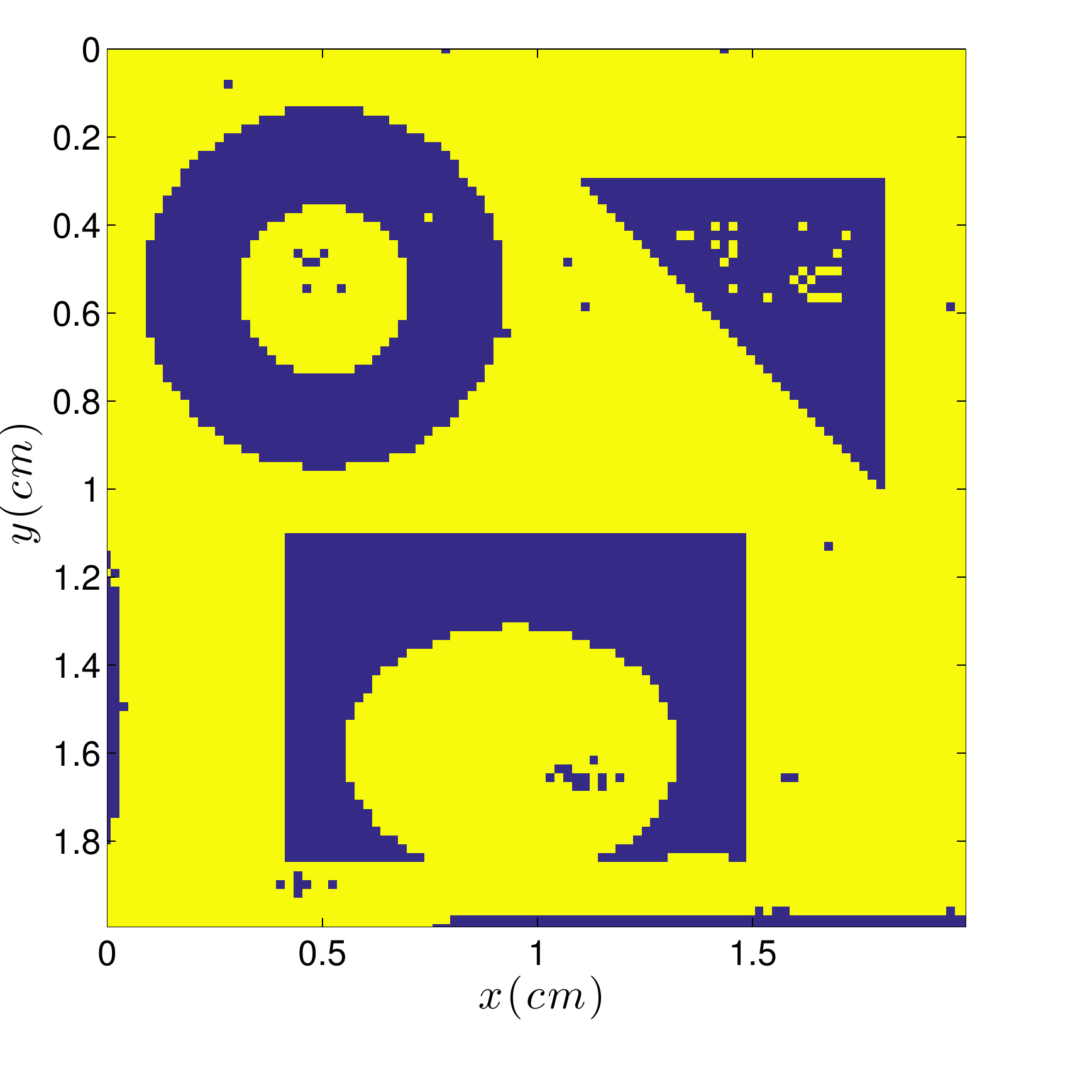}\put (45,-3) {(c.2)}
\end{overpic}
\\[.05cm]
\hspace{-.56cm}\begin{overpic}[trim={4cm 7cm 4cm 7cm}, clip,width=1.7in,height=1.65in]{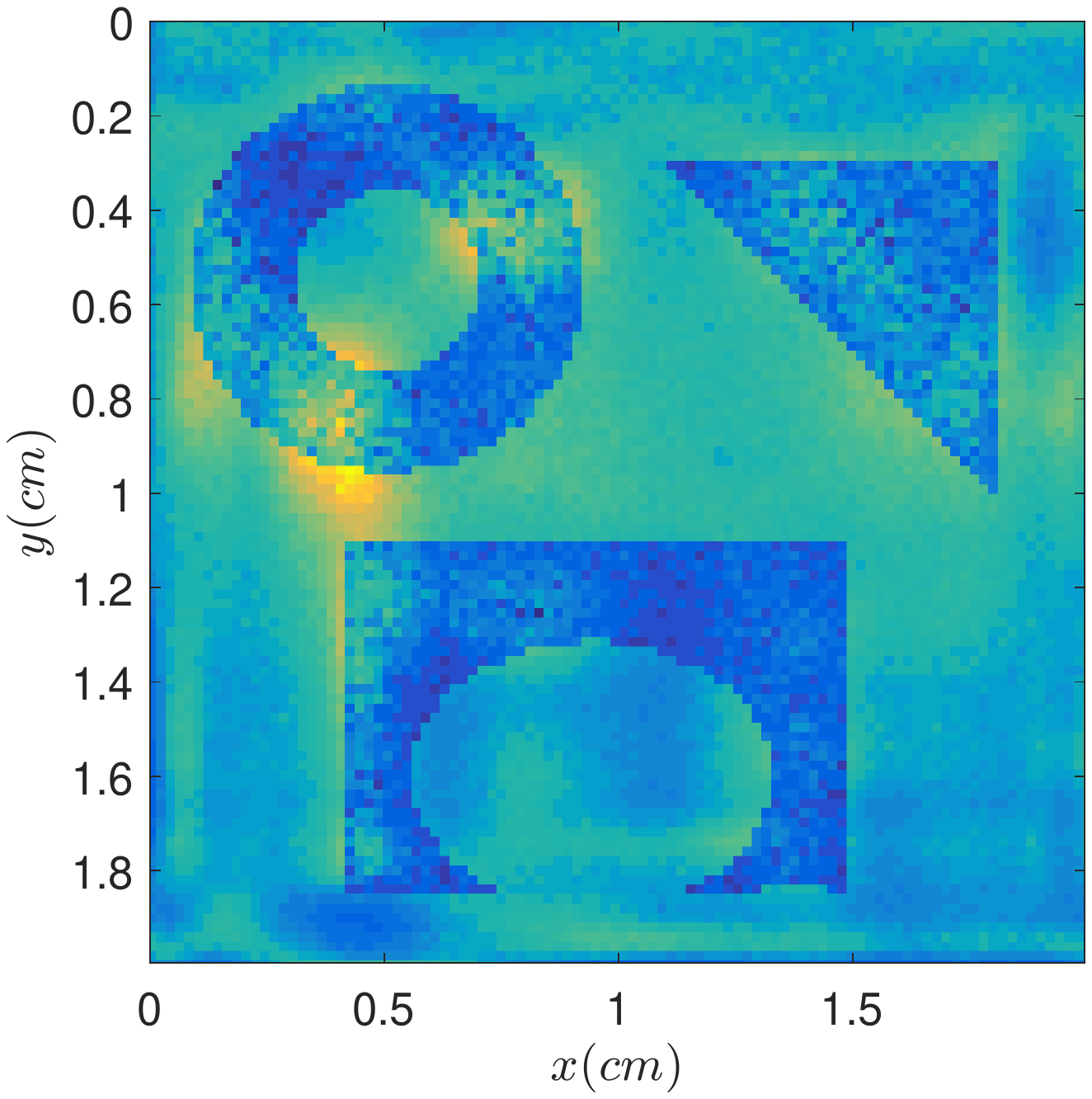}\put (48,-6) {(d.1)} \end{overpic}
&
\hspace{-.56cm}\begin{overpic}[trim={4cm 7cm 4cm 7cm}, clip,width=1.68in,height=1.65in]{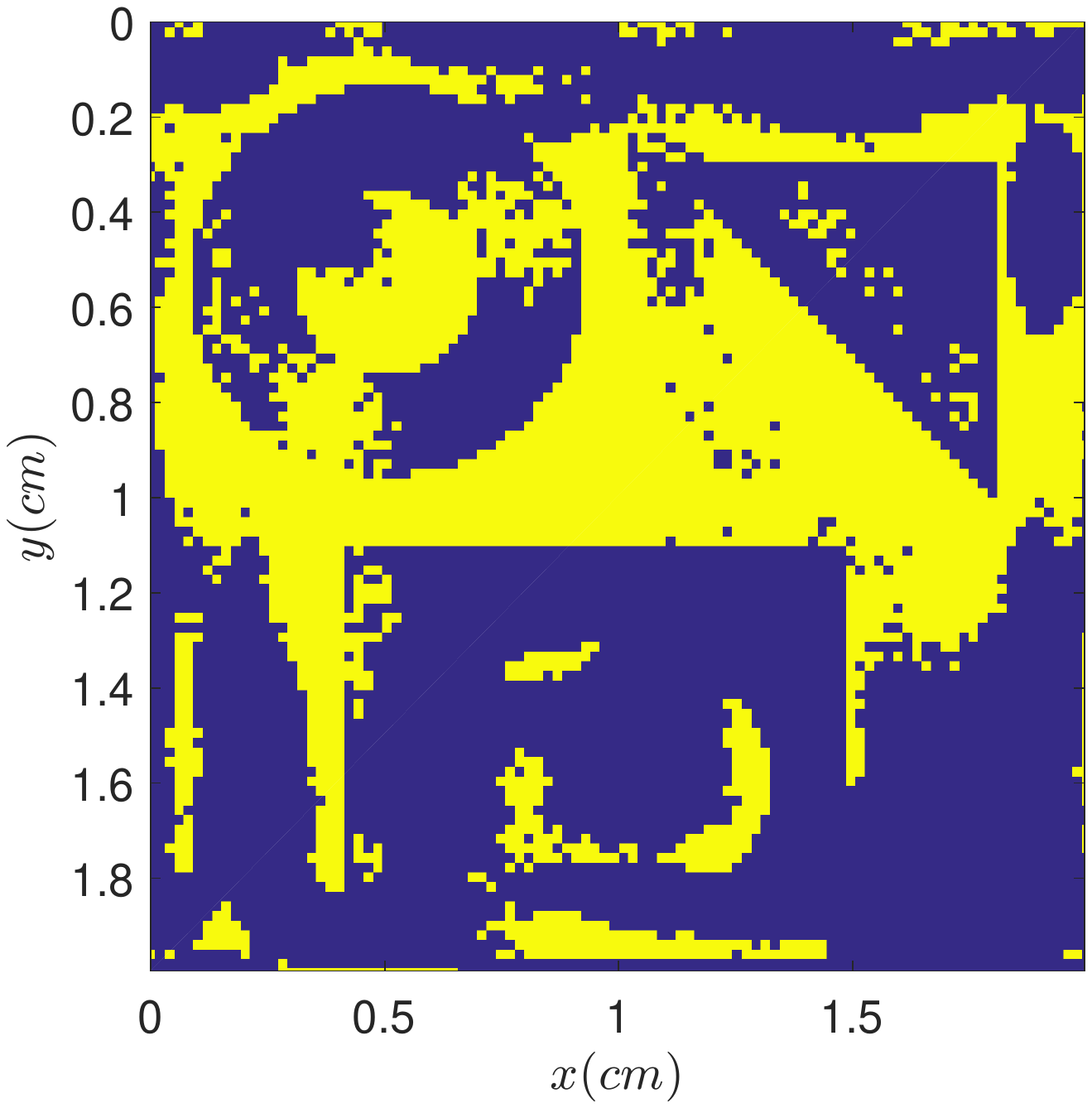}\put (48,-6) {(d.2)}
\end{overpic}
\end{tabular}
\caption{Demodulation of simulated data; (a) reference profile of $\rho(\boldsymbol{x})$; (b.1-3) observed reflections modulated by horizontal sweep profiles; (c.1) the demodulated image using the proposed algorithm using $\sigma_0=\sigma_1=10^{-5}$; (c.2) a binary approximation of the result in (c.1); (d.1) the recovered image using the nuclear norm minimization; (d.2) a binary approximation of the result in (d.1) }\label{fig3}
\end{figure}

We simulate the reflected signal from a dielectric slab with the $x$-$y$ profile depicted in Figure \ref{fig3}(a). The dielectric width is $d=100$ $\mu m$. For the binary dielectric slab the reflection coefficients are $\rho^0=0.3$ and $\rho^1 = 0.1$. The sample is probed with a bipolar THz pulse (simply derivative of a Gaussian) as
\[\chi(t) = (t_0 - t)\exp\left(-\frac{(t-t_0)^2}{2T^2}\right)\quad t\geq 0,
\]
where $t_0 = 1$ $ps$ and $T = t_0/4$. To model a sweep modulated signal, we assume that instead of point-wise observations at a constant $z=z_0$, the measurements are performed at $z = z_0 + \alpha_1 x + \alpha_2 y$, where $(x,y)$ is the planar coordinate, and $\alpha_1 = 10^{-6}$, $\alpha_2 = 10^{-4}$ are small constants. We take $M=10$ uniform samples of the reflected wave in a time period of $0.8$ $ps$. Among the recorded frames Figure \ref{fig3}(b) shows three sample images, where an almost vertically moving sweep distortion across the frames is observable. The samples are synthetically corrupted with white noise ($\mbox{SNR}=10$ $dB$).  

In recovering the binary profile, for the subspace construction step associated with this example we use $N_1=\cdots=N_{10} = 100$ most dominant wavelet basis. Consistently through this example and the remaining experiments we use the symlet wavelet family.

Figure \ref{fig3}(c.1) reports the reconstructed reflection profile $\rho$, using our proposed scheme. The algorithm converges in less than 20 iterations, and as may be observed, the result is no more in hold of the distortion traces. A simple rounding of the result to the closest binary value (between $\rho^0$ and $\rho^1$) yields Figure \ref{fig3}(c.2), which is reasonably close to the original profile depicted in Figure \ref{fig3}(a).

Figure \ref{fig3}(d.1) reports the reconstruction using the nuclear norm framework proposed in Section \ref{lowrank}. While some level of distinction between the two binary phases is achieved, the fact that our proposed scheme efficiently uses the binary prior helps us outperform the nuclear norm reconstruction. A similar binary approximation of the nuclear norm reconstruction is demonstrated in Figure \ref{fig3}(d.2), which clearly fails to characterize some of the main details in the original binary profile. 

To more extensively analyze the performance of our algorithm, we proceed by the sensitivity analysis of the main parameters affecting the reconstruction. For this purpose, two main scenarios are considered. First, an ideal case where the exact sweep distortion subspace is known a priori. From a theoretical standpoint, one way to characterize a subspace for the sweep distortion profiles is to run exactly the same experiment with a homogeneous slab and take the observations as the subspace basis. While experimentally such accurate characterization of the subspace is hard, here we present it as a hypothetical case for comparison purposes and to compare the results against an ideal setup. 

For the second class of experiments, we use the proposed wavelet thresholding framework to determine the distortion subspaces. For each proposed scenario we test the algorithm for various values of $M$ (number of available frames) and the SNR in the observations. Both the reconstructed $\rho$ as well as a binary rounded version are compared with the reference image depicted in Figure \ref{fig3}(a) and the mean squared error (MSE) is reported.   

\begin{figure}[t]
\centering
\includegraphics[width=3in,height=2.4in]{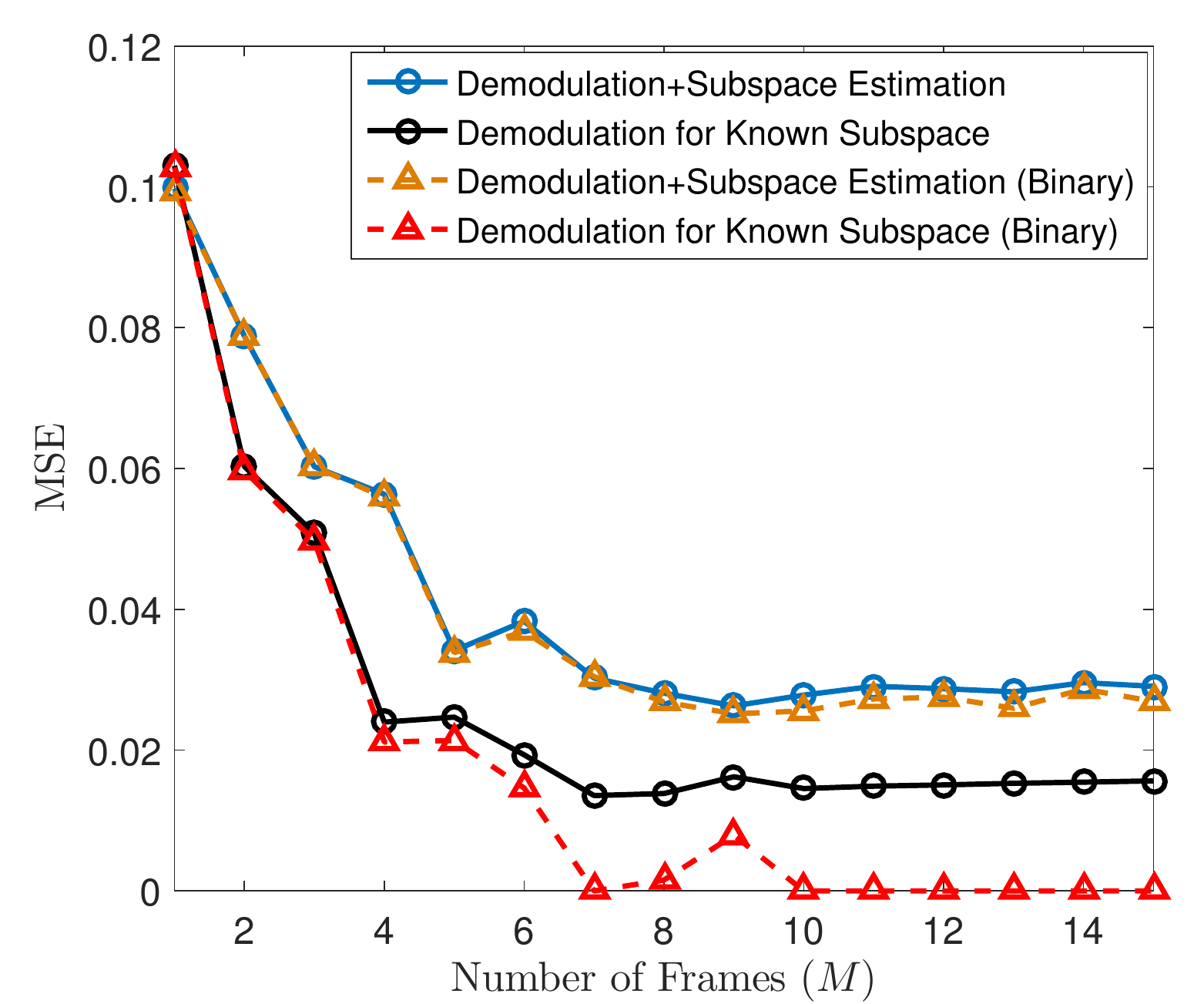}
\caption{MSE as a function of available number of frames, $M$, for various scenarios: a total of 20 frames using uniform sampling in time is made available through the simulation. For each MSE report $M$ frames are randomly selected from the 20 frames and passed to the algorithm. The process is performed multiple times for each $M$ and the average MSE is reported}
\label{fig4}
\end{figure}

Figure \ref{fig4} shows the MSE values for various number of available frames and a fixed SNR value of 10 $dB$. The dark solid curve corresponds to the case of exact subspace characterization and the dashed curve standing close to it is the MSE value for the reconstructed $\rho$ rounded to the closest binary value. We can see that for only 7 frames (or more), a perfect or close to perfect recovery of the binary profile is possible. For fewer number of available frames, the MSE still remains at a controlled level. 

The lighter solid line (and the dashed line standing close to it) report the MSE values of similar experiments when the sweep distortion subspaces are characterized through the wavelet thresholding. We can see that aside from a slight performance gap, the MSE values follow a similar reduction pattern as the ideal case. Based on the reported MSE values, the recovery is still close to the reference image. The slight performance gap between this case and the ideal case is the result of error in exact characterization of the sweep distortion subspaces.

\begin{figure}[t]
\centering
\includegraphics[width=3in,height=2.4in]{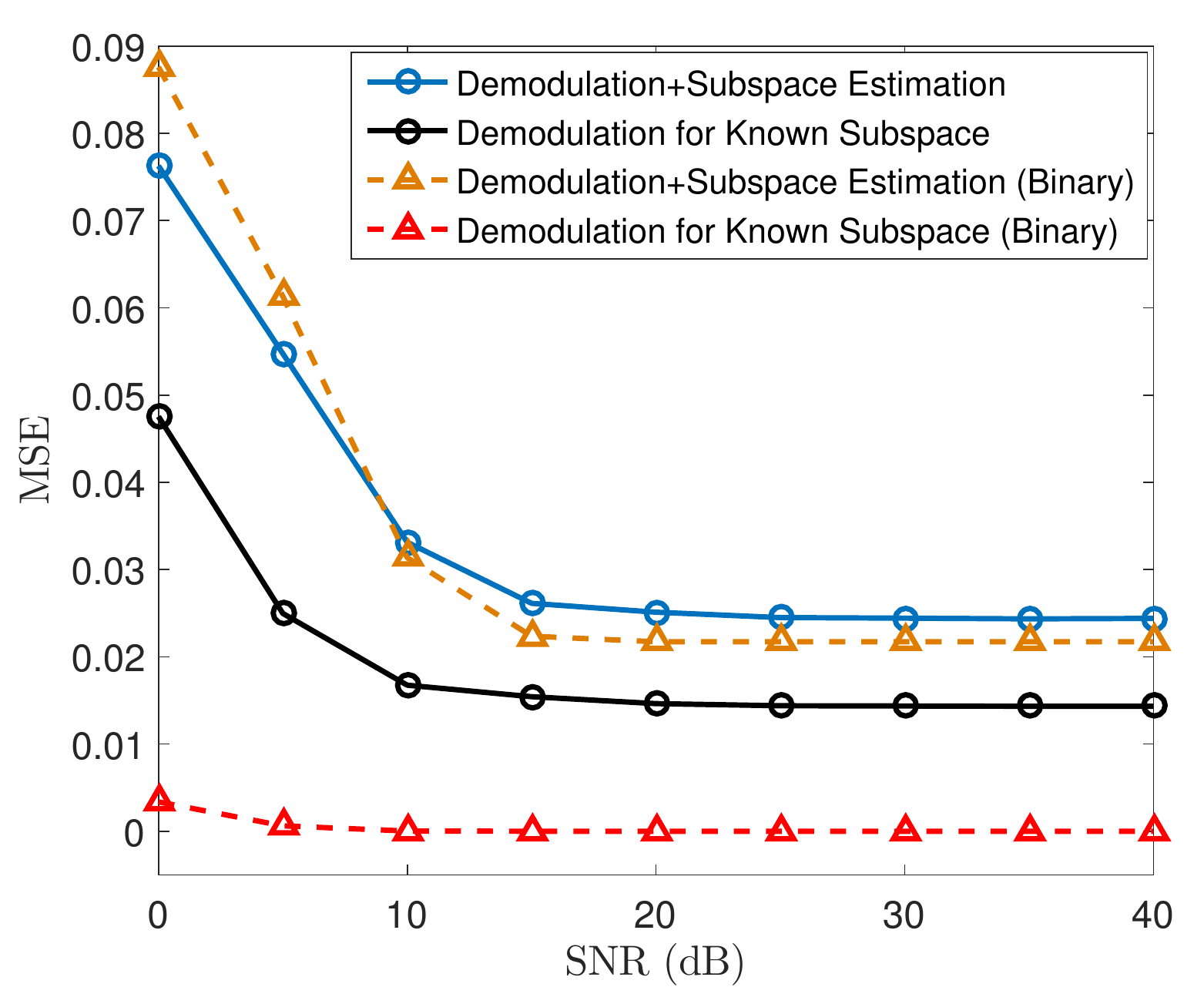}
\caption{MSE as a function of SNR for a fixed number of available frames ($M=10$)}
\label{fig5}
\end{figure}

Figure \ref{fig5} shows the MSE values for the case of fixed number of frames ($M=10$) and varying observation noise. For this experiment, in the case of known distortion subspace, a perfect recovery of the sweep distortion profile is possible for SNR values approximately exceeding 10 $dB$ (by rounding to the closest binary value). In the case of characterizing the sweep distortion subspace through wavelet thresholding, the MSE values follow a similar trend as the case of knowing the distortion subspace a priori. The only difference is a slight performance gap which is again linked to the error in exact characterization of the sweep distortion subspaces.

\subsection{Experimental Data}
\begin{figure}[t]
\begin{tabular}{ccc}
\hspace{-.2cm}\begin{overpic}[trim={0.2cm 0.5cm 0.5cm 0.2cm}, clip,width=1.1in,height=.7in]{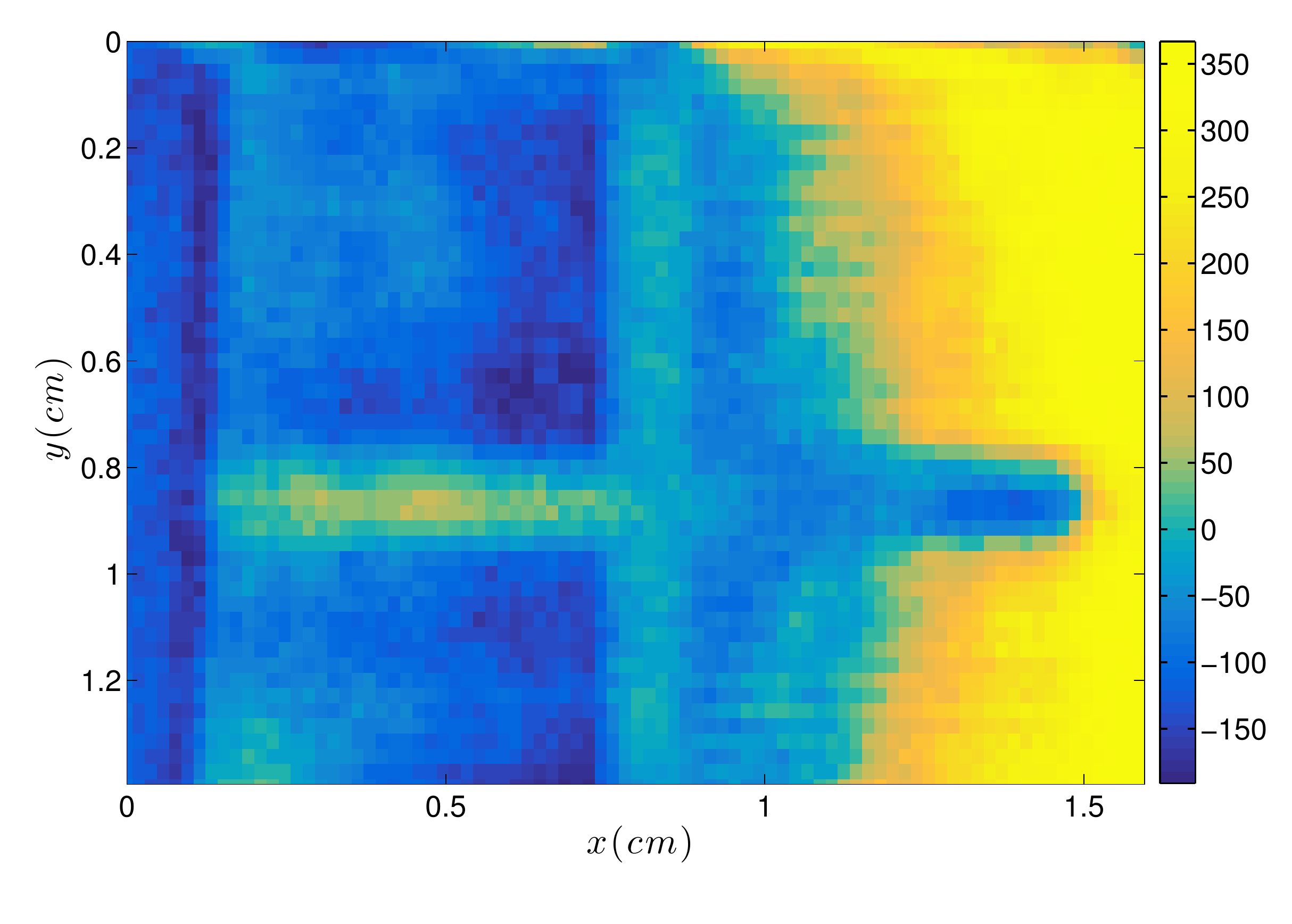}\put (45,-10) {(a.1)} \end{overpic}
&
\hspace{-.9cm}\begin{overpic}[trim={0.2cm 0.5cm 0.5cm 0.2cm}, clip,width=1.1in,height=.7in]{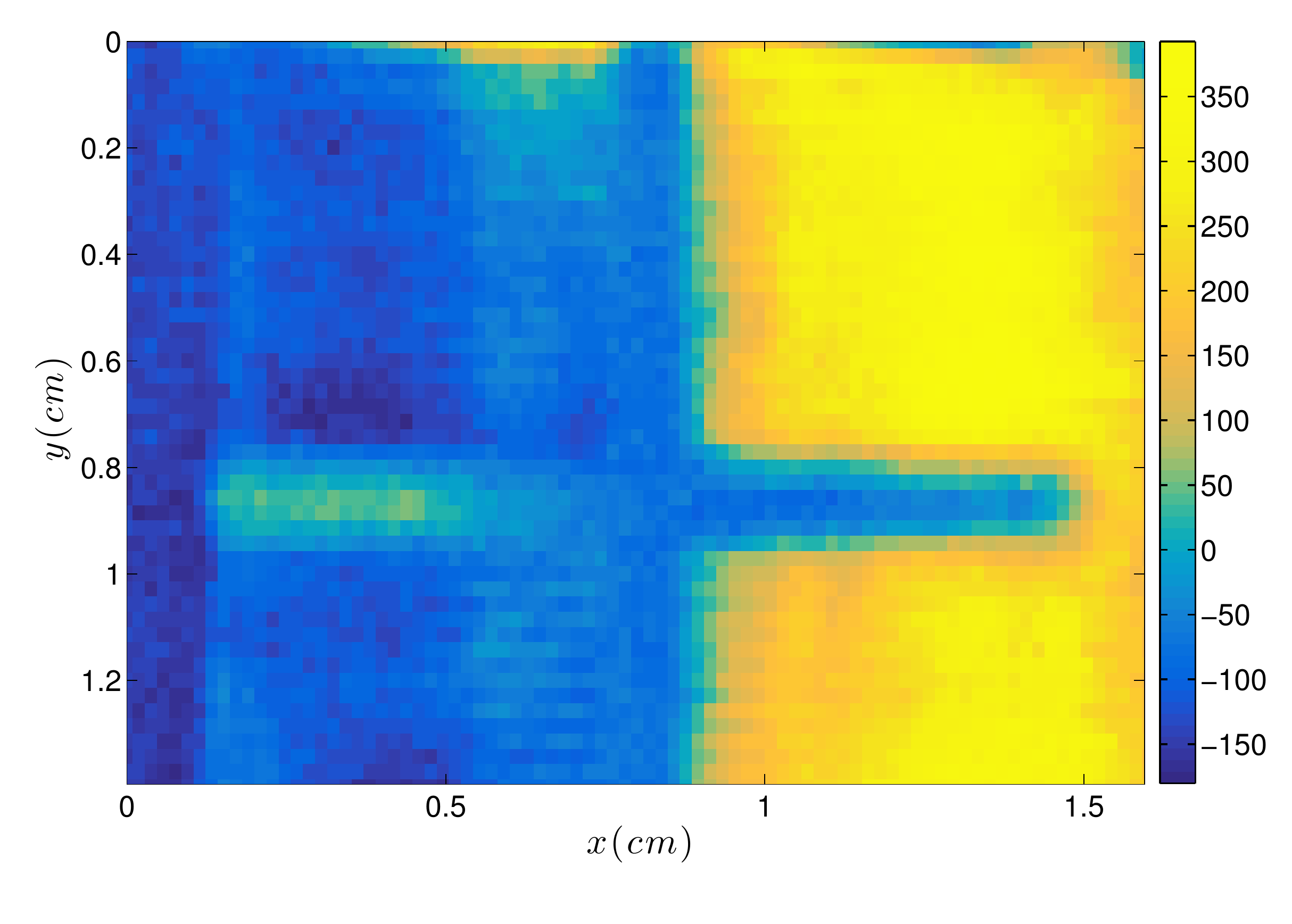}\put (45,-10) {(a.2)}\end{overpic}
&
\hspace{-.9cm}\begin{overpic}[trim={0.2cm 0.5cm 0.5cm 0.2cm}, clip,width=1.1in,height=.7in]{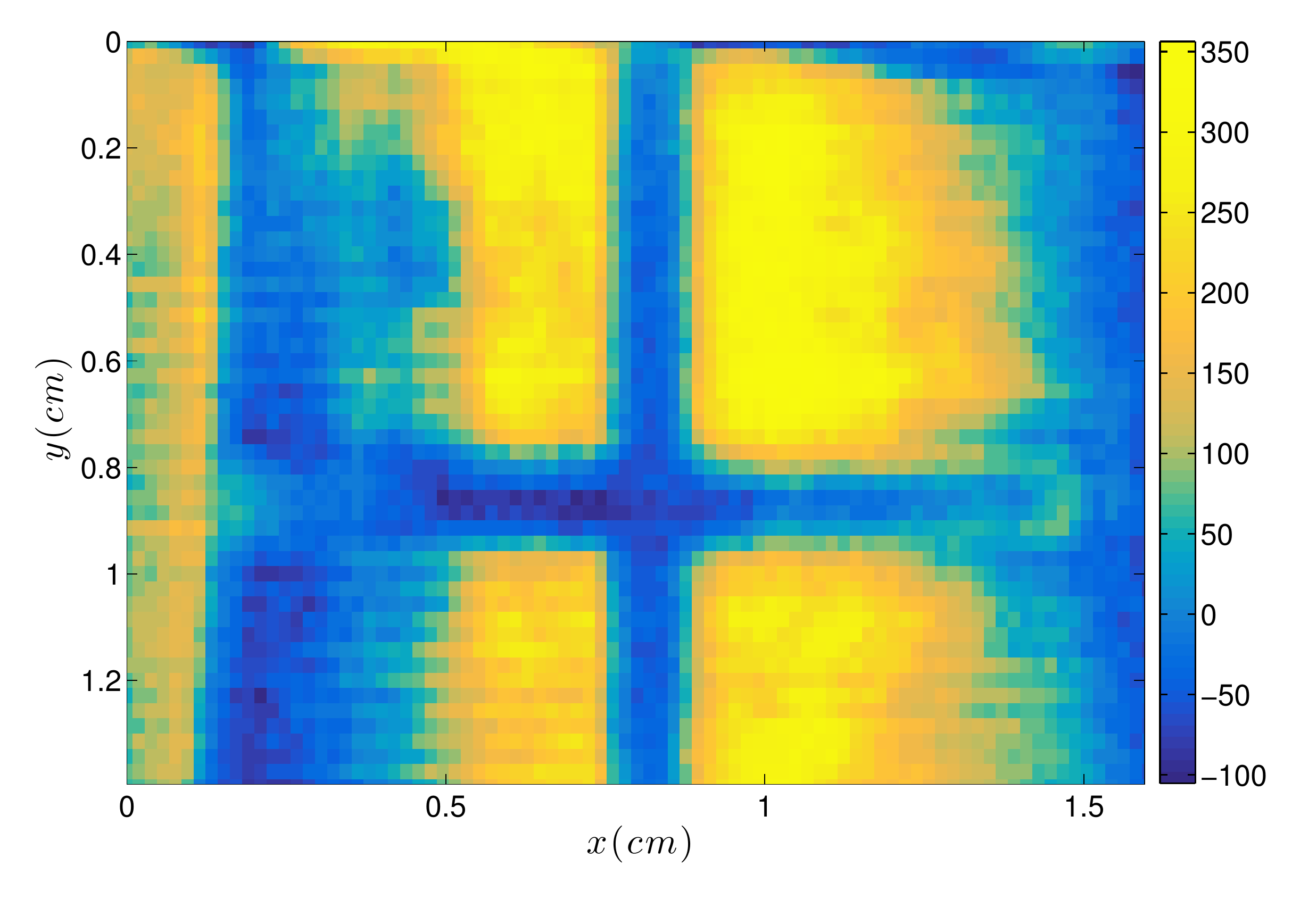}\put (45,-10) {(a.3)}
\end{overpic}
\\[.2cm]
\hspace{-.2cm}\begin{overpic}[trim={0.2cm 0.5cm 0.5cm 0.2cm}, clip,width=1.1in,height=.7in]{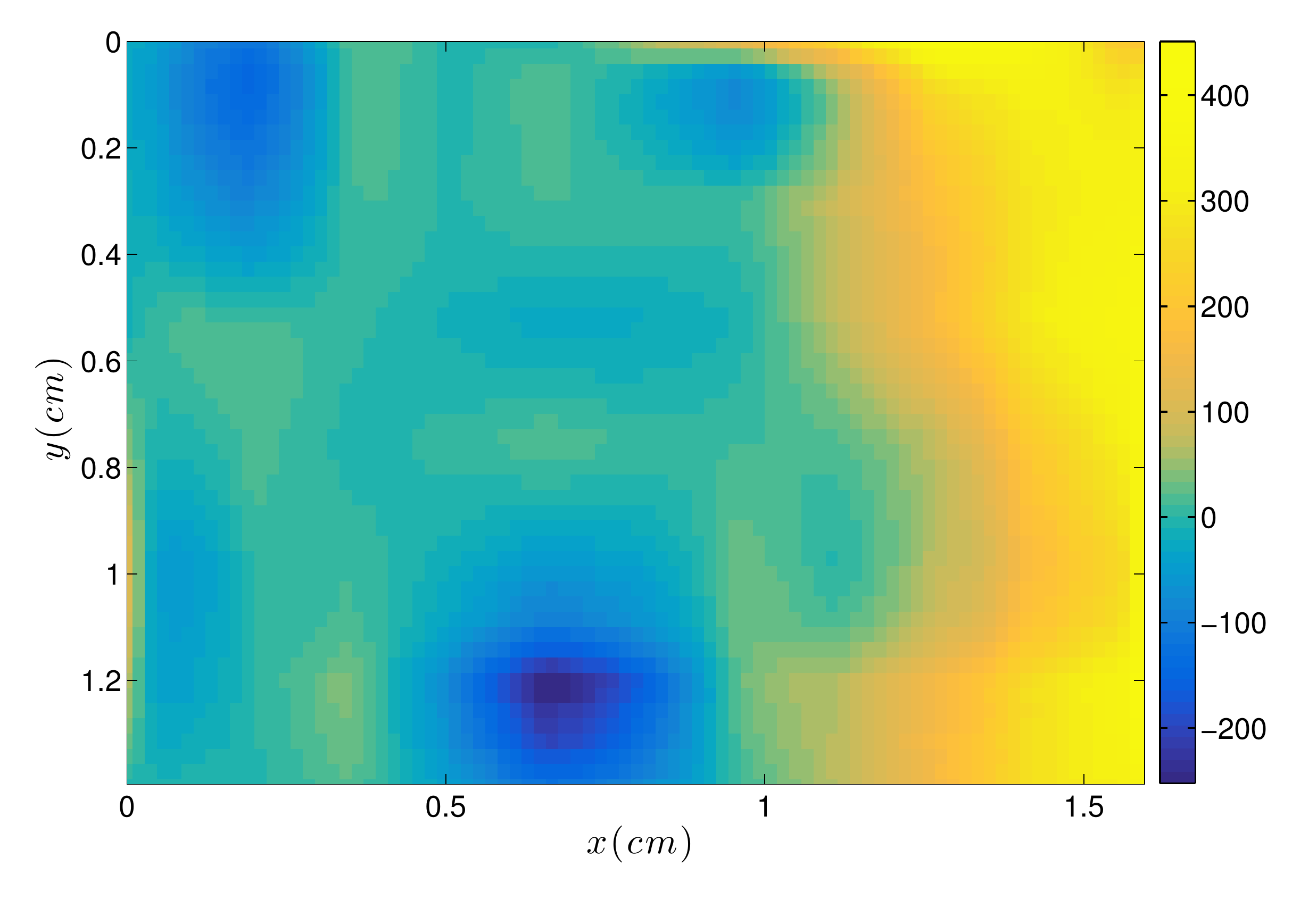}\put (45,-10) {(b.1)} \end{overpic}
&
\hspace{-.9cm}\begin{overpic}[trim={0.2cm 0.5cm 0.5cm 0.2cm}, clip,width=1.1in,height=.7in]{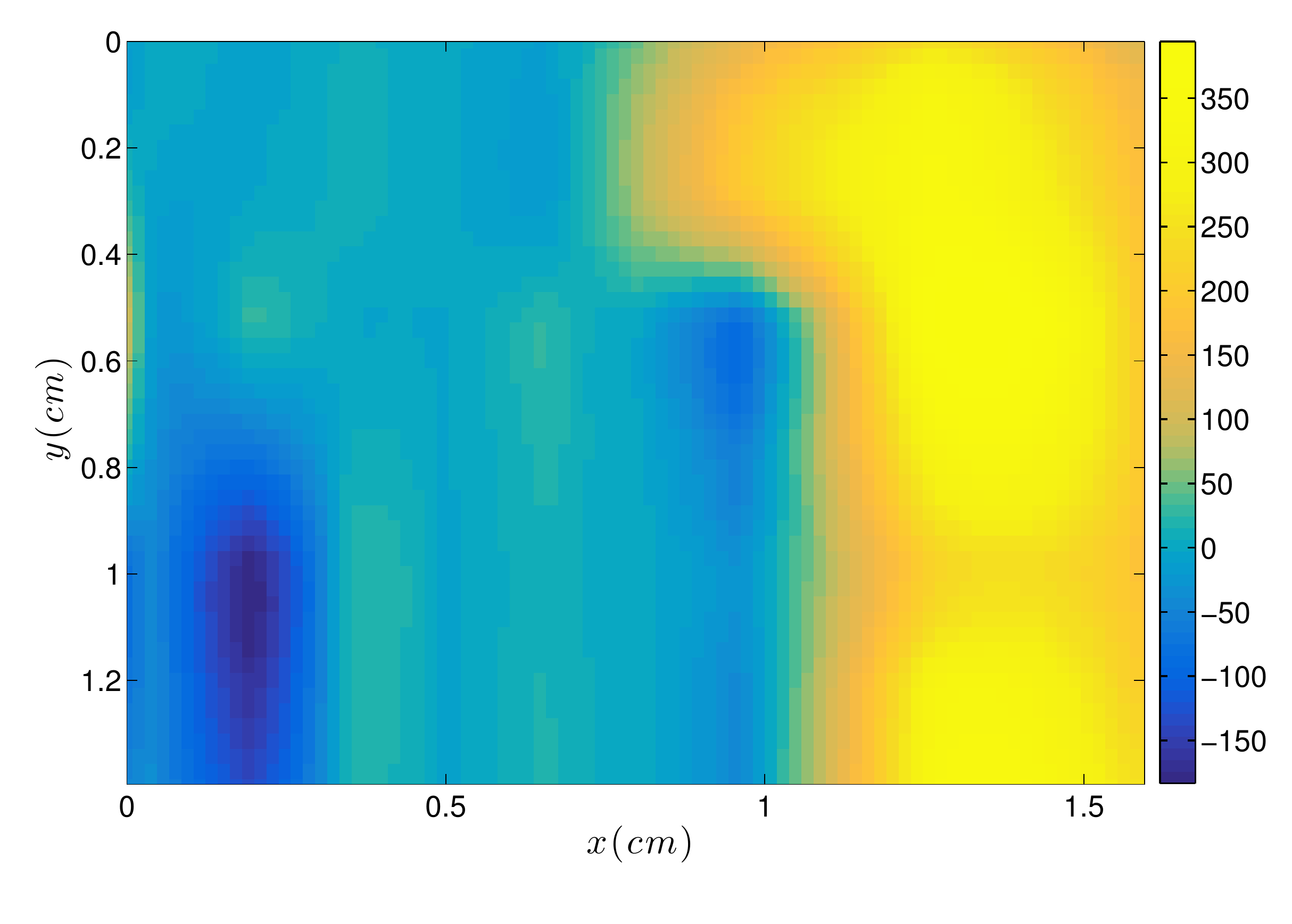}\put (45,-10) {(b.2)}\end{overpic}
&
\hspace{-.9cm}\begin{overpic}[trim={0.2cm 0.5cm 0.5cm 0.2cm}, clip,width=1.1in,height=.7in]{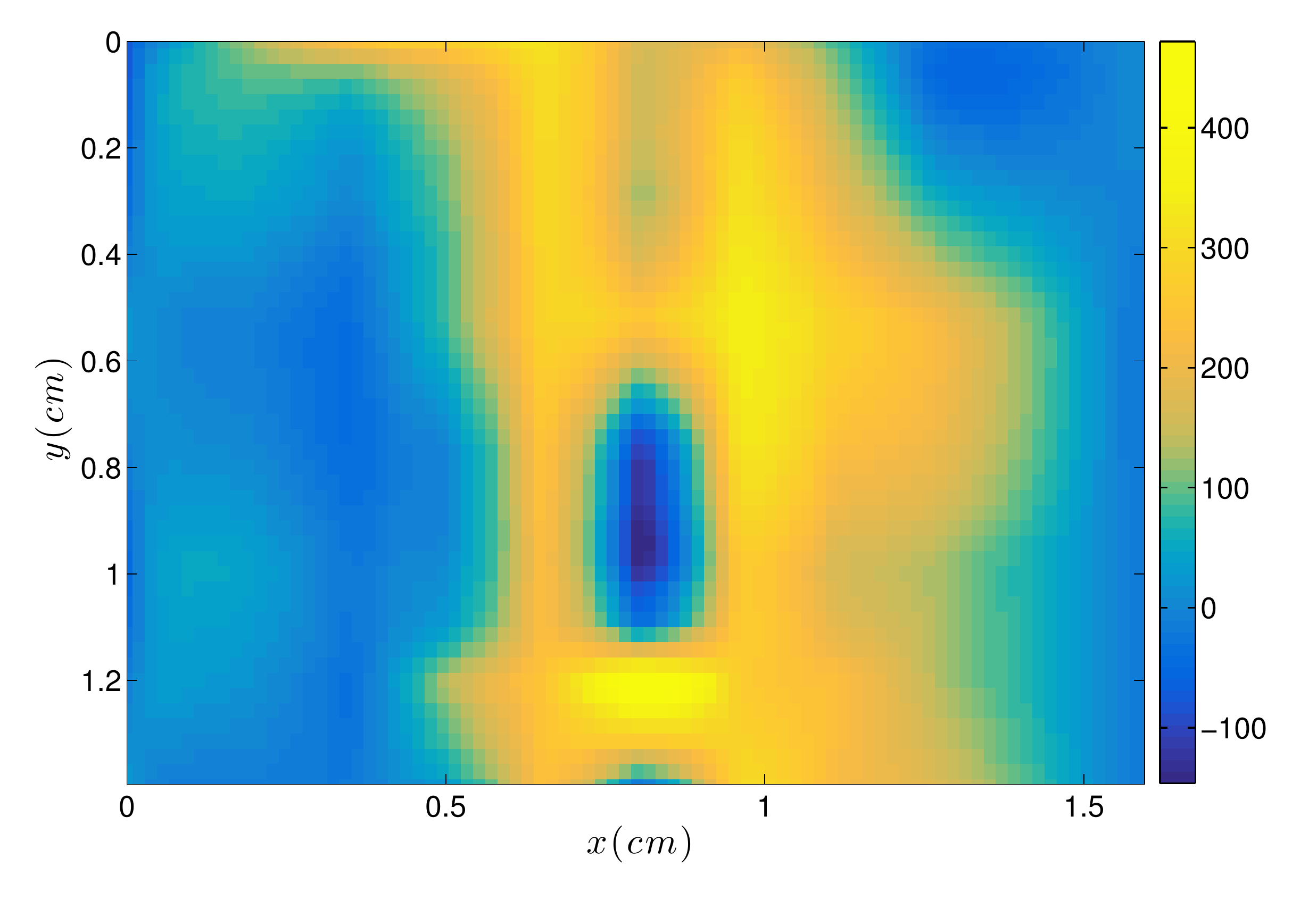}\put (45,-10) {(b.3)}
\end{overpic}
\\[.2cm]
&
\hspace{-1.0cm}\begin{overpic}[trim={0.2cm 0.5cm 0.5cm 0.2cm}, clip,width=1.6in,height=1in]{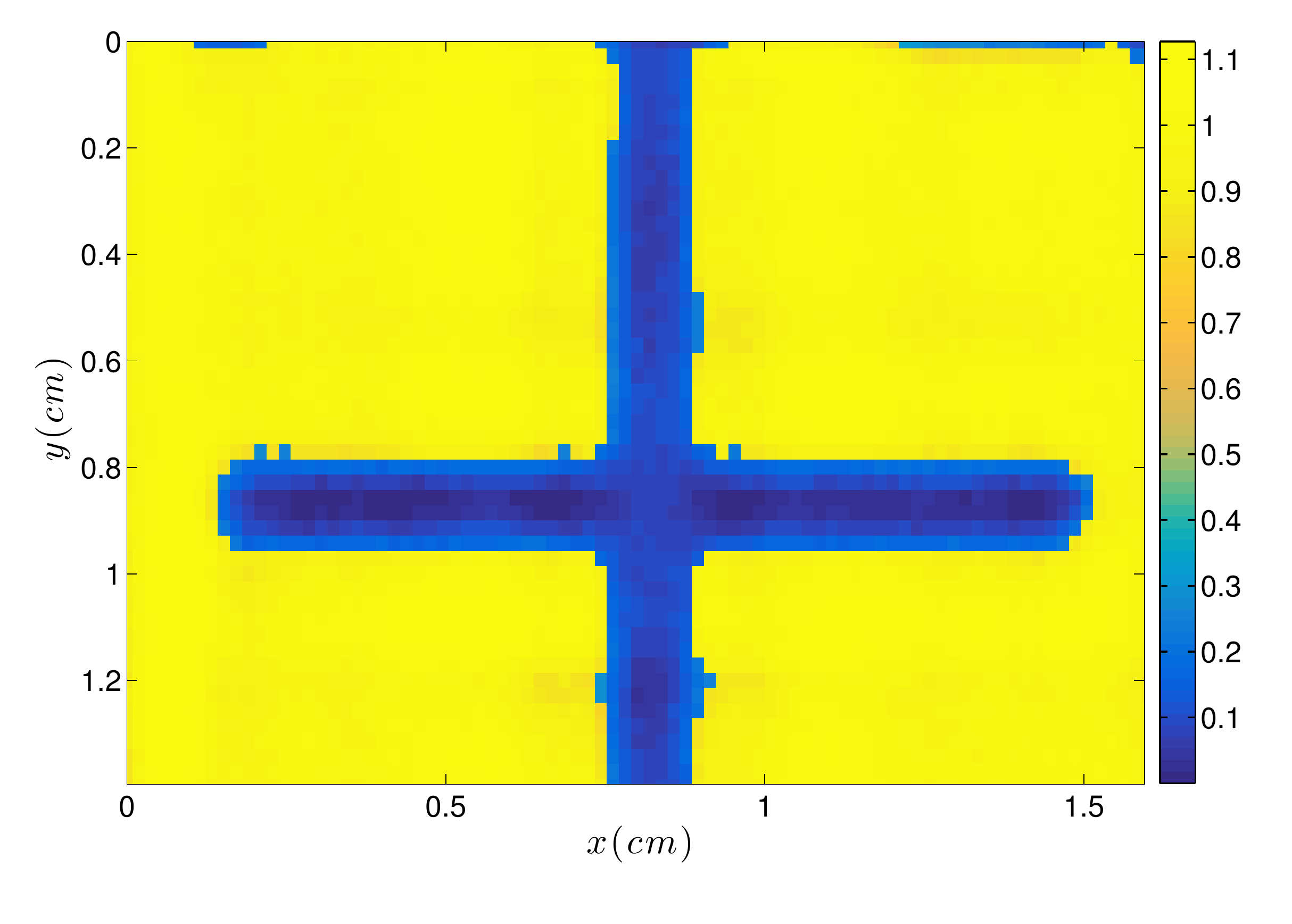}\put (45,-5) {(c)}\end{overpic}
&
\end{tabular}
\caption{ Experimental demonstration of sweep decoupling for a metallic surface; (a.1-3) three time instances of the recorded image from the sample, vertical sweep distortions are dominantly over the cross pattern; (b.1-3) the recovered distortion profiles corresponding to the observations shown; (c) the recovered binary profile }\label{fig6}
\end{figure}

To experimentally apply the demodulation process we used a standard THz-TDS system in reflection geometry. In the first experiment (Figure \ref{fig6}) we used a binary sample with metallic surface and a cross shaped pattern carved into a $15$ $mm\times 15$ $mm$ metal surface with 2 $mm$ depth. As noted in Figure \ref{fig6}(a1-3), the spatial sweep distortions are dominantly present in the time domain observations. For this experiment $M=25$ frames were available, three of which are shown here.

 Figures \ref{fig6}(b1-3) and \ref{fig6}(c) report the decoupled sweep distortions from the binary cross profile. We can observe a good characterization of the cross profile thanks to the large number of available frames and high SNR observations. A more challenging experiment with less number of available frames and noisier data is presented next.  

\begin{figure}[htb!]
\begin{tabular}{ccc}
\hspace{-.2cm}\begin{overpic}[trim={0.2cm 0.5cm 0.5cm 0.2cm}, clip,width=1.1in,height=.7in]{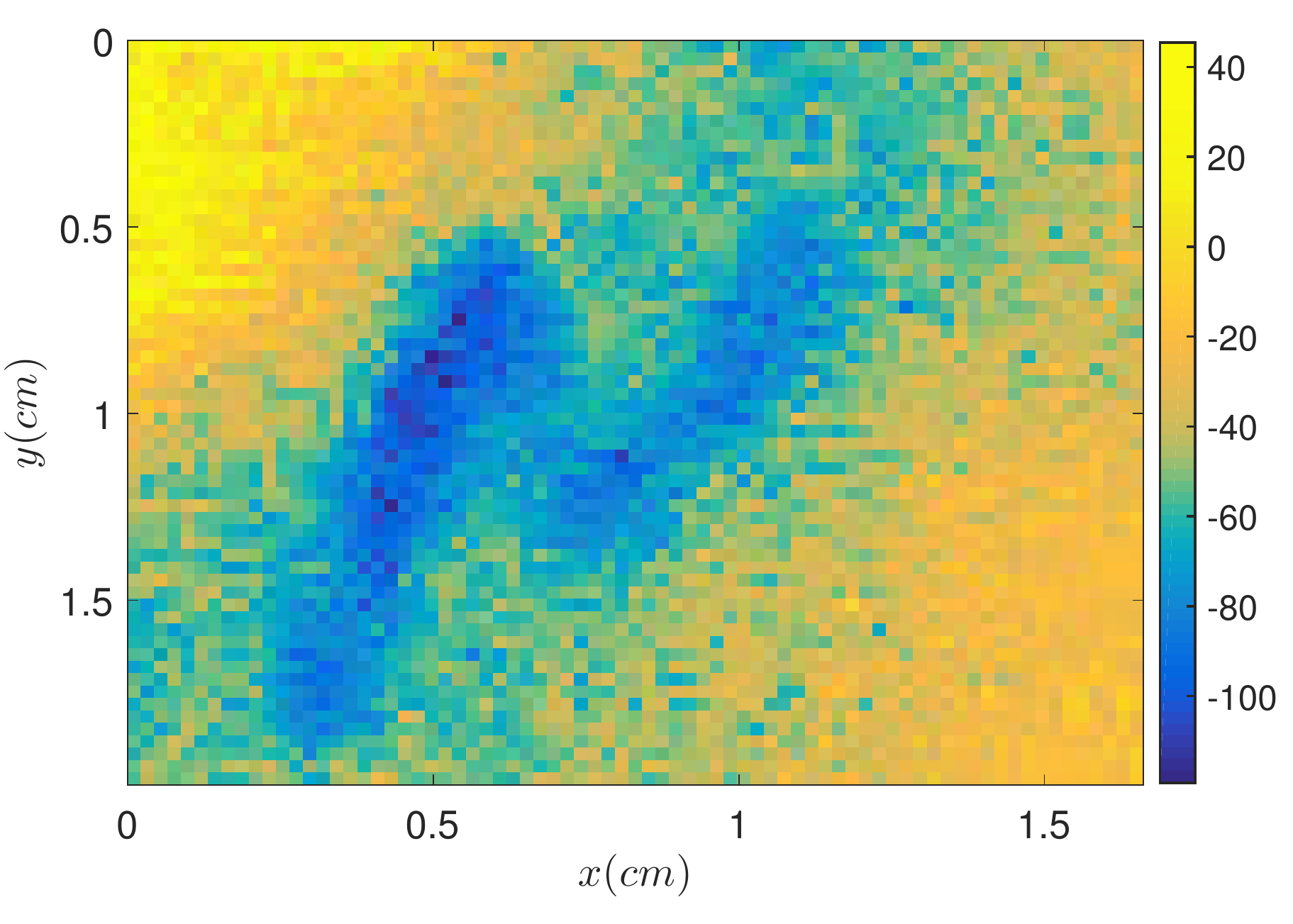}\put (45,-10) {(a.1)} \end{overpic}
&
\hspace{-.23cm}\begin{overpic}[trim={0.2cm 0.5cm 0.5cm 0.2cm}, clip,width=1.1in,height=.7in]{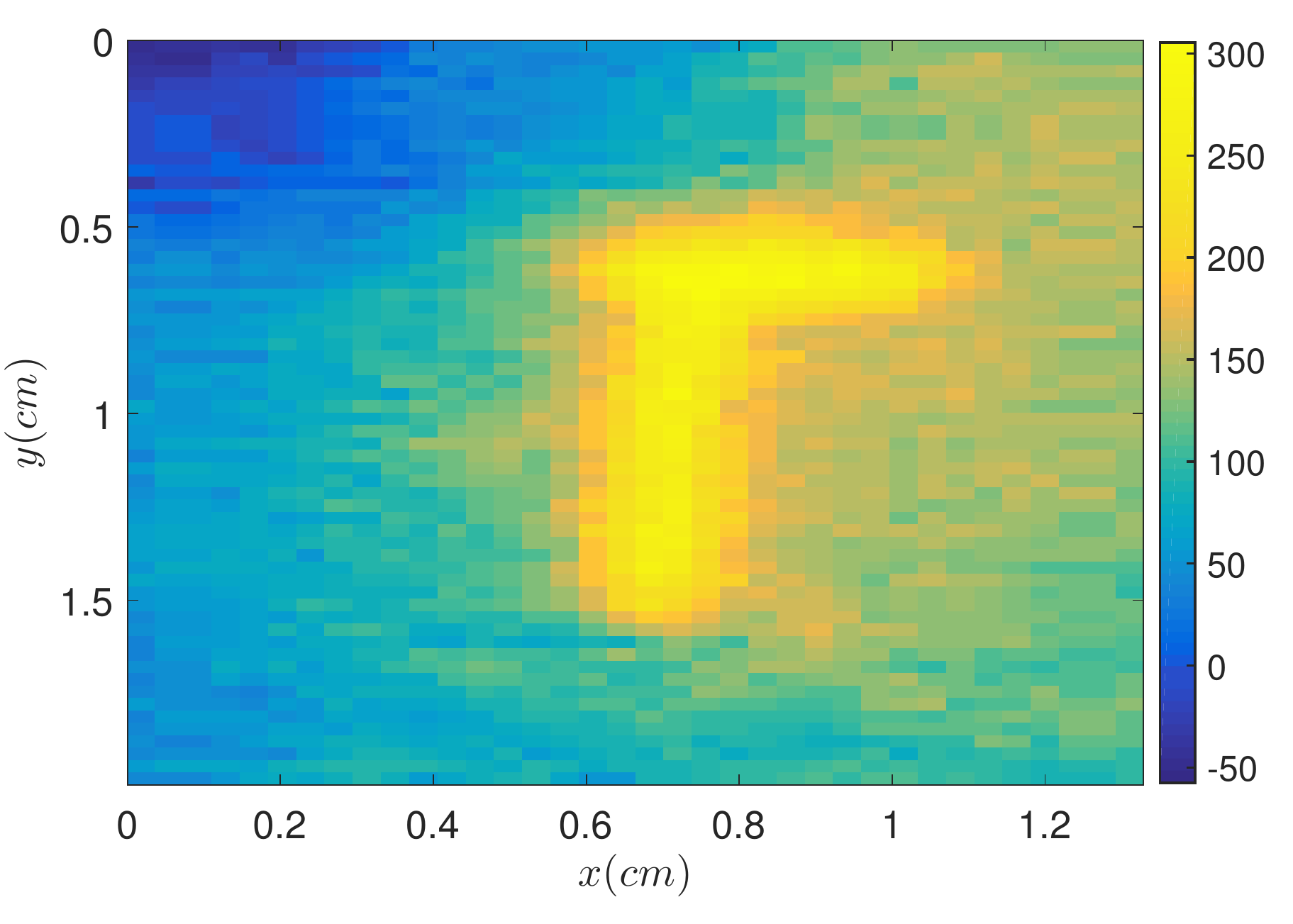}\put (45,-10) {(b.1)}\end{overpic}
&
\hspace{-.23cm}\begin{overpic}[trim={0.2cm 0.5cm 0.5cm 0.2cm}, clip,width=1.1in,height=.7in]{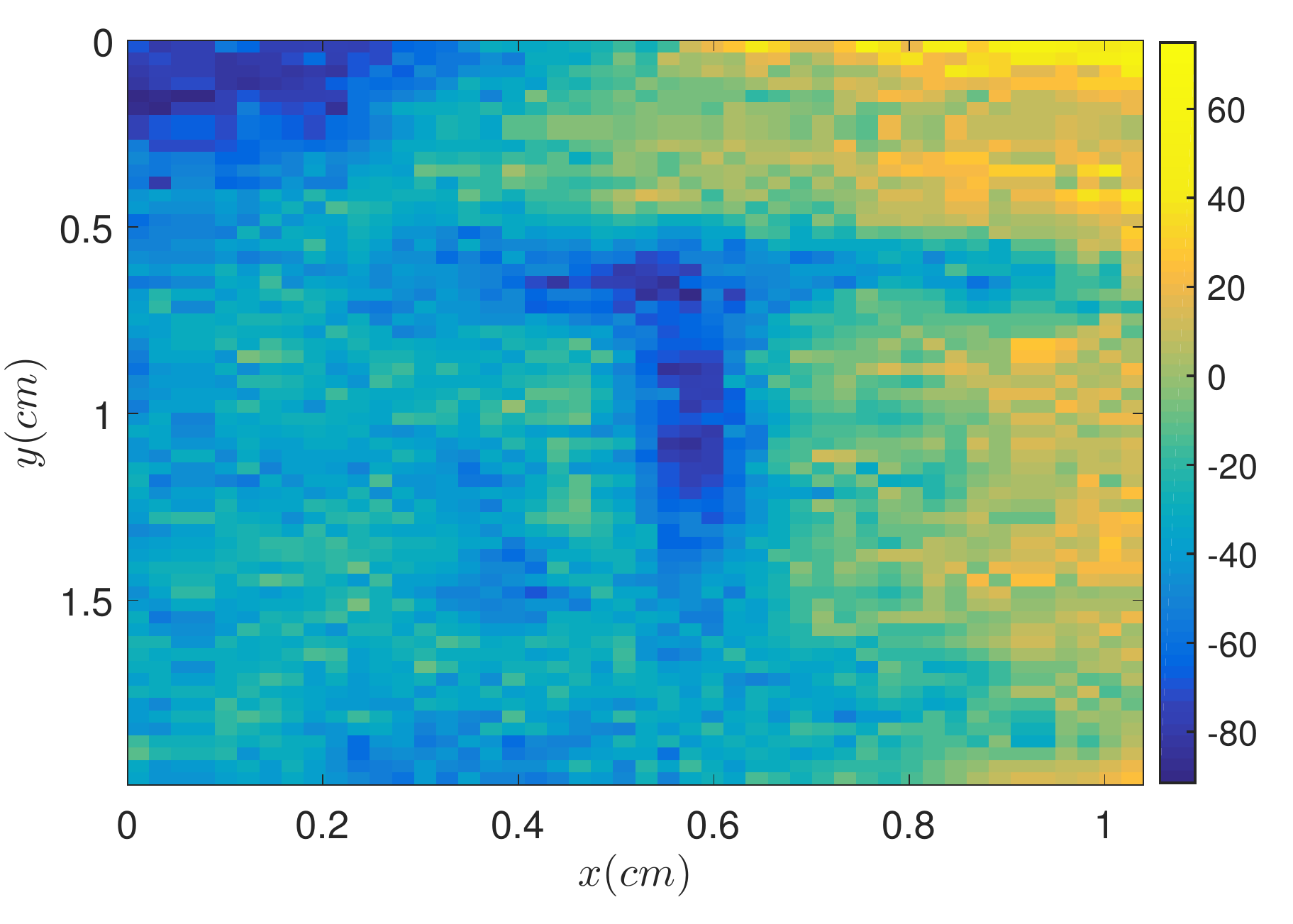}\put (45,-10) {(c.1)}
\end{overpic}
\\[.2cm]
\hspace{-.2cm}\begin{overpic}[trim={0.2cm 0.5cm 0.5cm 0.2cm}, clip,width=1.1in,height=.7in]{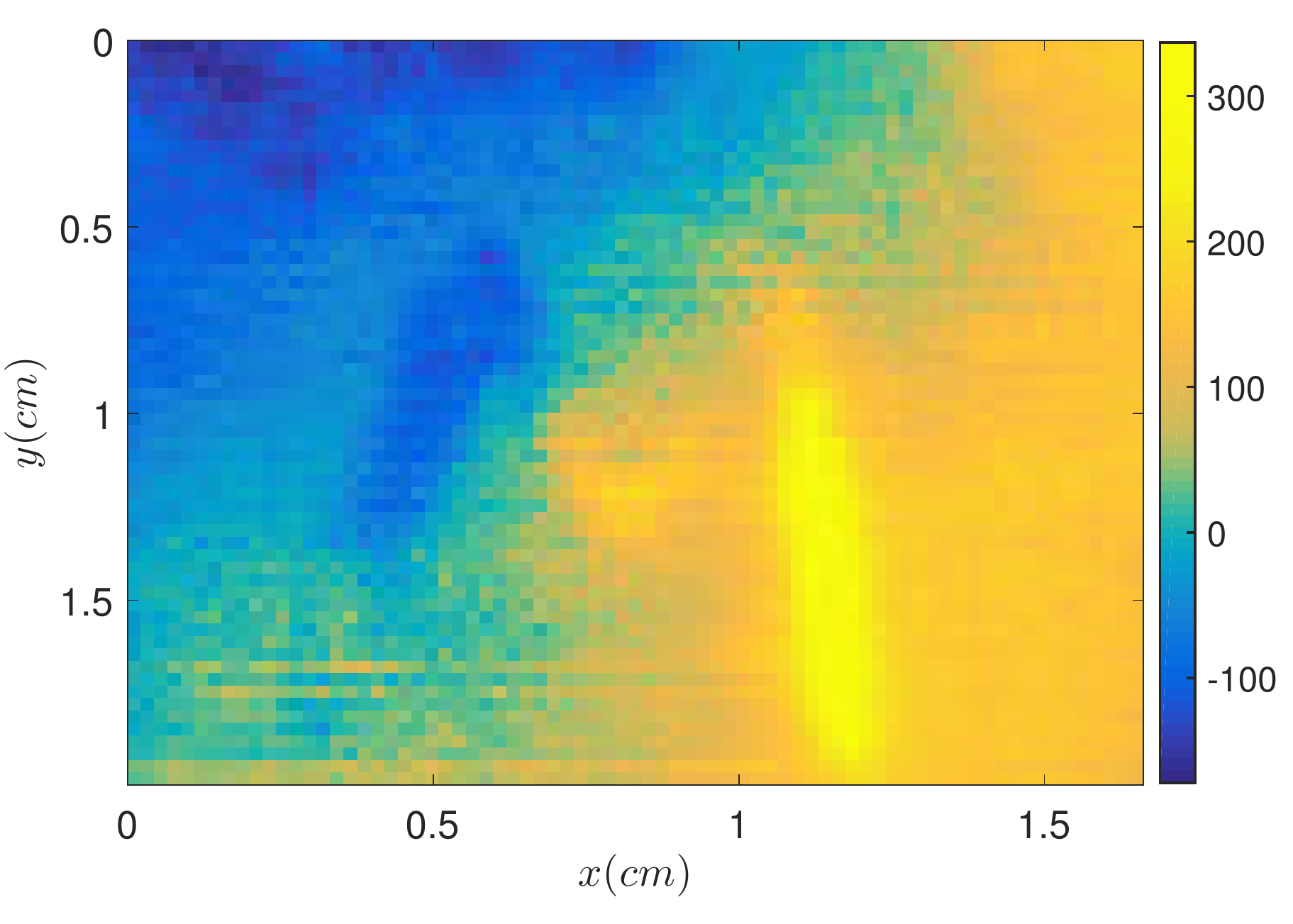}\put (45,-10) {(a.2)} \end{overpic}
&
\hspace{-.23cm}\begin{overpic}[trim={0.2cm 0.5cm 0.5cm 0.2cm}, clip,width=1.1in,height=.7in]{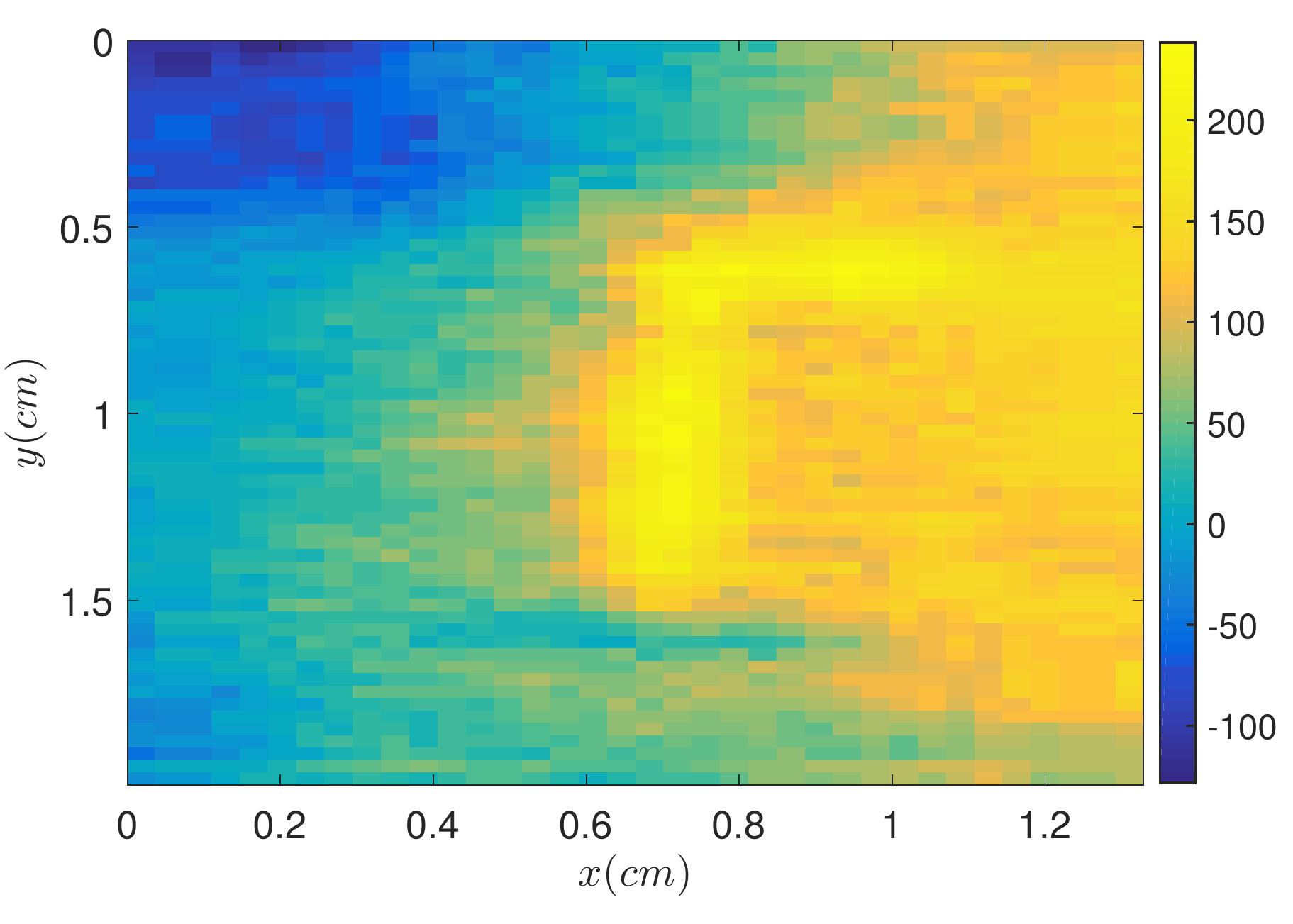}\put (45,-10) {(b.2)}\end{overpic}
&
\hspace{-.23cm}\begin{overpic}[trim={0.2cm 0.5cm 0.5cm 0.2cm}, clip,width=1.1in,height=.7in]{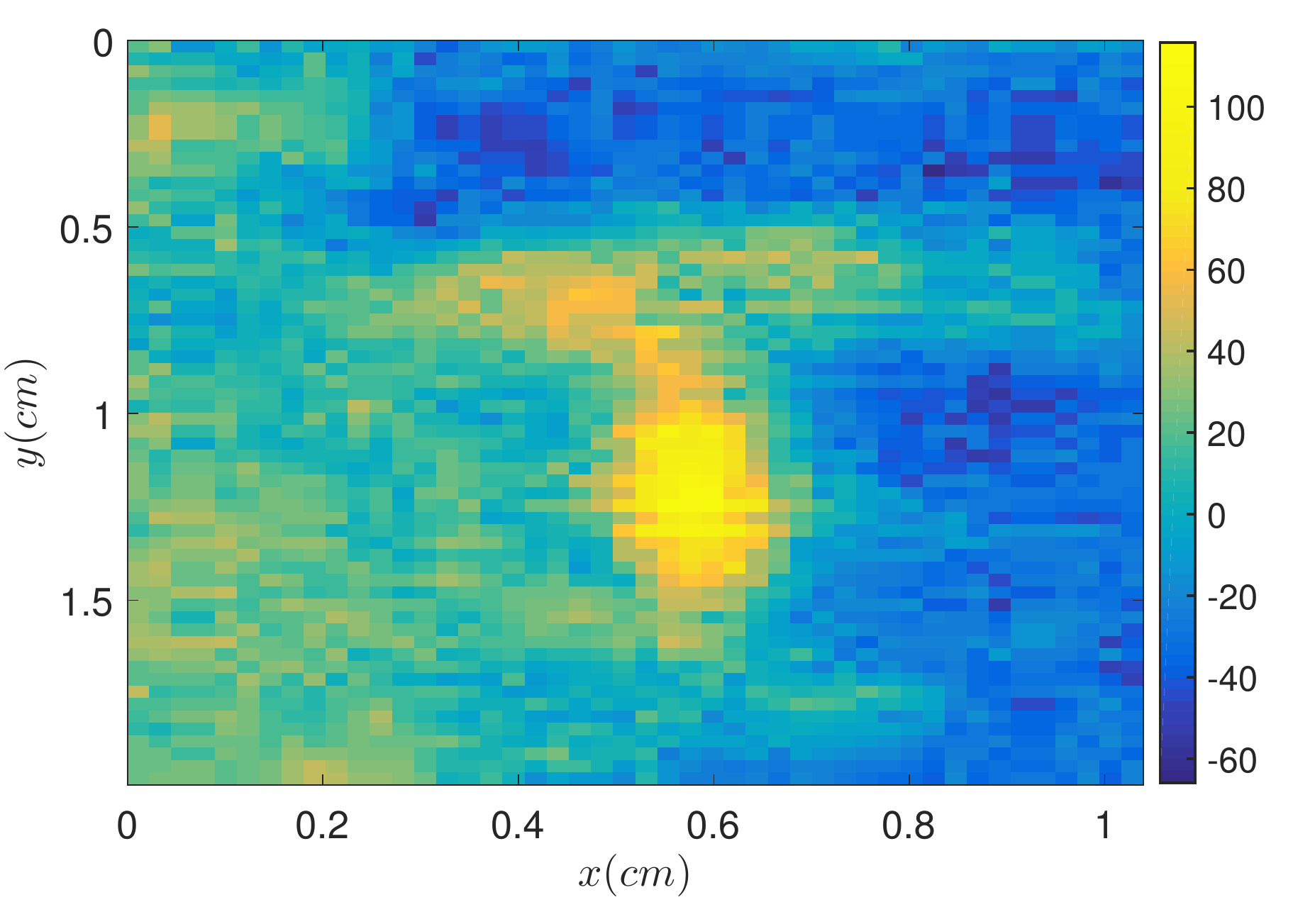}\put (45,-10) {(c.2)}
\end{overpic}
\\[.2cm]
\hspace{-.2cm}\begin{overpic}[trim={0.2cm 0.5cm 0.5cm 0.2cm}, clip,width=1.1in,height=.7in]{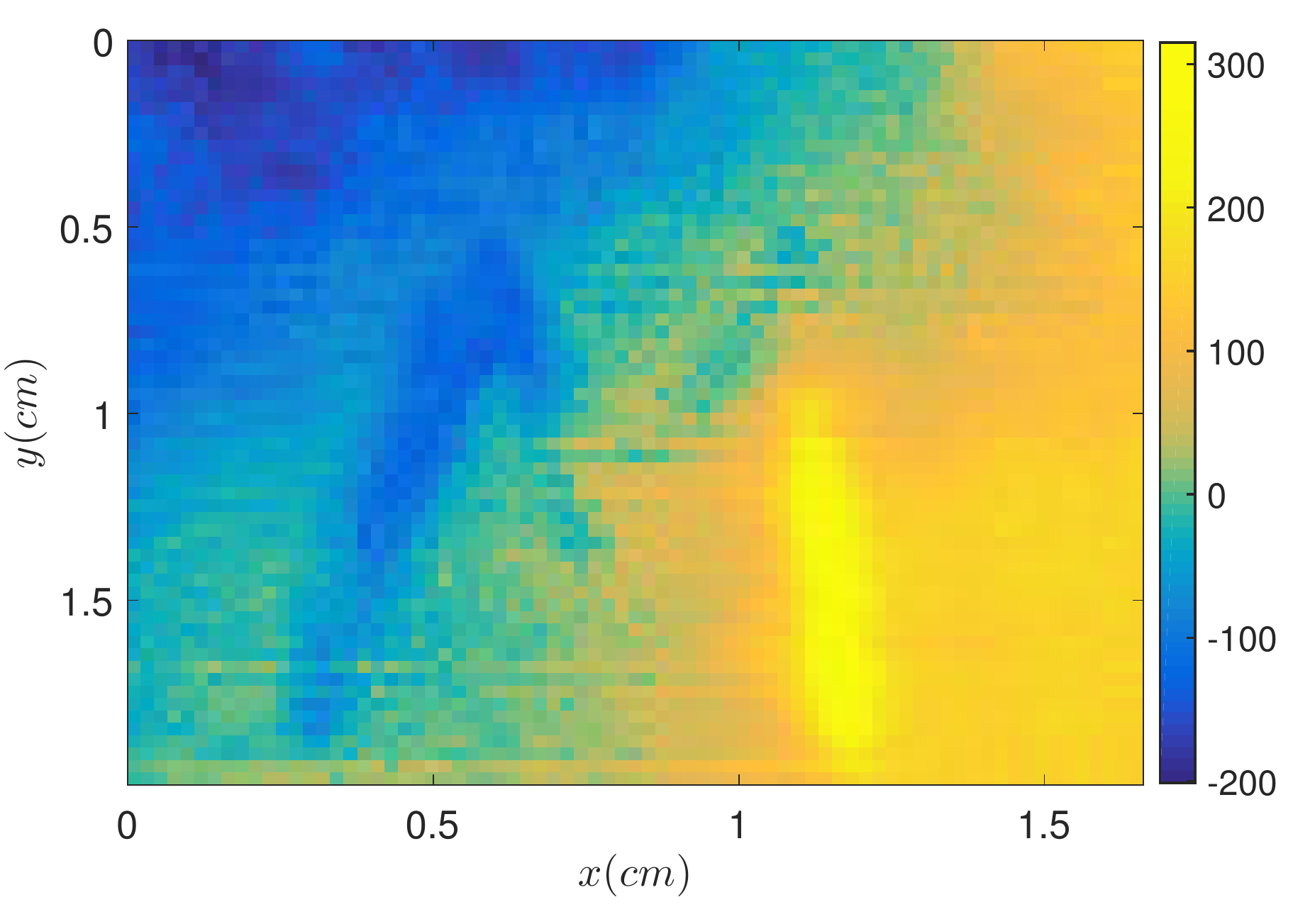}\put (45,-10) {(a.3)} \end{overpic}
&
\hspace{-.23cm}\begin{overpic}[trim={0.2cm 0.5cm 0.5cm 0.2cm}, clip,width=1.1in,height=.7in]{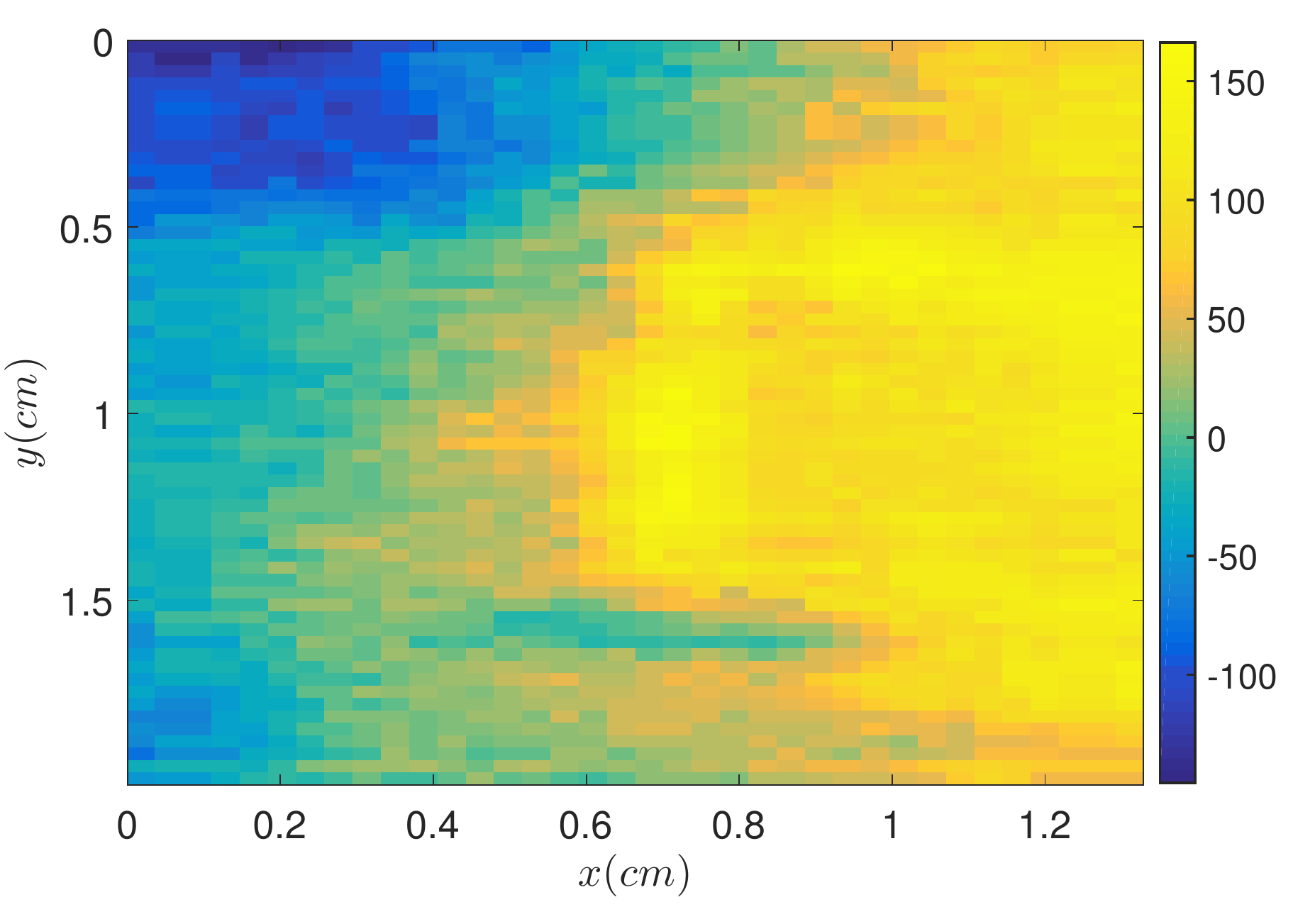}\put (45,-10) {(b.3)}\end{overpic}
&
\hspace{-.23cm}\begin{overpic}[trim={0.2cm 0.5cm 0.5cm 0.2cm}, clip,width=1.1in,height=.7in]{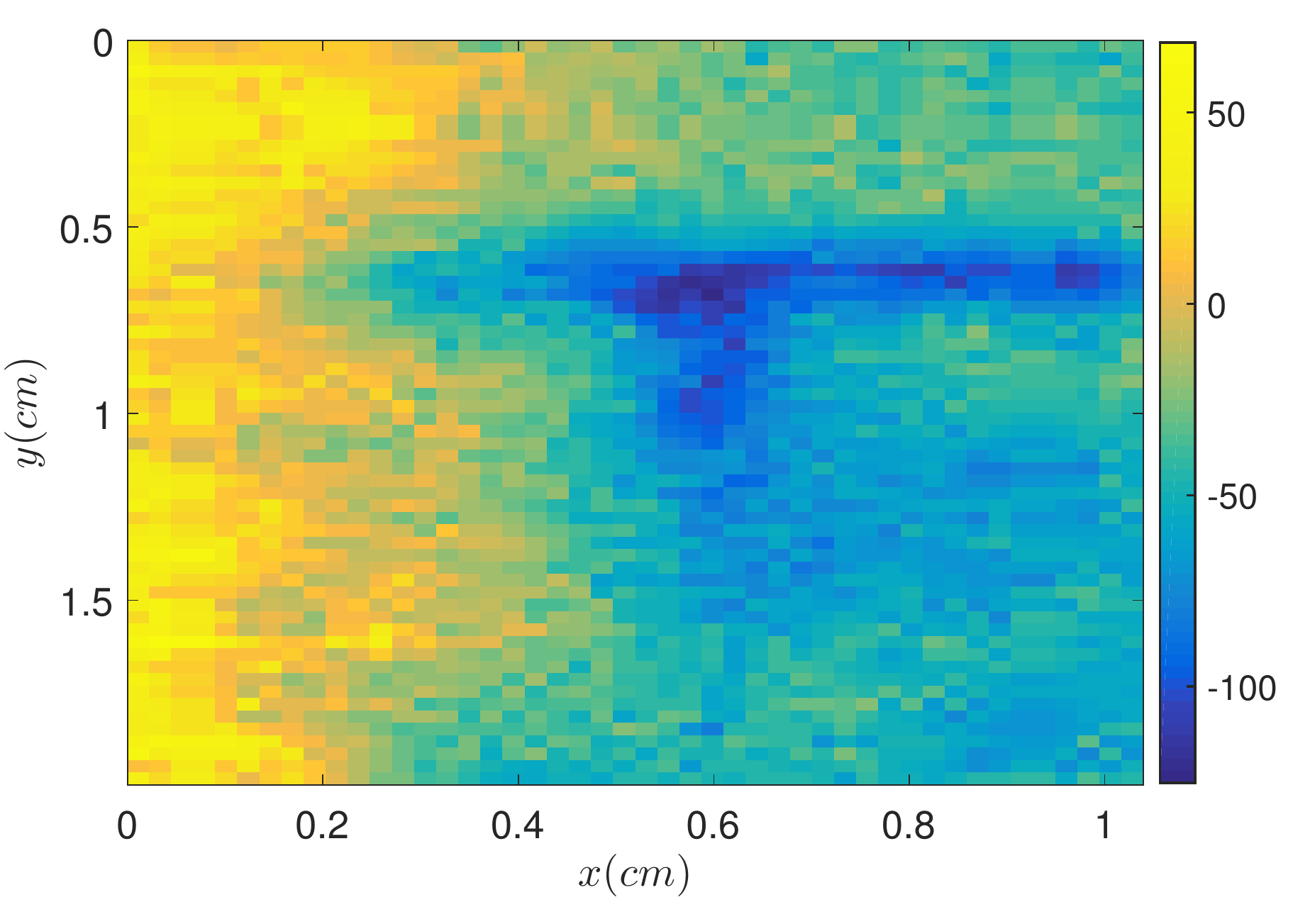}\put (45,-10) {(c.3)}
\end{overpic}
\\[.2cm]
\hspace{-.2cm}\begin{overpic}[trim={0.2cm 0.5cm 0.5cm 0.2cm}, clip,width=1.1in,height=.7in]{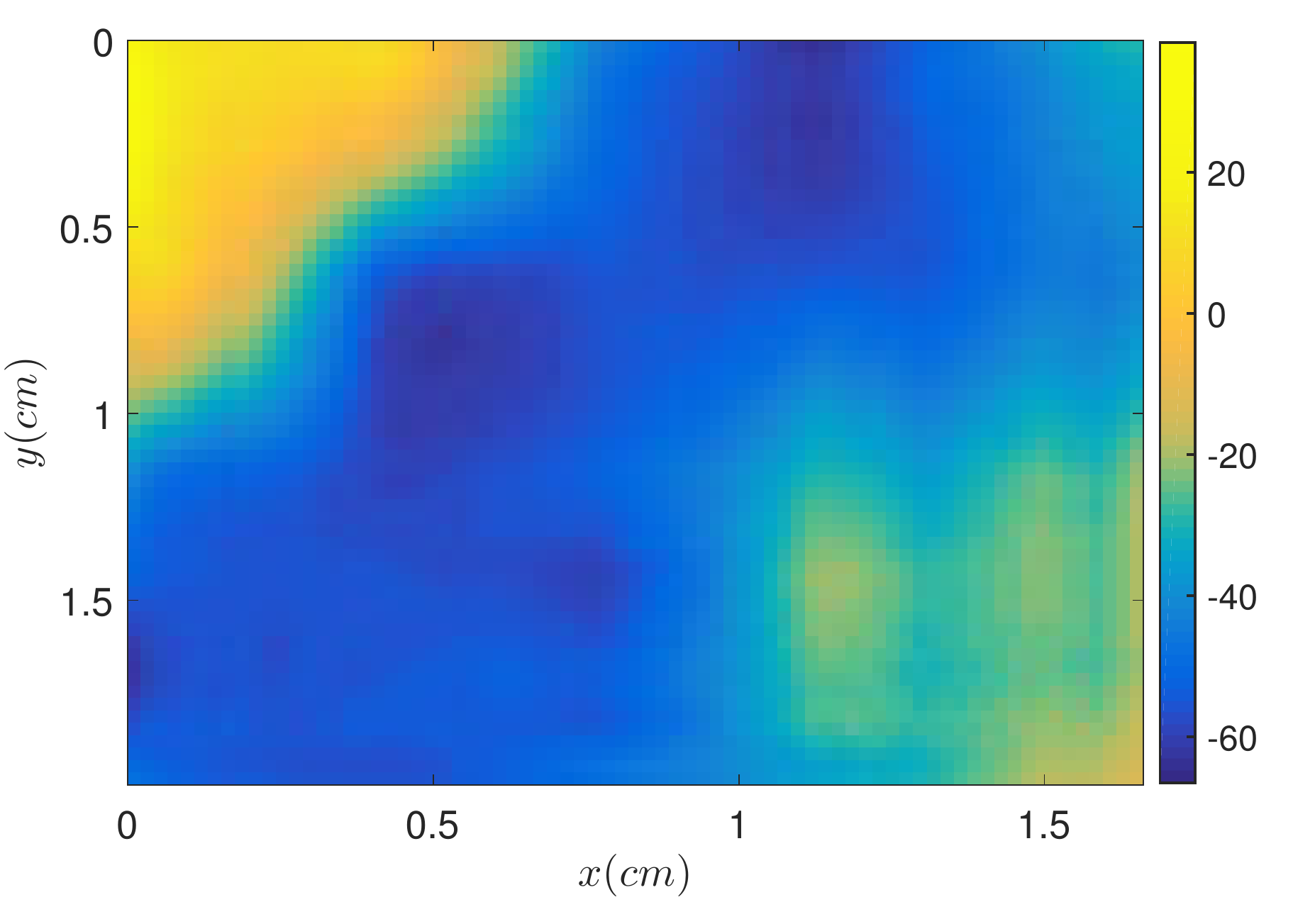}\put (45,-10) {(a.4)} \end{overpic}
&
\hspace{-.23cm}\begin{overpic}[trim={0.2cm 0.5cm 0.5cm 0.2cm}, clip,width=1.1in,height=.7in]{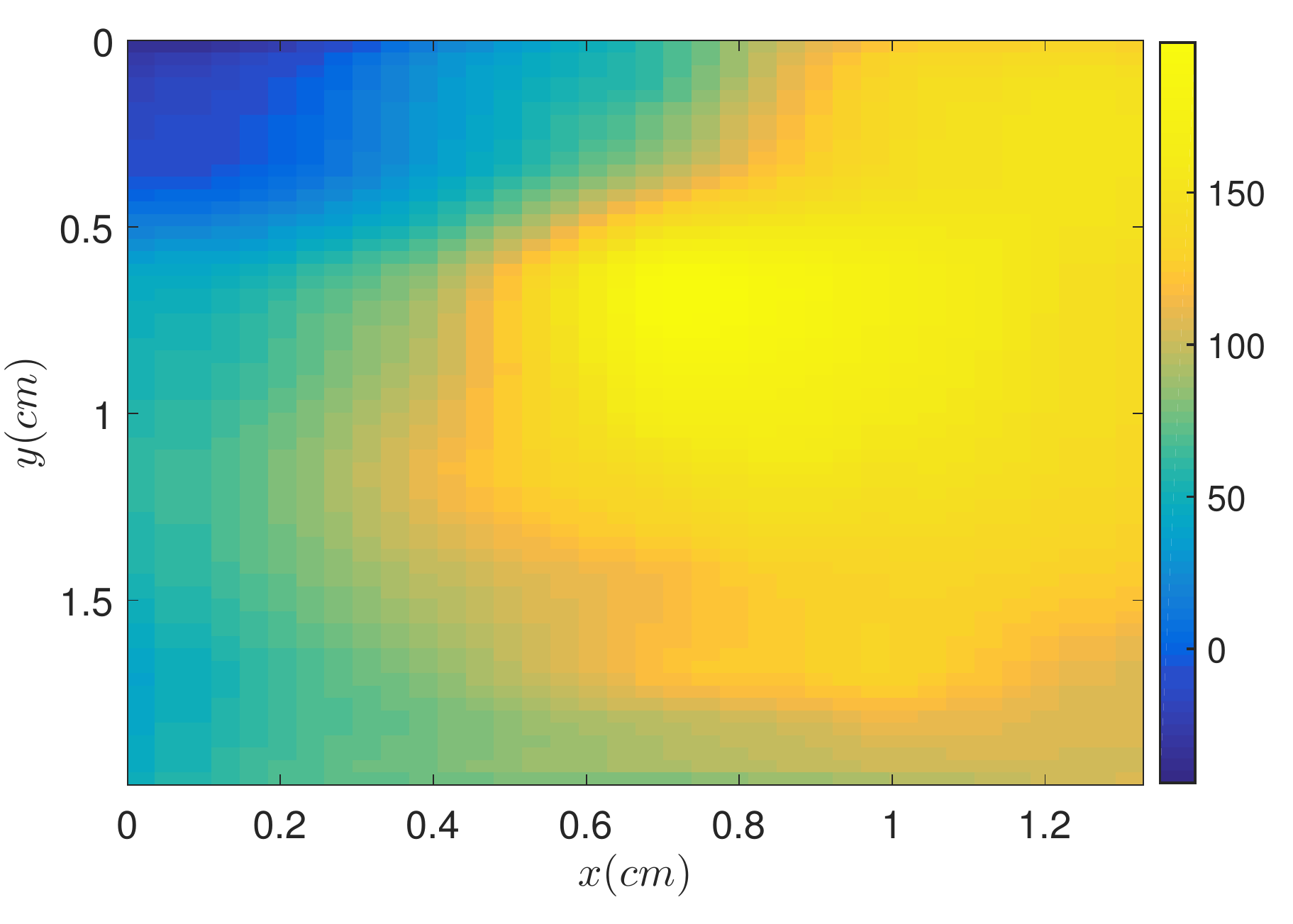}\put (45,-10) {(b.4)}\end{overpic}
&
\hspace{-.23cm}\begin{overpic}[trim={0.2cm 0.5cm 0.5cm 0.2cm}, clip,width=1.1in,height=.7in]{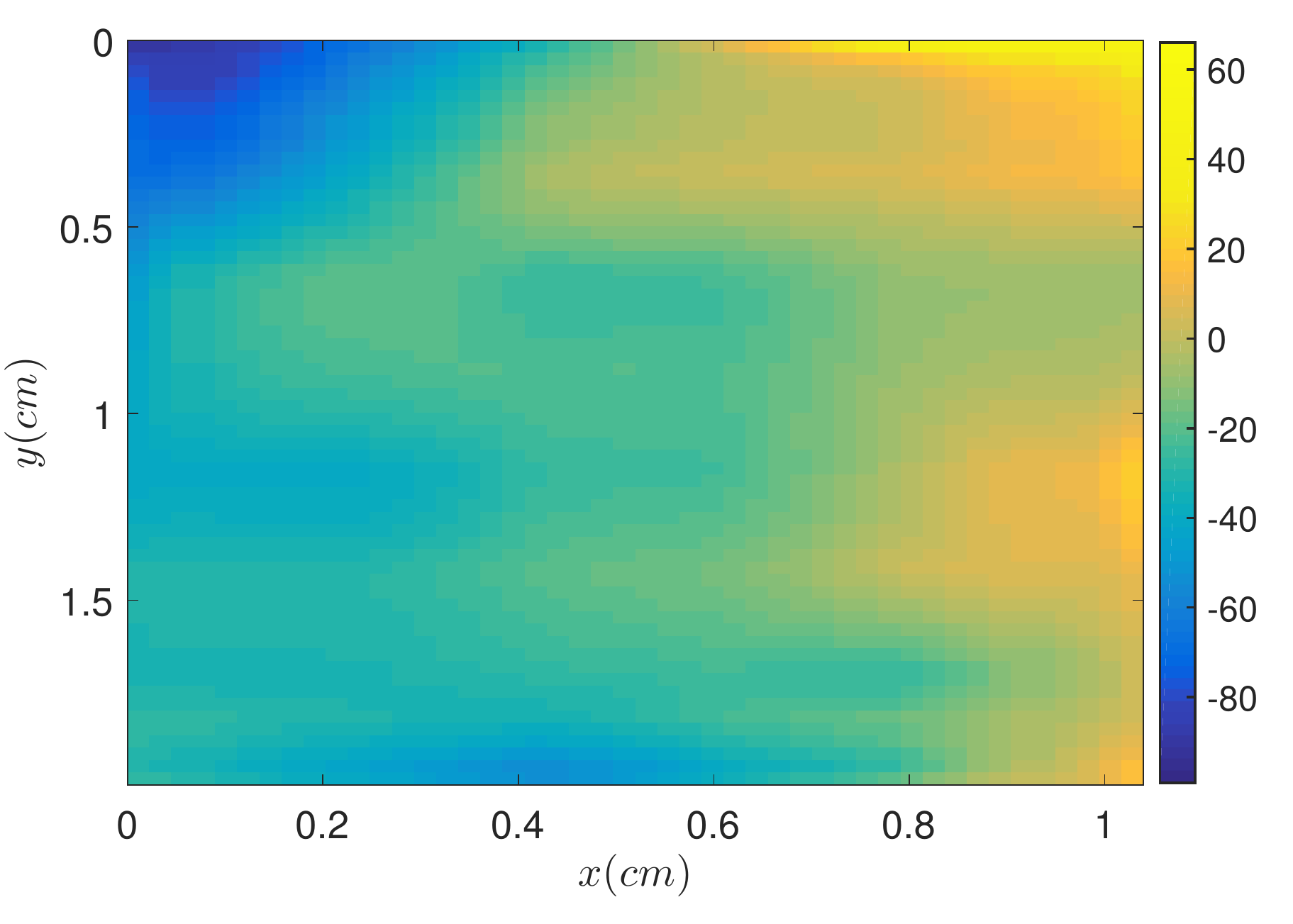}\put (45,-10) {(c.4)}
\end{overpic}
\\[.2cm]
\hspace{-.2cm}\begin{overpic}[trim={0.2cm 0.5cm 0.5cm 0.2cm}, clip,width=1.1in,height=.7in]{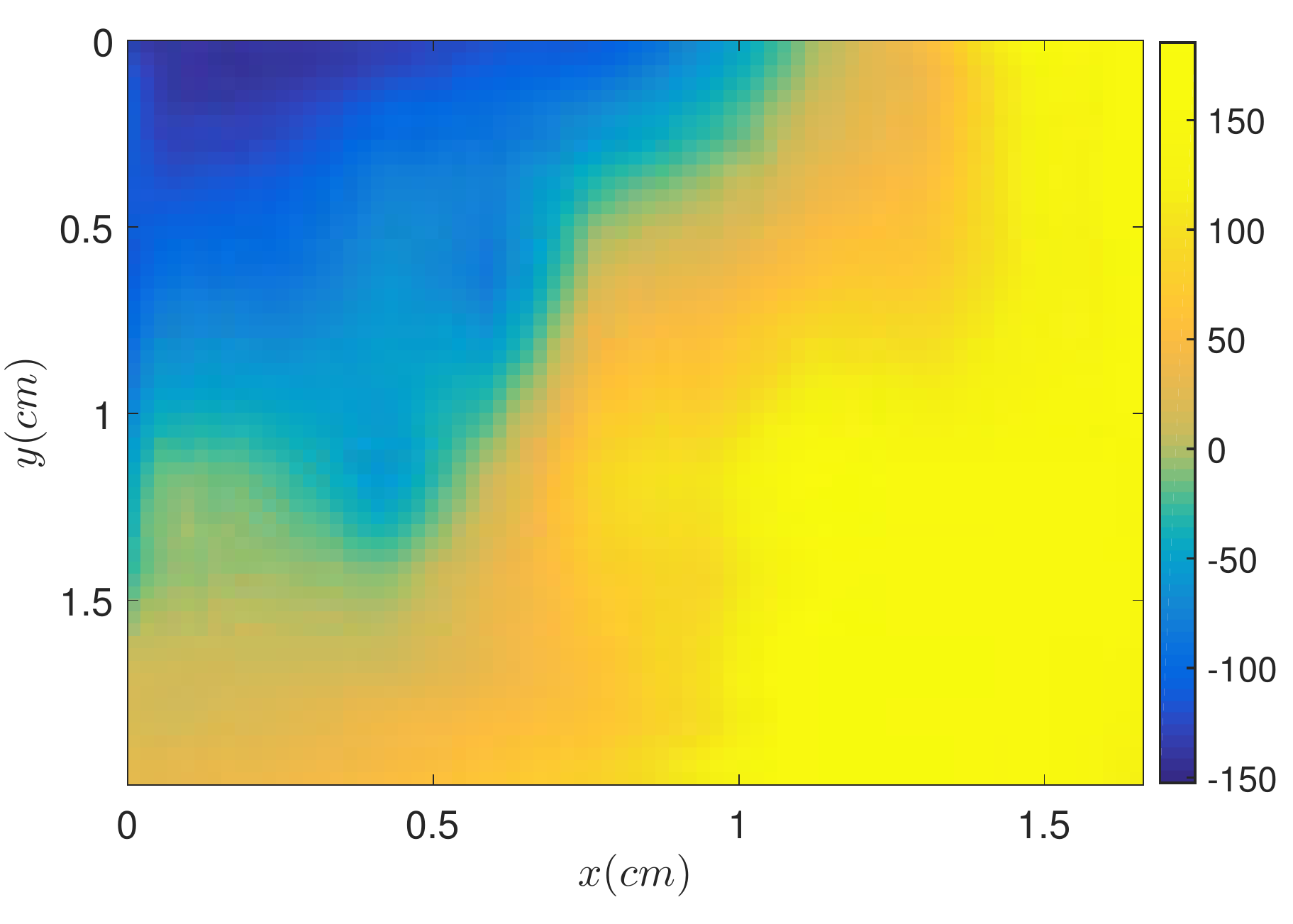}\put (45,-10) {(a.5)} \end{overpic}
&
\hspace{-.23cm}\begin{overpic}[trim={0.2cm 0.5cm 0.5cm 0.2cm}, clip,width=1.1in,height=.7in]{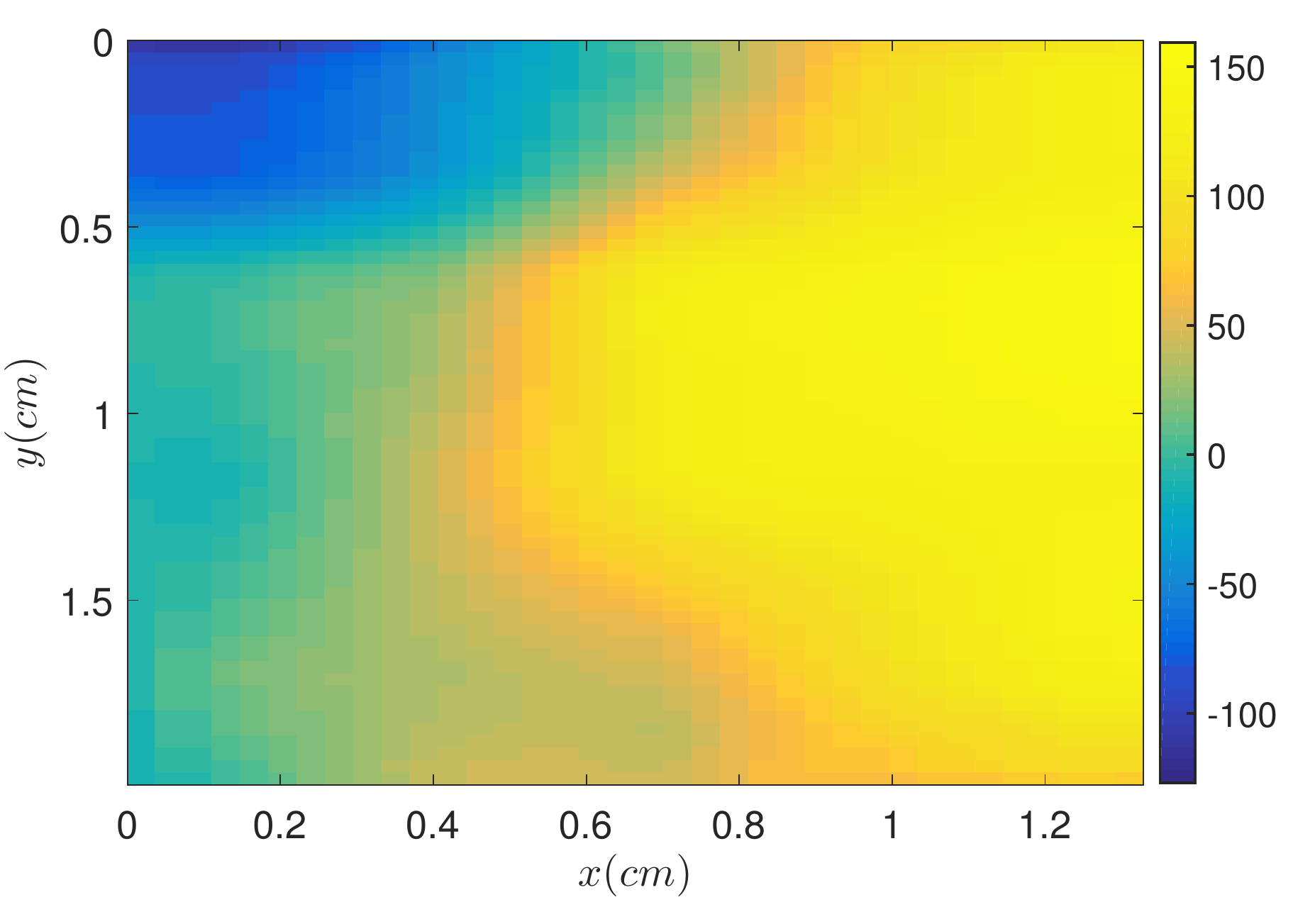}\put (45,-10) {(b.5)}\end{overpic}
&
\hspace{-.23cm}\begin{overpic}[trim={0.2cm 0.5cm 0.5cm 0.2cm}, clip,width=1.1in,height=.7in]{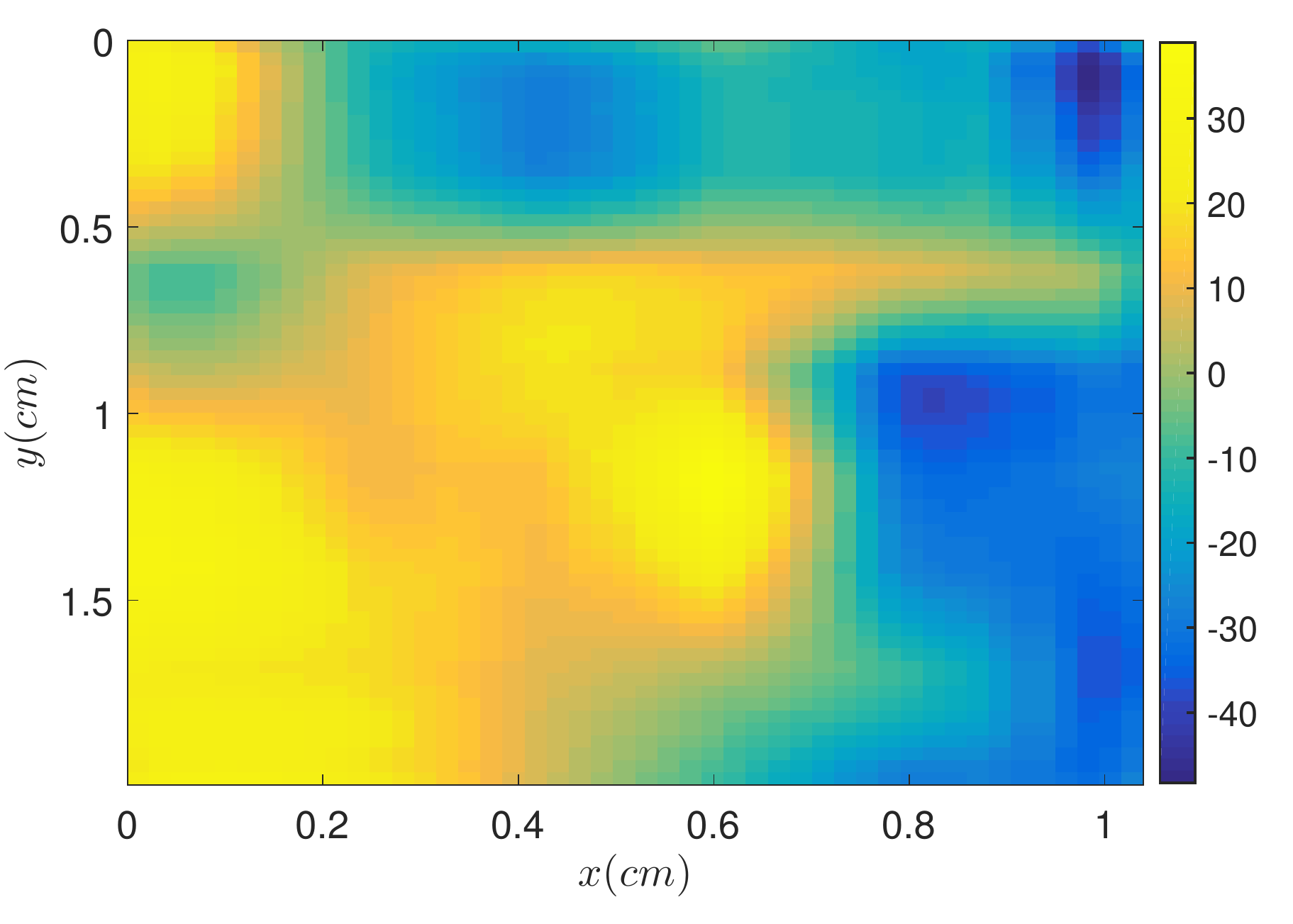}\put (45,-10) {(c.5)}
\end{overpic}
\\[.2cm]
\hspace{-.2cm}\begin{overpic}[trim={0.2cm 0.5cm 0.5cm 0.2cm}, clip,width=1.1in,height=.7in]{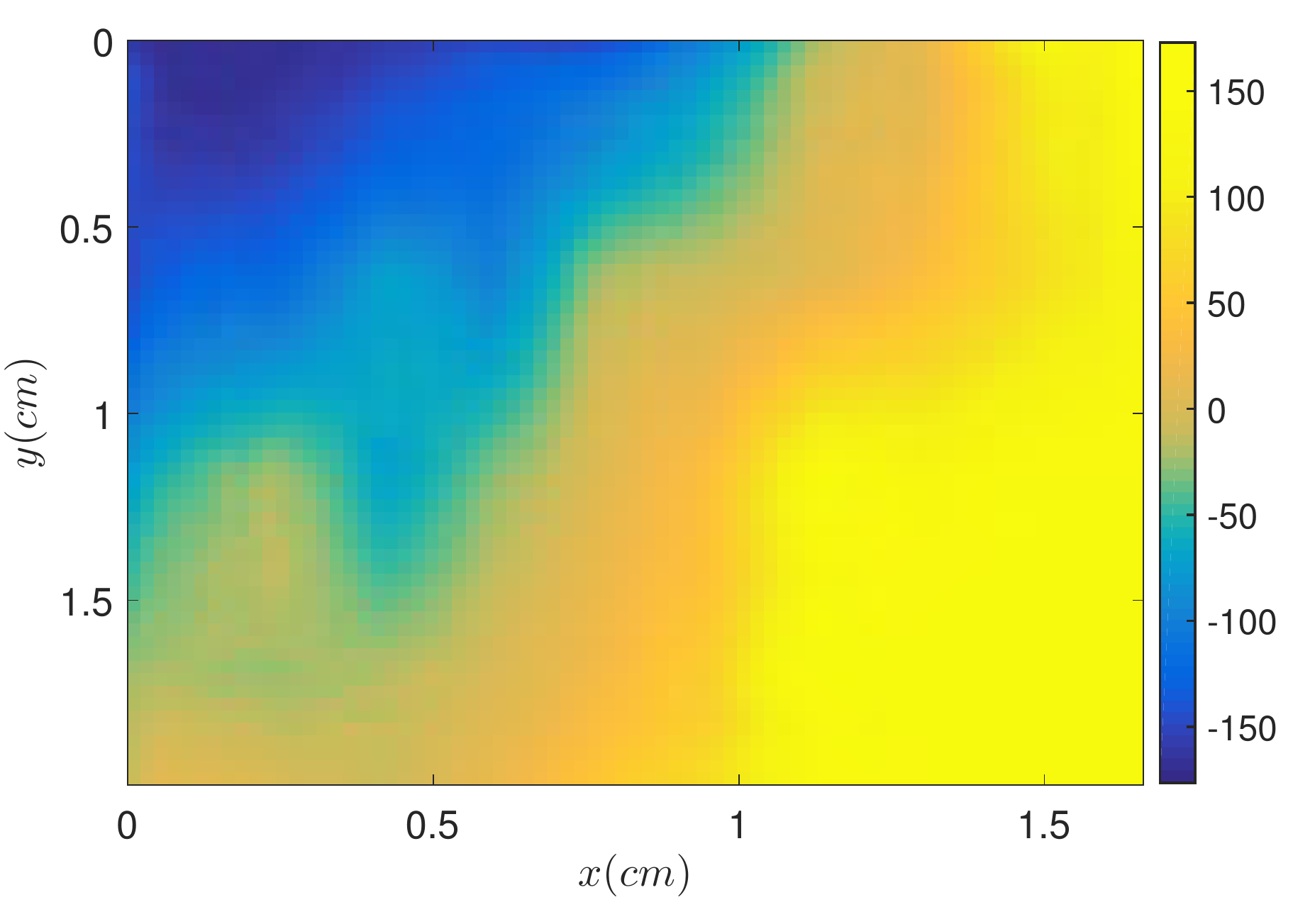}\put (45,-10) {(a.6)} \end{overpic}
&
\hspace{-.23cm}\begin{overpic}[trim={0.2cm 0.5cm 0.5cm 0.2cm}, clip,width=1.1in,height=.7in]{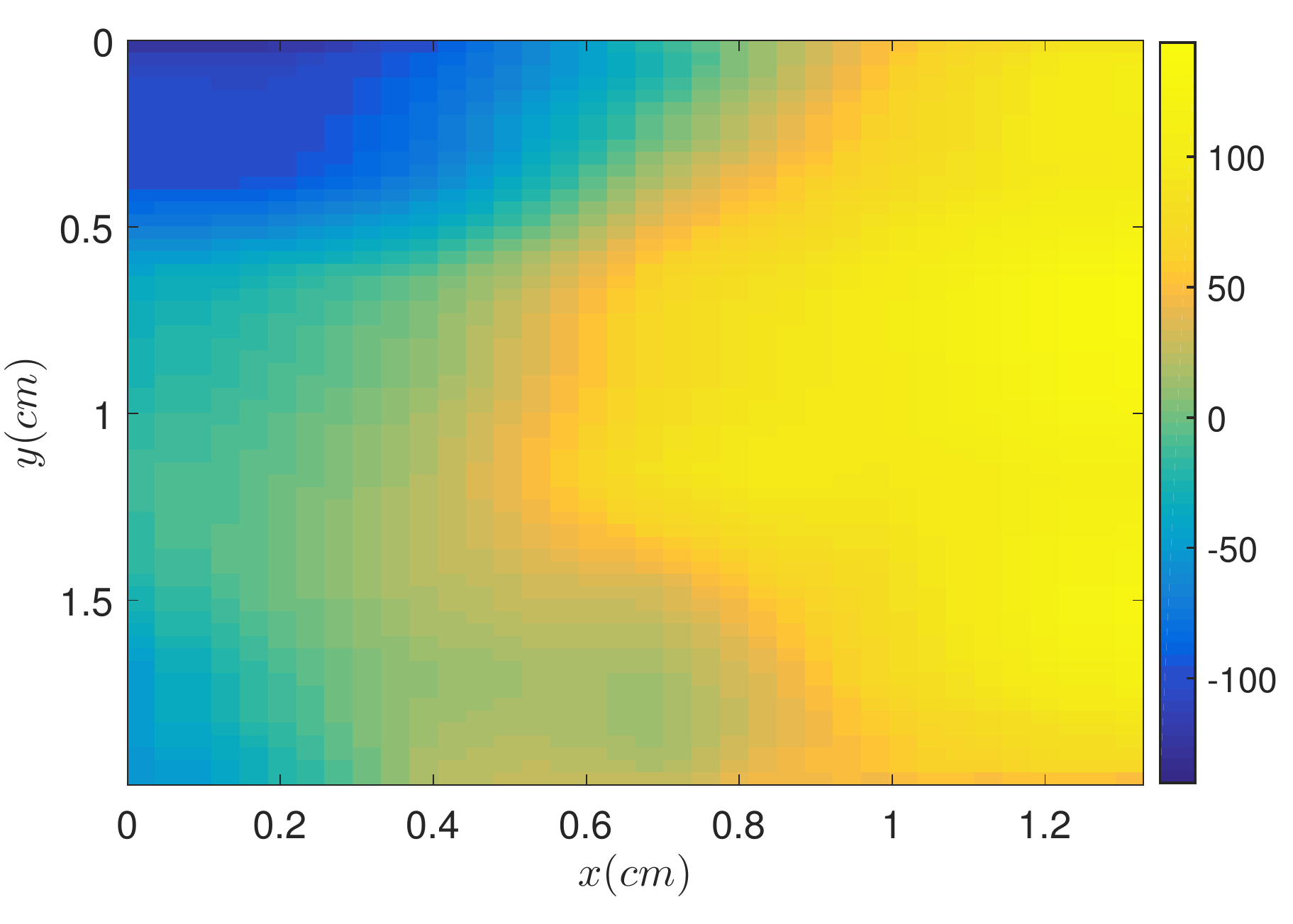}\put (45,-10) {(b.6)}\end{overpic}
&
\hspace{-.23cm}\begin{overpic}[trim={0.2cm 0.5cm 0.5cm 0.2cm}, clip,width=1.1in,height=.7in]{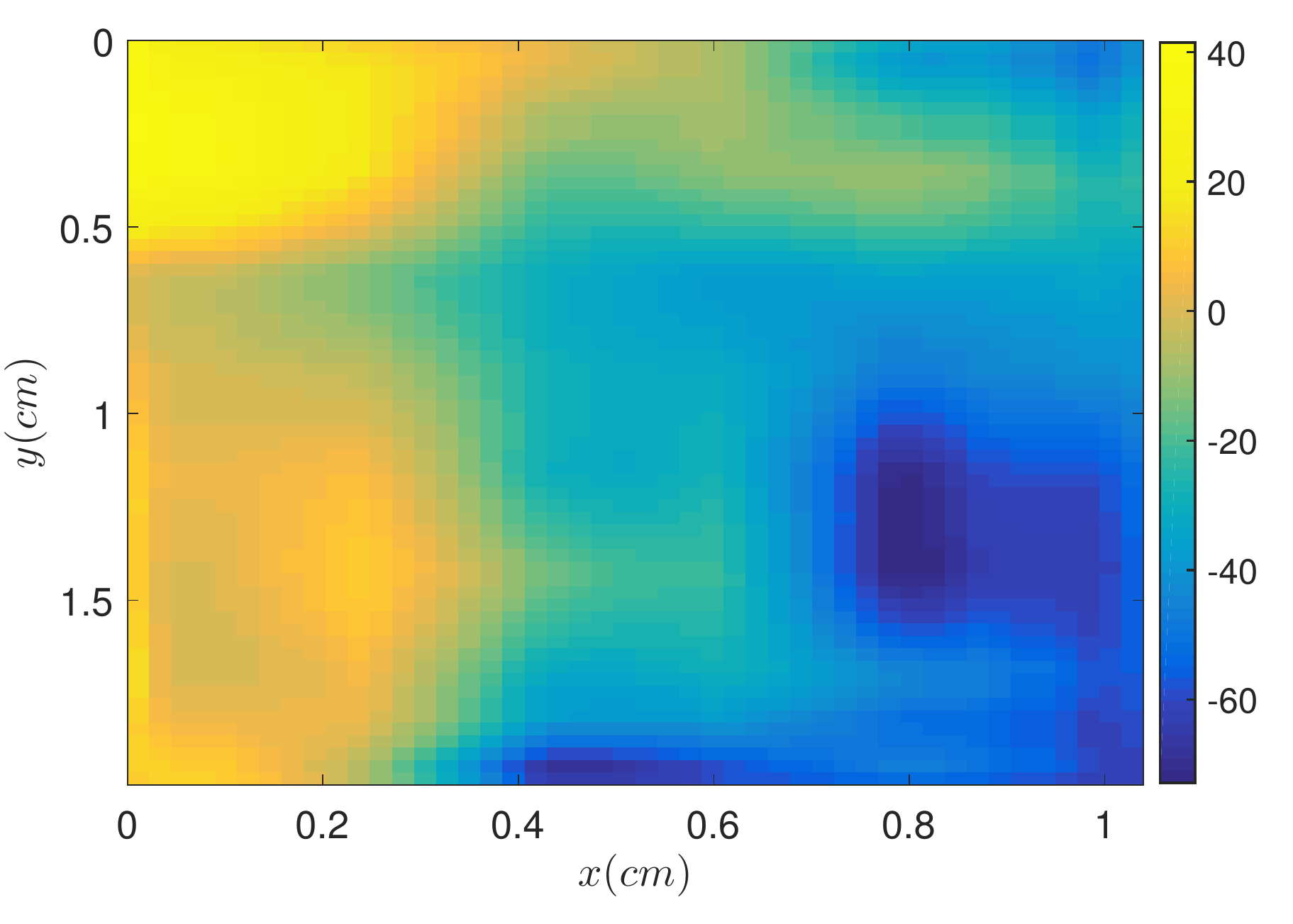}\put (45,-10) {(c.6)}
\end{overpic}
\\[.2cm]
\hspace{-.2cm}\begin{overpic}[trim={0.2cm 0.5cm 0.5cm 0.2cm}, clip,width=1.1in,height=.7in]{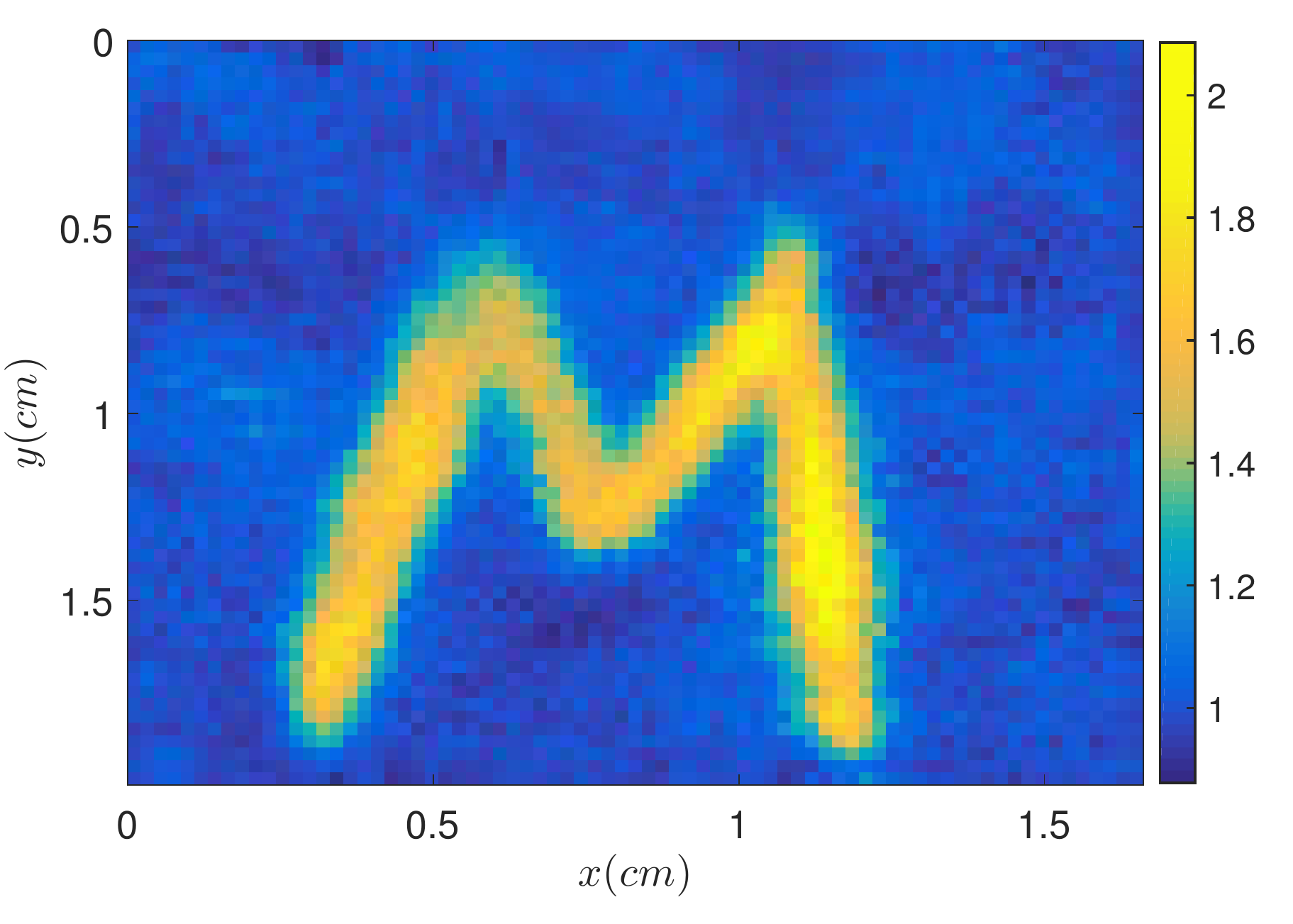}\put (45,-10) {(a.7)} \end{overpic}
&
\hspace{-.23cm}\begin{overpic}[trim={0.2cm 0.5cm 0.5cm 0.2cm}, clip,width=1.1in,height=.7in]{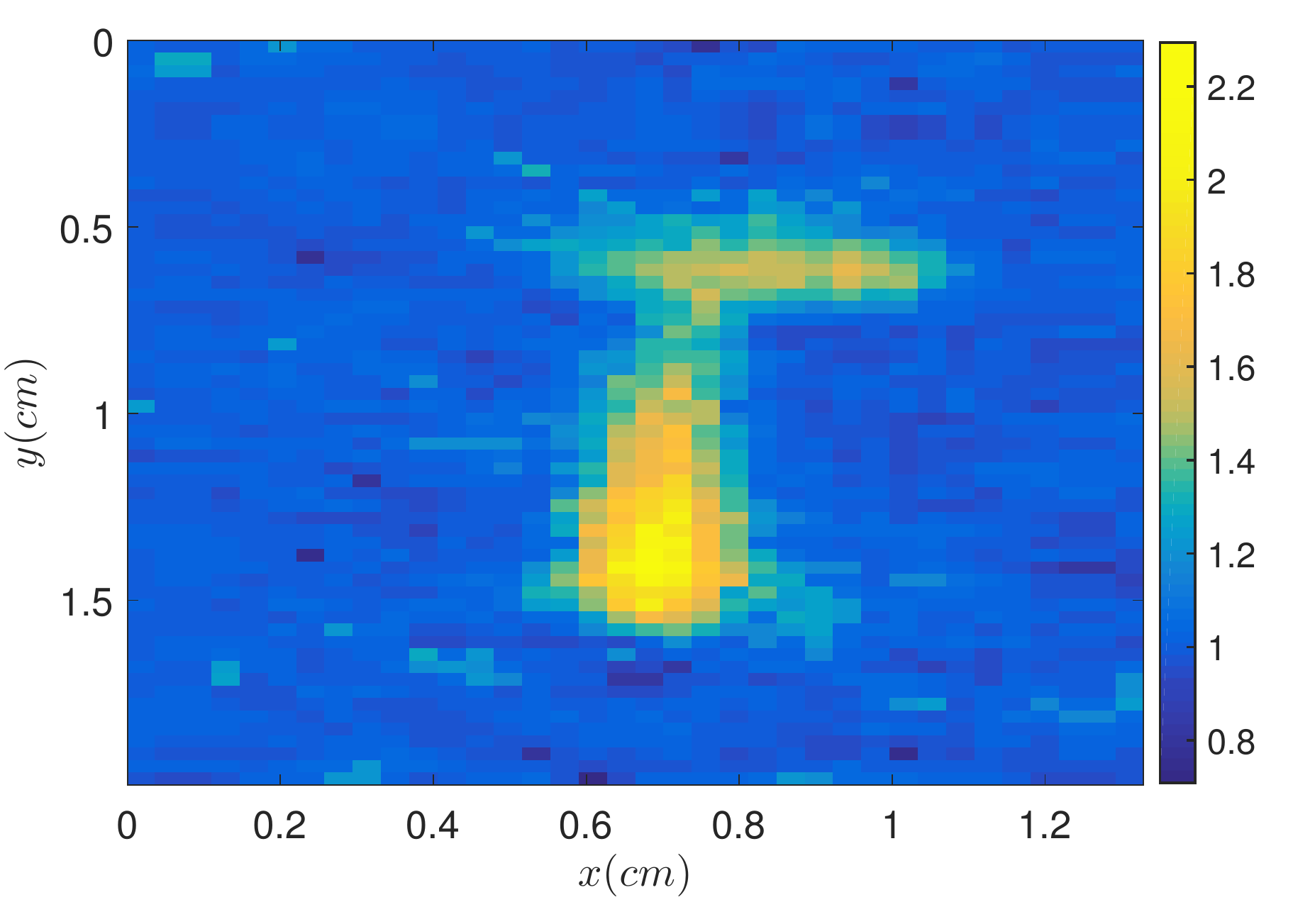}\put (45,-10) {(b.7)}\end{overpic}
&
\hspace{-.23cm}\begin{overpic}[trim={0.2cm 0.5cm 0.5cm 0.2cm}, clip,width=1.1in,height=.7in]{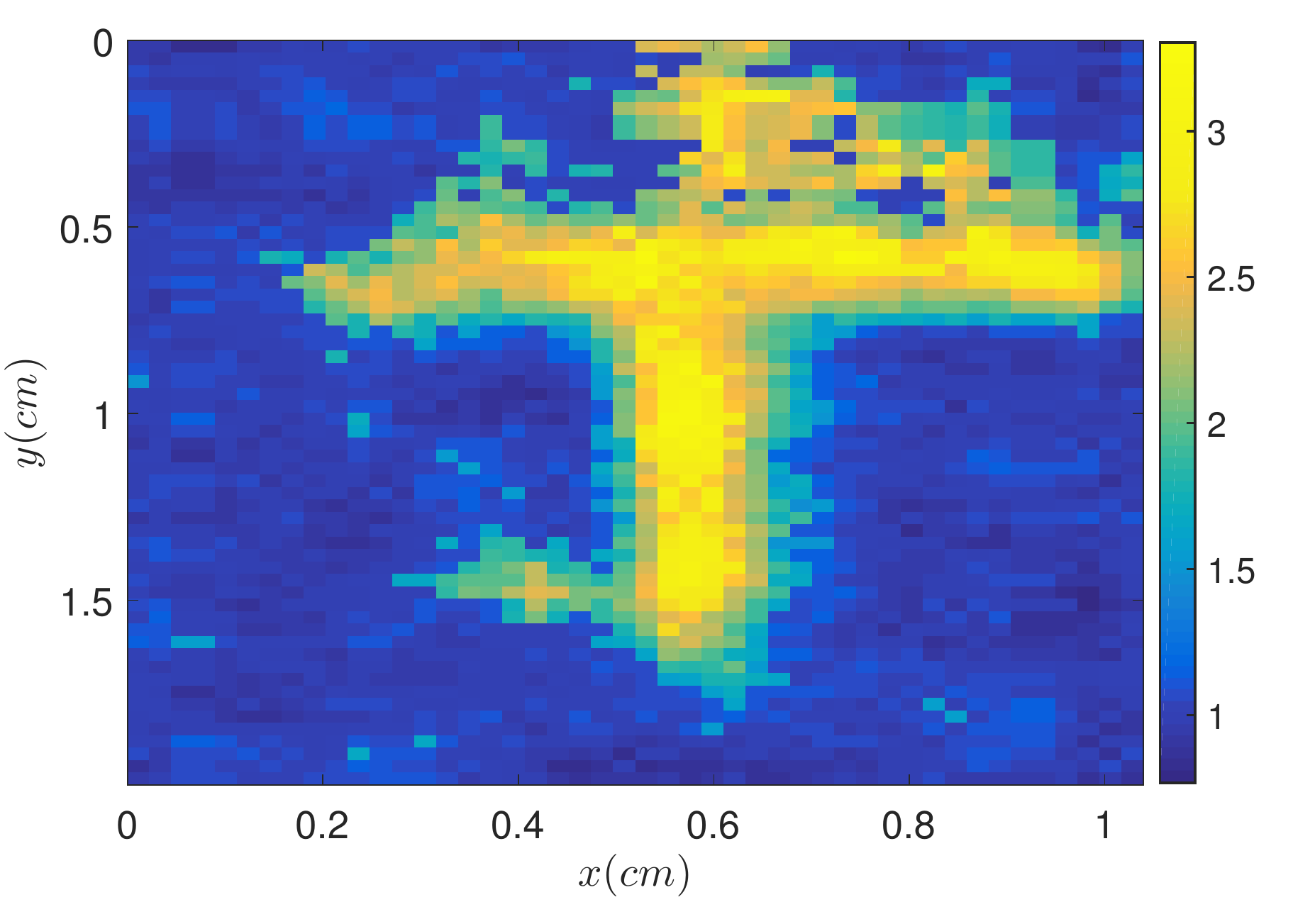}\put (45,-10) {(c.7)}
\end{overpic}
\\[.2cm]
\end{tabular}
\caption{Experimental demonstration of sweep decoupling for a paper stack sample; (a.1-3) three time instances of the recorded image from the first page; (b.1-3) second page; (c.1-3) third page, irregular distortions are dominantly present; (a.4-6) the algorithm has demodulated the distortion from the first page; (b.4-6) for the second page; (c.4-6) third page; (a.7) recovered letter “M” on the first page; (b.7) recovered letter “I” on the second page; (c.7) recovered letter “T” on the third page   }\label{fig7}
\end{figure}

A more challenging experiment uses a three layer paper sample with letters M-I-T printed on each page from front to back respectively. The page dimensions are 15 $mm\times 30$ $mm$ and the thicknesses are $300$ $\mu m$, stacked flush next to each other ($\sim \! 30$ $\mu m$ gap) similar to the structure of a book. Figure \ref{fig7}(a1-3) shows the reflected signal from the first layer at three different instances of time. Figures \ref{fig7}(b1-3) and \ref{fig7}(c1-3) show reflection samples from the second and third layers respectively. The available number of frames corresponding to the first, second and third layer are $M =$ 25, 6 , 8, respectively.

Followed by the observations in each column of Figure \ref{fig7}, the demodulation results are presented for the three letters at different pages of the sample, as the THz pulse travels deeper into the pages. The rows 4-6 in Figure \ref{fig7} correspond to the sweep distortion profiles decoupled from the observed data. The last row corresponds to the recovered character profiles.

For this multilayer setup, after the first layer the SNR drops and a larger number of inter-reflections is induced, thus the recovered letters ``I'' and ``T'' in panels (b.7) and (c.7) have more noise compared to the recovered letter ``M'' on the first page (panel (a.7)).  

The clean results from the first experiment (Figure \ref{fig6}) on the metallic surface are anticipated due to high SNR and perfectly flat surface of the polished steel, however for the paper the reflection is rather small (<0.15), the distortion contrast is at the same level of the signal contrast itself, and the pages have slight depth variations across them. The results from the paper sample in Figure \ref{fig7} indicate the true robustness of the decoupling technique for practical samples.  

It is worth noting that the major burden in characterizing the sample inhomogeneity profiles (characters in this example) is the varying signal level in the observed images. For instance, while the contrast between the character ``M'' and the background can be visually detected in Figure \ref{fig7}(a.3), the pixel values drastically vary from one portion of the letter to the other. As a result, characterizing the letter based on the pixel values becomes an inconceivable task. However, once a binary profile is reconstructed through the proposed algorithm, other post-processing schemes can be employed to characterize the reconstructions, especially in the case of the more noisy reconstructions such as Figure \ref{fig7}(c.7). The interested reader is referred to \cite{redo2016Terahertz, aghasi2015convex, aghasi2013sparse, aghasi2016object}, where advanced shape composition techniques are employed to accurately identify binary inclusions in noisy images. This class of techniques can accurately characterize the inclusion in an image like Figure \ref{fig7}(c.7), even when the noise is higher, parts of the object are missing due to signal occlusions, and we have clutter or overlapping characters caused by the inter-reflections from neighboring layers.

\subsection{Concluding Remarks}
The main outcome of this research is developing a general demodulation scheme to process the reflected signals, conveniently applicable to THz imaging. While a careful processing of the THz data can produce high resolution images thanks to its high frequency, dealing with phenomena such as inter-reflections, noise and modulated distortions is inevitable. We showed that reformulation of the problem as a bilinear inverse problem and making practical assumptions, such as the binary prior, can assist us with a successful inversion of THz measurements in THz-TDS systems.   

Using basic statistical assumptions about the image and the observation, we were able to develop an iterative inversion scheme, which is computationally efficient, distributable and scalable to big data sets. In fact, the main computational load at every step of the algorithm is a least squares solve. The remaining computations are in the form of thresholding-type operations. 

Since in many applications such as water profilometry, extracting coded structures, extracting sub surface cracks, the main objective is a binary characterization of the inhomogeneity in the samples, we used a binary prior in our modeling. The algorithm can be naturally generalized to the case of polyadic prior, where more than two levels are considered for the image profile. Basically, the proposed algorithm can be further modified or combined with post or pre-processing tools to handle a larger class of imaging applications.



\end{document}